\documentclass{article}
\usepackage[accepted]{icml2025}

\usepackage{amsmath,amsfonts,bm}

\def\eqref#1{equation~\ref{#1}}
\def\1{\bm{1}}

\DeclareMathAlphabet{\mathsfit}{\encodingdefault}{\sfdefault}{m}{sl}
\SetMathAlphabet{\mathsfit}{bold}{\encodingdefault}{\sfdefault}{bx}{n}

\newcommand\ts[1]{\texttt{#1}}

\newcommand{\bx}{{\bf x}}

\newcommand{\bv}{{\bf v}}

\newcommand{\bI}{{\bf I}}

\newcommand{\bw}{{\bf w}}
\newcommand{\bW}{{\bf W}}

\newcommand{\mc}{\mathcal}
\newcommand{\pre}{\text{pre}}

\newcommand\prob[1]{\mathbb{P}\left[ #1 \right]}

\newcommand\inner[2]{\langle #1, #2 \rangle}
\newcommand\norm[1]{\left\| #1 \right\|_2^2}
\newcommand\normt[1]{\left\| #1 \right\|_2}
\newcommand\ind[1]{\mathds{1}\big( #1 \big)}
\newcommand\abs[1]{\left| #1 \right|}
\newcommand\bigo[1]{\mathcal{O}\left( #1 \right)}
\newcommand\bigtheta[1]{\Theta\left( #1 \right)}
\newcommand\bigthetat[1]{\widetilde{\Theta}\left( #1 \right)}
\newcommand\bigO[1]{\mathcal{O}\left( #1 \right)}

\newcommand\bigw[1]{\Omega \left( #1 \right)}

\newcommand{\lbr}{\left[}
\newcommand{\rbr}{\right]}

\newcommand{\expt}{\mathbb{E}}

\newcommand{\bmu}{\bm{\mu}}
\newcommand{\snr}{\mathrm{SNR}}

\newcommand{\bxi}{\bm{\xi}}
\newcommand{\lmweight}{\bw_{j,r,k}}
\newcommand{\lmweightt}{\widetilde{\bw}_{j,r,k}}

\newcommand{\lm}{\bW_k}
\newcommand{\lmt}{\widetilde{\bW}_k}
\newcommand{\lgam}{\gamma_{j,r,k}}

\newcommand{\lrho}{\rho_{j,r,k,i}}
\newcommand{\lprho}{\overline{\rho}_{j,r,k,i}}
\newcommand{\lnrho}{\underline{\rho}_{j,r,k,i}}
\newcommand{\ggam}{\Gamma_{j,r}}

\newcommand{\ggamv}{\mathbb{G}_{j,r,k}}
\newcommand{\effsnr}{\widehat{\gamma}}

\newcommand{\grho}{P_{j,r,k,i}}

\newcommand{\gprho}{\overline{P}_{j,r,k,i}}

\newcommand{\gprhov}{\overline{\mathbb{P}}_{j,r,k,i}}
\newcommand{\gprhovy}{\overline{\mathbb{P}}_{y_{k,i},r,k,i}}
\newcommand{\gnrho}{\underline{P}_{j,r,k,i}}

\newcommand{\gnrhov}{\underline{\mathbb{P}}_{j,r,k,i}}

\newcommand{\gmweight}{\bw_{j,r}}

\newcommand{\gm}{\bW}
\newcommand{\wopt}{\bW^*}
\newcommand{\woptw}{\bw^*}
\newcommand{\lderiv}{{\ell'}_{k,i}}
\newcommand{\lderivk}{{\ell'}_{k',i}}
\newcommand{\lderivd}{{\ell'}_{k',i'}}
\newcommand{\lderivone}{{\ell'}_{k,1}}
\newcommand\noisederiv[1]{\sigma^{'}\big( \langle #1, \bxi_{k,i} \rangle\big)}
\newcommand\signalderiv[1]{\sigma^{'}\big( \langle #1, y_{k,i}\bmu \rangle\big)}
\newcommand{\nn}{\nonumber}

\usepackage{url}
\usepackage{amsmath,amsfonts,bm}
\usepackage[normalem]{ulem}
\usepackage{nicefrac}
\usepackage{enumitem}[enumerate,itemize,description]
\usepackage{amssymb}
\usepackage{mathtools, nccmath, textcomp}
\usepackage{amsmath}
\usepackage[utf8]{inputenc} % allow utf-8 input
\usepackage[T1]{fontenc}    % use 8-bit T1 fonts
\usepackage{booktabs}       % professional-quality tables
\usepackage{wrapfig,lipsum,booktabs}
\usepackage{bm} % blackboard math symbols
\usepackage{nicefrac}       % compact symbols for 1/2, etc.
\usepackage{microtype}      % microtypography
\usepackage{xcolor}         % colors
\usepackage{algorithm}
\usepackage{algorithmic}
\usepackage{amsthm}
\usepackage{titletoc}
\usepackage{subfig}
\usepackage{multirow}
\usepackage{makecell}
\usepackage{mathtools}
\usepackage{adjustbox}
\usepackage{booktabs}
\usepackage{wrapfig}
\usepackage{dsfont}
\usepackage{tcolorbox}
\usepackage[font=small]{caption}
\usepackage{hyperref}
\usepackage{cleveref}

\DeclarePairedDelimiter\floor{\lfloor}{\rfloor}

\newtheorem{lemma}{Lemma}
\newtheorem{theorem}{Theorem}
\newtheorem*{thm}{Theorem}
\newtheorem{definition}{Definition}

\newtheorem{proposition}{Proposition}

\newtheorem{assumption}{Assumption}
\newtheorem{condition}{Condition}

\icmltitlerunning{Initialization Matters: Unraveling the Impact of Pre-Training on Federated Learning}

\begin{document}

\twocolumn[
\icmltitle{Initialization Matters: Unraveling the Impact of Pre-Training on Federated Learning}

\icmlsetsymbol{equal}{*}

\begin{icmlauthorlist}
\icmlauthor{Divyansh Jhunjhunwala}{yyy}
\icmlauthor{Pranay Sharma}{yyy}
\icmlauthor{Zheng Xu}{comp}
\icmlauthor{Gauri Joshi}{yyy}
\end{icmlauthorlist}

\icmlaffiliation{yyy}{Carnegie Mellon University}
\icmlaffiliation{comp}{Google Research}

\icmlcorrespondingauthor{Divyansh Jhunjhunwala}{djhunjhu@andrew.cmu.edu}

\icmlkeywords{Pre-training, Initialization, CNN, Federated Learning}

\vskip 0.3in
]



\printAffiliationsAndNotice{}  % leave blank if no need to mention equal contribution


\begin{abstract}
Initializing with pre-trained models when learning on downstream tasks is becoming standard practice in machine learning. Several recent works explore the benefits of pre-trained initialization in a federated learning (FL) setting, where the downstream training is performed at the edge clients with heterogeneous data distribution. These works show that starting from a pre-trained model can substantially reduce the adverse impact of data heterogeneity on the test performance of a model trained in a federated setting, with no changes to the standard FedAvg training algorithm. In this work, we provide a deeper theoretical understanding of this phenomenon. To do so, we study the class of two-layer convolutional neural networks (CNNs) and provide bounds on the training error convergence and test error of such a network trained with FedAvg. We introduce the notion of \textit{aligned} and \textit{misaligned} filters at initialization and show that the data heterogeneity only affects learning on misaligned filters. Starting with a pre-trained model typically results in fewer misaligned filters at initialization, thus producing a lower test error even when the model is trained in a federated setting with data heterogeneity.
Experiments in synthetic settings and practical FL training on CNNs verify our theoretical findings.
\end{abstract}

\section{Introduction}
\label{sec:intro}

Federated Learning (FL) \citep{mcmahan2017communication} has emerged as the de-facto paradigm for training a Machine Learning (ML) model over data distributed across multiple clients with privacy protection due to its no data-sharing philosophy. Ever since its inception, it has been observed that heterogeneity in client data can severely slow down FL training and lead to a model that has poorer generalization performance than a model trained on Independent and Identically Distributed (IID) data \citep{kairouz2021advances, li2020federatedchallenge, yang2021characterizing}. This has led works to propose several \textit{algorithmic} modifications to the popular Federated Averaging (\ts{FedAvg}) algorithm such as variance-reduction \citep{acar2021federated, karimireddy2020scaffold}, contrastive learning \citep{li2021model, tan2022federated} and sophisticated model-aggregation techniques \citep{lin2020ensemble, wang2020federated}, to combat the challenge of data heterogeneity.

\begin{figure}
\centering 
\includegraphics[width=0.33\textwidth]{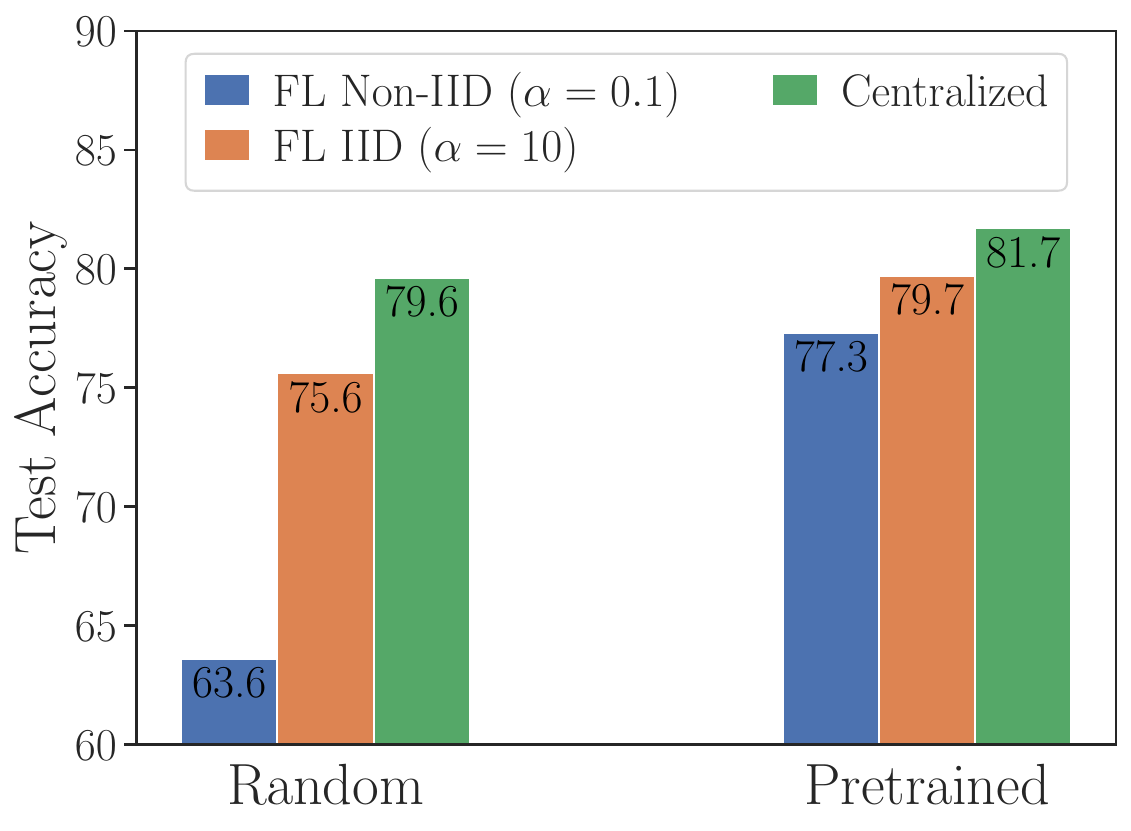}\vspace{-1em}
\caption{Test accuracy ($\%$) on CIFAR10 with SqueezeNet model \cite{iandola2016squeezenet} under random and pretrained initializations for FL and centralized training. Pre-training benefits FL more than centralized setting and significantly reduces the gap between IID and non-IID FL model performance.}
\label{fig:test_acc_gap}
\vspace{-2em}
\end{figure}

A recent line of work \citep{chen2022importance, nguyen2022begin} has sought to understand the benefits of starting from \textit{pre-trained} models instead of randomly initializing the global model when doing FL. This idea has been popularized by results in the centralized setting \citep{Devlin2019BERTPO, radford2019language, he2019rethinking, dosovitskiy2020image}, which show that starting from a pre-trained model can lead to state-of-the-art accuracy and faster convergence on downstream tasks. Pre-training is usually done on internet-scale public data \citep{schuhmann2022laion, thomee2016yfcc100m, raffel2020exploring, gao2020pile} in order for the model to learn fundamental data representations \citep{sun2017revisiting,mahajan2018exploring,radford2019language}, that can be easily applied for downstream tasks. 
Thus, while it would not be unexpected to see some gains of using pre-trained models even in FL, what is surprising is the sheer scale of improvement.
In many cases \cite{nguyen2022begin, chen2022importance} show that just starting from a pre-trained model can significantly reduce the gap between the performance of a model trained in a federated setting with non-IID versus IID data partitioning with \textit{no algorithmic modifications}.
\Cref{fig:test_acc_gap} shows our own replication of this phenomenon, where starting from a pre-trained model can lead to almost $14\%$ improvement in accuracy for FL with non-IID data (i.e., high data heterogeneity) compared to $4\%$ for FL with IID data and $2\%$ in the centralized setting. This observation leads us to ask the question:
\begin{center}
\vspace{-0.5em}
    \textit{Why can pre-trained initialization drastically improve model performance in FL?}
\end{center}

One reason suggested by \cite{nguyen2022begin} is a lower value of the training loss at initialization when starting from pre-trained models. However, this observation can only explain improvement in training convergence speed
(see Theorem V in \cite{karimireddy2021breaking}) and not the significantly improved generalization performance of the trained model. Also, a pre-trained initialization can have larger loss than random initialization while continuing to have faster convergence and better generalization (see Table 1 in \cite{nguyen2022begin}). \citet{chen2022importance, nguyen2022begin}, also observe some optimization-related factors when starting from a pre-trained model including smaller distance to optimum, better conditioned loss surface (smaller value of the largest eigen value of Hessian) and more stable global aggregation. However, it has not been formally proven that these factors can reduce the adverse effect of non-IID data. Thus, there is still a lack of fundamental understanding of why pre-trained initialization benefits generalization for non-IID FL.

\textbf{Our contributions.} In this work we provide a deeper theoretical understanding of the importance of initialization for \ts{FedAvg} by studying two-layer ReLU Convolutional Neural Networks (CNNs) for binary classification. This class of neural networks lends itself to tractable analysis while providing valuable insights that extend to training deeper CNNs as shown by several recent works \citep{cao2022benign, du2018gradient,kou2023benign, zou2021understanding, jelassi2022towards, baoprovable, oh2024provable}.
 Our data generation model, also studied in \cite{cao2022benign, kou2023benign}, allows us to utilize a \textit{signal-noise decomposition} result (see \Cref{prop:decomposition}) to perform a fine-grained analysis of the CNN filter weight updates than can be done with general non-convex optimization.
Some highlights of our results are as follows:
\vspace{-1em}
\begin{enumerate}[leftmargin=*]
    \item We introduce the notion of \textit{aligned} and \textit{misaligned} filters at initialization (\Cref{lem:main_paper_lemma_signal_growth}) and show that data heterogeneity affects signal learning only on misaligned filters while noise memorization is unaffected by data heterogeneity (see \Cref{lem:main_paper_noise_growth}). A pre-trained model is expected to have fewer misaligned filters, which can explain the reduced effect of non-IID data.
    
    \item We provide a test error upper bound for \ts{FedAvg} that depends on the number of misaligned filters at initialization and data heterogeneity. The effect of data heterogeneity on misaligned filters is exacerbated as clients perform more local steps, which explains why FL benefits more from pre-trained initialization than centralized training. To our knowledge, this is the first result where the test error for \ts{FedAvg} explicitly depends on initialization conditions (\Cref{thm:test_error}).
    
    \item We prove the training error convergence of \ts{FedAvg} by adopting a two-stage analysis: a first stage where the local loss derivatives are lower bounded by a constant and second stage where the model is in the neighborhood of a global minimizer with nearly convex loss landscape. Our analysis shows a provable benefit of using local steps in the first stage to reduce communication cost.

    \item We experimentally verify our upper bound on the test error in a synthetic data setting  (see \Cref{sec:main_results} as well as conduct experiments on practical FL tasks which show that our insights extend to deeper CNNs (see \Cref{sec:expts}). 
\end{enumerate}

\textbf{Related Work.} The two-layer CNN model that we study in this work was originally introduced in \cite{zou2021understanding} for the purpose of analyzing the generalization error of the \ts{Adam} optimizer in the centralized setting. Later \cite{cao2022benign} study the same model to analyze the phenomenon of \textit{benign overfitting} in two-layer CNN, i.e., give precise conditions under which the CNN can perfectly fit the data while also achieving small population loss. \cite{oh2024provable} use this model to prove the benefit of patch-level data augmentation techniques such as Cutout and CutMix. \cite{kou2023benign} relaxes the the polynomial ReLU activation in \cite{cao2022benign} to the standard ReLU activation and also introduces label-flipping noise when analyzing benign overfitting in the centralized setting. We do not consider label-flipping in our work for simplicity; however this can be easily incorporated as future work. To the best of our knowledge, we are only aware of two other works \citep{huang2023understanding, baoprovable} that analyze the two-layer CNN in a FL setting. The focus in \cite{huang2023understanding} is on showing the benefit of collaboration in FL by considering signal heterogeneity across the data in clients while \cite{baoprovable} considers signal heterogeneity to show the benefit of local steps. Both \cite{huang2023understanding} and \cite{baoprovable} do not consider any label heterogeneity and there is no emphasis on the importance of initialization, making their analysis quite different from ours. We defer more discussion on other related works to the Appendix. 

\section{Problem Setup}
\label{sec:problem_setup}
We begin by introducing the data generation model and the two-layer convolutional neural network, followed by our FL objective and a brief primer on the \ts{FedAvg} algorithm. We note that given integers $a,b$, we denote by $[a:b]$ the set of integers $\{a,a+1,\dots,b \}$. Also, $[n]$ denotes $\{1,2,\dots,n\}$. 

 \paragraph{Data-Generation Model.} 
 Let $\mathcal{D}$ be the global data distribution. 
 A datapoint $(\bx, y) \sim \mathcal{D}$ contains feature vector $\bx = [\bx(1)^{\top}, \bx(2)^{\top}]^\top \in \mathbb{R}^{2d}$ with two components $ \bx(1), \bx(2) \in \mathbb{R}^d$ and label $y \in \{+1,-1\}$, that are generated as follows:
 \vspace{-1em}
 \begin{enumerate}[leftmargin=*]
     \item Label $y \in \{-1, 1\}$ is generated as  $\prob{y=1} = \prob{y=-1} = 1/2$.
     \item One of $\bx(1)$, $\bx(2)$ is chosen at random and assigned as $y \bmu$, where $\bmu \in \mathbb{R}^{d}$ is the signal vector that we are interested in learning. The other of $\bx(1)$, $\bx(2)$ is set to be the noise vector $\bxi \in \mathbb{R}^{d}$, which is generated from the Gaussian distribution $\mathcal{N}(\bm{0},\sigma_p^2\cdot(\bI - \bmu\bmu^{\top}\cdot\normt{\bmu}^{-2}))$. 
 \end{enumerate}
 \vspace{-1em}
This data generation model is inspired by image classification tasks \cite{cao2022benign} where it has been observed that only some of the image patches (for example, the foreground) contain information (i.e. the signal) about the label. We would like the model to predict the label by focusing on such informative image patches and ignoring background patches that act as noise and are irrelevant to the classification. Note that by definition, the noise vector $\bxi$ is orthogonal to the signal $\bmu$, i.e., $\bxi^{\top}\bmu = 0$. We assume orthogonality just for simplicity of analysis and can be easily relaxed as done in . Our theoretical insights will remain the same with the only difference being that we need a slightly stronger condition on the dimension of the filters (\ref{assump_d}). 

\paragraph{Measure of Data Heterogeneity.} We consider $n$ datapoints drawn from the distribution $\mathcal{D}$, and partitioned across $K$ clients such that each client has $N = n/K$ datapoints. The assumption of equal-sized client datasets is made for simplicity of analysis and can be easily relaxed. The data partitioning determines the level of heterogeneity across clients.

Let $D_{+,k}$ and $D_{-,k}$ denote the set of samples at client $k$ with positive ($y = +1$) and negative ($y = -1$) labels respectively. Define
\begin{align}
\textstyle
    h := \mfrac{\sum_{k=1}^K \min\left(\abs{D_{+,k}},\abs{D_{-,k}}\right)}{n} \in [ 0,\nicefrac{1}{2} ]. \label{eq:data_hetero}
\end{align}
Note that a smaller $h$ implies a higher data heterogeneity. In the IID setting, with uniform partitioning across clients, we expect $ \min(\abs{D_{+,k}},\abs{D_{-,k}}) \approx \nicefrac{n}{2K}$
for all $k \in [K]$, and therefore $h \approx \nicefrac{1}{2}$. In the extreme  non-IID setting where each client only has samples from one class, $h = 0$.

\vspace{-0.5em}
\paragraph{Two-Layer CNN.} We now describe our two-layer CNN model. The first layer in our model consists of $2m$ filters $\{ \bw_{j, r}\}_{r=1}^{m}, j \in \{\pm 1\}$, where each $\bw_{j,r} \in \mathbb{R}^d$ performs a 1-D convolution on the feature $\bx$ with stride $d$ followed by ReLU activation and average pooling \cite{lin2013network, yu2014mixed}. The weights in the second layer then aggregate the outputs produced after pooling to get the final output and are fixed as $2/m$ for $j=+1$ filters and $-2/m$ for $j=-1$ filters. Formally, we have,

\begin{align}
\textstyle
    f(\bW, \bx)
    & = \underbrace{\mfrac{1}{m}\sum_{r=1}^m \left[ \sigma\left( \inner{\bw_{+1,r}}{y\bmu} \right) +  \sigma\left( \inner{\bw_{+1,r}}{\bxi} \right)\right]}_{:=F_{+1}(\bW_{+1},\bx)} \nn \\
    \vspace{-0.5em}
    & - \underbrace{\mfrac{1}{m}\sum_{r=1}^{m} \left[ \sigma\left( \inner{\bw_{-1,r}}{y\bmu} \right) +  \sigma\left( \inner{\bw_{-1,r}}{\bxi} \right)\right]}_{:=F_{-1}(\bW_{-1},\bx)}
    \label{eq:simplify_cnn_exp}.
\end{align}
Here $\bW \in \mathbb{R}^{2md}$ parameterizes all the weights of our neural network, $\bW_{+1}, \bW_{-1} \in \mathbb{R}^{md}$ parameterize the weights of the $j=+1$ filters and $j=-1$ filters respectively, and $\sigma(z) = \max(0,z)$ is the ReLU activation. Intuitively $F_{j}(\bW_j,\bx)$ represents the `logit score' that the model assigns to label $j$. 

 \paragraph{FL Training and Test Objectives.} Let $\{(\bx_{k,i}, y_{k,i})\}_{i=1}^N$ be the local dataset at client $k$. Then the global FL objective can be written as follows:
 \begin{align}
 \label{eq:global_obj}
 \textstyle
     & \min_{\bW \in \mathbb{R}^{2d}} \left\{ L(\bW) = \mfrac{1}{K} \textstyle\sum_{k=1}^K L_k(\bW)\right\},  \nonumber \\
     & L_k(\bW) = \mfrac{1}{N} \textstyle \sum_{i=1}^N \ell(y_{k,i} f(\bW,\bx_{k,i})),
 \end{align}
where $L_k(\bW)$ is the local objective at client $k$ and $\ell(z) = \log (1+\exp(-z))$ is the cross-entropy loss. We also define the test-error $L_{\mathcal{D}}^{0-1}$ as the probability that $\bW$ will misclassify a point $(\bx,y) \sim \mathcal{D}$:
\begin{align}
    L_{\mathcal{D}}^{0-1}(\bW) := \mathbb{P}_{(\bx,y) \sim \mathcal{D}} \left(y \neq \mathrm{sign}(f(\bW, \bx))\right).
\end{align}

\paragraph{The FedAvg Algorithm.} The standard approach to minimizing objectives of the form in \Cref{eq:global_obj} is the \ts{FedAvg} algorithm. In each round $t$ of the algorithm, the central server sends the current global model $\bW^{(t)}$ to the clients. Clients initialize their local models to the current global model by setting $\bW_k^{(t,0)} = \bW^{(t)}$, for all $k \in [K]$, and run $\tau$ local steps of gradient descent (GD) as follows
\begin{align}
\textstyle
    \text{Local GD:} \hspace{5pt} \bW_k^{(t,s+1)} = \bW_k^{(t, s)} - \eta \nabla L_k(\bW_k^{(t,s)}) \label{eq:localGD}
\end{align}
for all $s \in [0:\tau-1]$ and for all $k \in [K]$. After $\tau$ steps of Local GD, the clients send their local models $\{\bW_k^{(t,\tau)}\}$ to the server, which aggregates them to get the global model for the next round: $\bW^{(t+1)} = \sum_{k=1}^K \bW_k^{(t,\tau)}/K$. While we focus on FedAvg with local GD in this work, we note that several modifications such as stochastic gradients instead of full-batch GD, partial client participation \cite{yang2021achieving} and server momentum \cite{reddi2020adaptive} are considered in both theory and practice. Studying these modifications is an interesting future research direction.

\section{Main Results}
\label{sec:main_results}
In this section we first introduce our definition of filter alignment at initialization and a fundamental result regarding the signal-noise decomposition of the CNN filter weights. We then state our main result regarding the convergence of \ts{FedAvg} with random initialization for the problem setup described in \Cref{sec:problem_setup} and the impact of data heterogeneity and filter alignment at initialization on the test-error. Later we discuss why starting from a pre-trained model can improve the test accuracy of \ts{FedAvg}.

\subsection{Filter Alignment at Initialization} 

Given datapoint $(\bx,y)$, for the CNN to correctly predict the label $y$ and minimize the loss $\ell(yf(\bW,\bx))$, from \eqref{eq:simplify_cnn_exp}-\eqref{eq:global_obj}, we want $yf(\bW,\bx) = F_{y}(\bW_{y},\bx) - F_{-y}(\bW_{-y},\bx)) \gg 0$.  At an individual filter $r \in [m]$, this can happen either with $\inner{\bw_{y,r}}{y\bmu} \gg 0$ or $\inner{\bw_{y,r}}{\bxi} \gg 0$. However, we want the model to focus on the signal $y\bmu$ in $\bx$ while making the prediction. Therefore, for filter $(j,r)$ we want $\inner{\bw_{j,r}}{y\bmu} \gg 0$ if $j = y$ and $\inner{\bw_{j,r}}{y\bmu} \ll 0$ if $j = -y$. Depending on the initialization of our CNN, we have the following definition of \textit{aligned} and \textit{misaligned} filters.

\begin{definition}
\label{def:signal_alignment} The $(j,r)$-th filter (with $j \in \{\pm 1\}, r \in [m]$) is said to be aligned (with signal) at initialization if $\inner{\bw_{j,r}^{(0)}}{j\bmu} \geq 0$ and misaligned otherwise.
\end{definition}

We shall see in \Cref{subsec:signal_noise_growth} that the alignment of a filter at initialization plays a crucial role in how well it learns the signal and also the overall generalization performance of the CNN in \Cref{thm:test_error}.

\subsection{Signal Noise Decomposition of CNN Filter Weights}
\label{sec:signal_decomp}

One of the key insights in \cite{cao2022benign} is that when training the two-layer CNN with GD, the filter weights at each iteration can be expressed as a linear combination of the initial filter weights, signal vector and noise vectors. Our first result below shows that this is true for \ts{FedAvg} as well.

\begin{proposition}
\label{prop:decomposition}
Let $\{\bw_{j,r}^{(t)}\}$, for $j \in \{\pm 1\}$ and $r \in [m]$, be the global CNN filter weights in round $t$. Then there exist unique coefficients $\Gamma_{j,r}^{(t)} \geq 0$ and $\{ P_{j,r,k,i}^{(t)} \}_{k,i}$ such that
\begin{align}
\textstyle
\label{eq:global_filter_decomp}
    \bw_{j,r}^{(t)} & = \bw_{j,r}^{(0)} + \underbrace{j\Gamma_{j,r}^{(t)}\cdot\normt{\bmu}^{-2}\cdot\bmu}_{\text{Signal Term}} \nonumber \\
    & + \underbrace{\textstyle\sum_{k=1}^K\sum_{i=1}^N \grho^{(t)}\cdot\normt{\bxi_{k,i}}^{-2}\cdot\bxi_{k,i}}_{\text{Noise Term}},
\end{align}
where $k \in [K]$ and $i \in [N]$ denote the client and sample index respectively.
\end{proposition}

This decomposition allows us to decouple the effect of the signal and noise components on the CNN filter weights, and analyze them separately throughout training. 

As we run more communication rounds (denoted by $t$), we expect the weights to learn the signal $y\bmu$, hence it is desirable for $\ggam^{(t)}$ to increase with $t$.
In addition, the filter weights also inevitably memorize noise $\bxi$ and overfit to it, therefore the noise coefficients $\{ \grho^{(t)}\}$ will also grow with $t$. We are primarily interested in the growth of positive noise coefficients $\gprho^{(t)} = \grho^{(t)}\ind{\grho^{(t)} \geq 0}$ since the negative noise-coefficients $\gnrho^{(t)} := \grho^{(t)} \ind{\grho^{(t)} \leq 0}$ remain bounded (see \Cref{thm:coeff_bound} in \Cref{sec:assumptions_and_results}) and we can show that $\sum_{k,i}\grho^{(t)} = \Theta (\sum_{k,i} \gprho^{(t)})$. Henceforth, we refer to $\ggam^{(t)}$ and $\sum_{k,i} \gprho^{(t)}$, as the \textit{signal learning} and \textit{noise memorization} coefficients of filter $(j,r)$ respectively. As we see later in \Cref{thm:test_error}, the ratio of signal learning to noise memorization $\ggam^{(t)}/\sum_{k,i} \gprho^{(t)}$ is fundamental to the generalization performance of the CNN.

\begin{figure*}[t]
  \centering
  \subfloat[]{\includegraphics[width=0.3\linewidth]{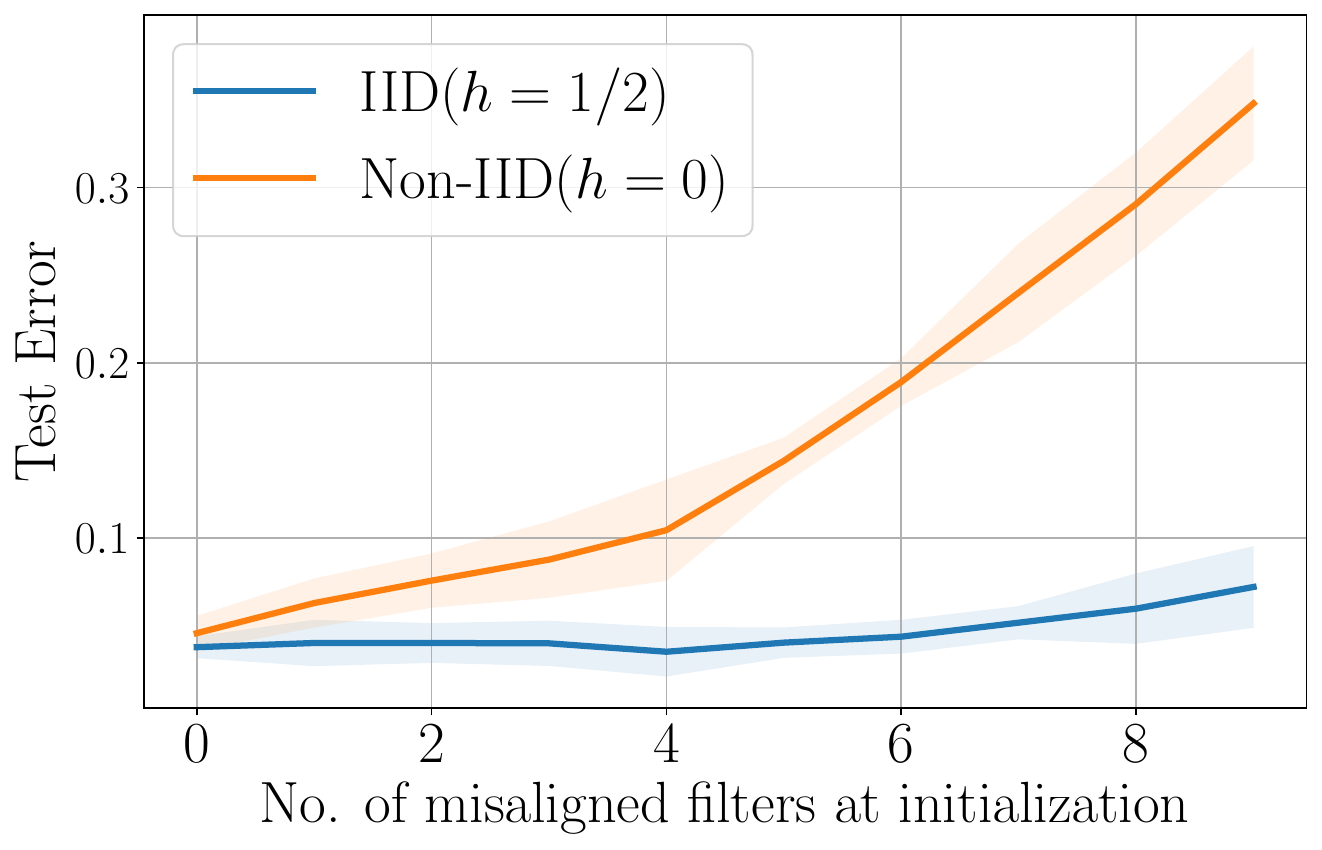}
  \label{subfig:filter_alignment}}
   \subfloat[]{\includegraphics[width=0.3\linewidth]{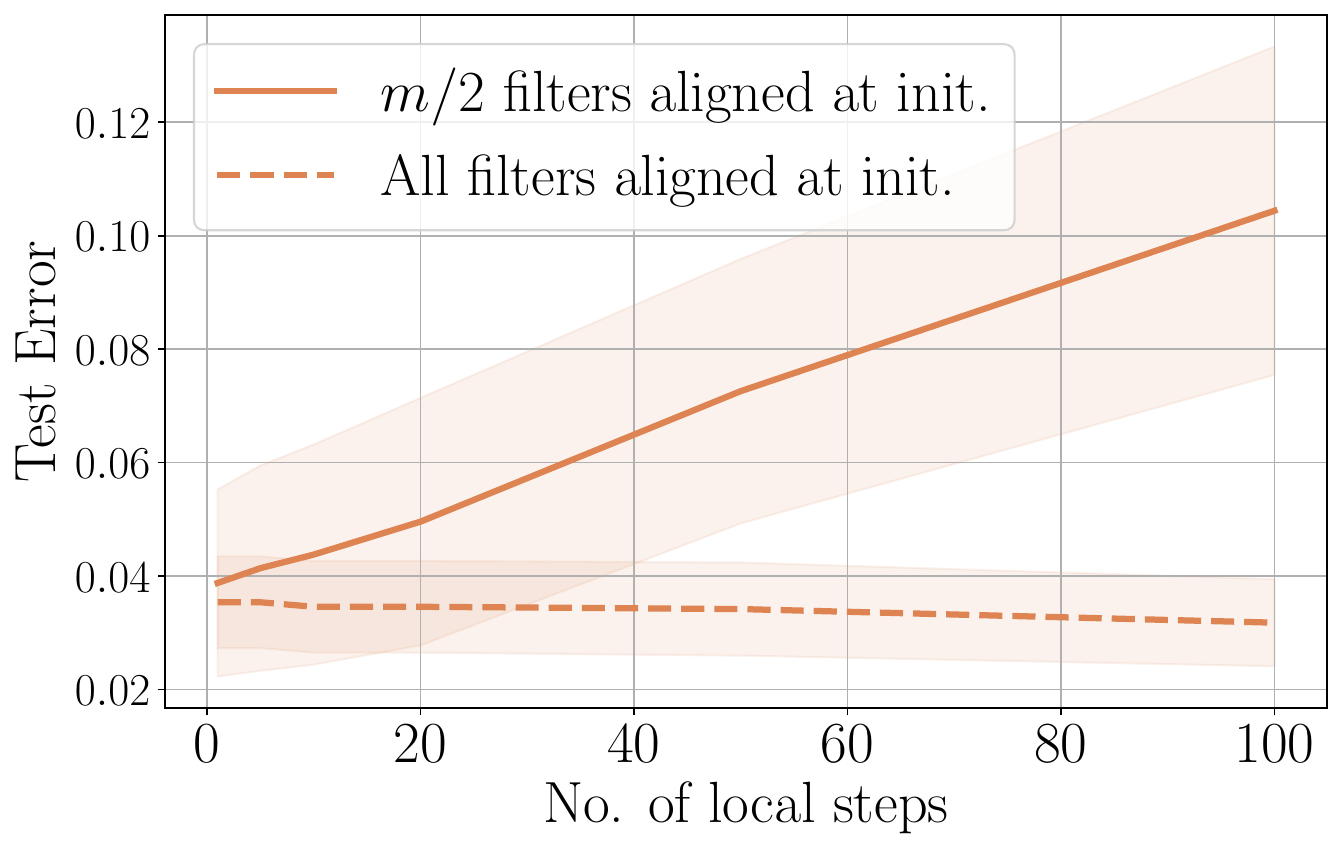}
   \label{subfig:local_steps}}
   \subfloat[]{\includegraphics[width=0.3\linewidth]{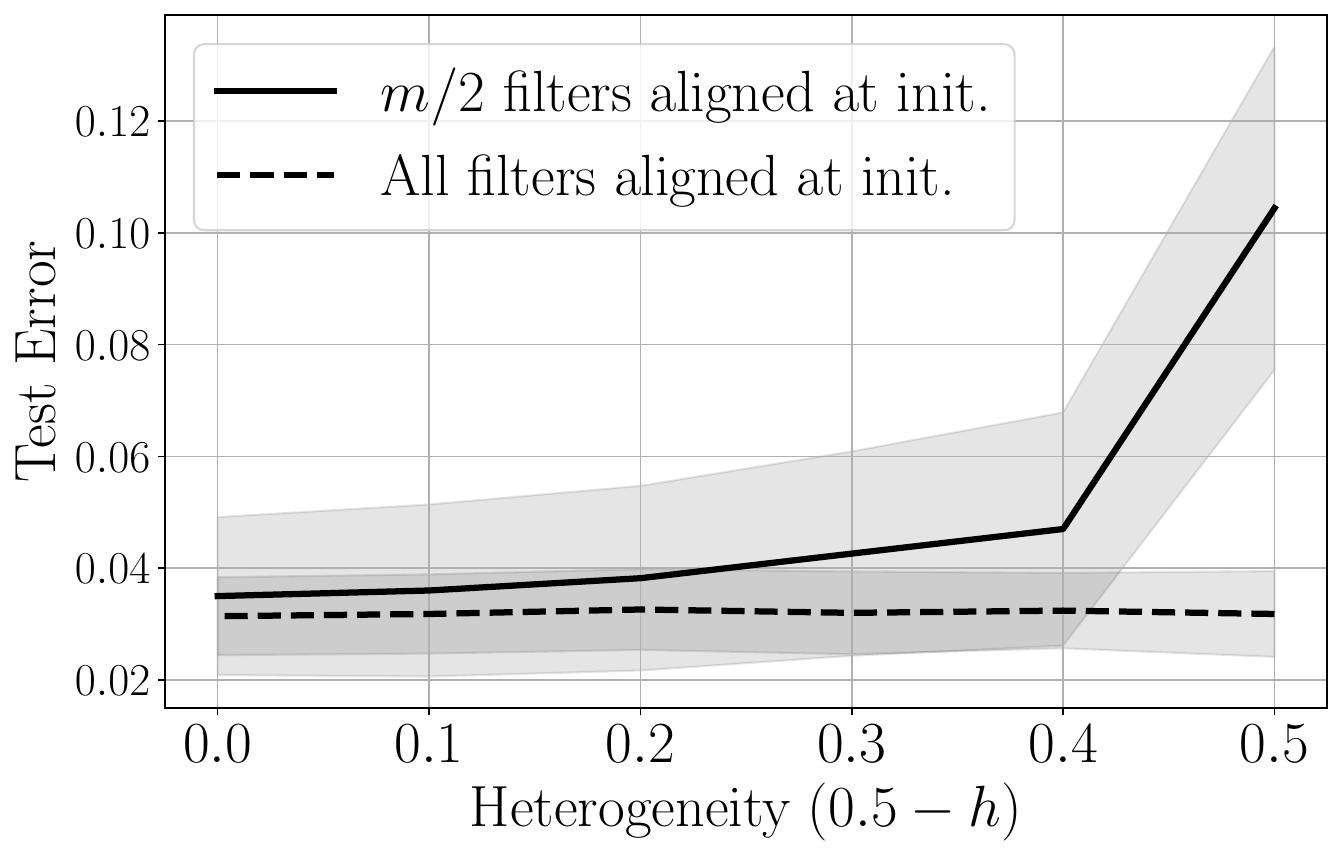}
    \label{subfig:heterogeneity}}
\vspace{-0.5em}
\caption{ Empirical results on synthetic dataset to verify the upper bound on test error in \Cref{thm:test_error}. We fix the training error $\epsilon = 0.1$.
\Cref{subfig:filter_alignment}: Test error increases as we increase the number of misaligned filters, with much larger rate of increase in the non-IID setting. Figures \ref{subfig:local_steps} and \ref{subfig:heterogeneity}: Test error increases with local steps and heterogeneity when $m/2$ filters are misaligned at initialization, remains constant when all the filters are aligned.
\label{fig:emp_verification}}
\end{figure*}

\subsection{Training Loss Convergence and Test Error Guarantee}

Next, we state our main result regarding the convergence of \ts{FedAvg} with random initialization. We assume the CNN weights are initialized as $\bw_{j,r}^{(0)} \sim \mathcal{N}(\mathbf{0}, \sigma_0^2\mathbf{I}_d)$ for all filters, where $\mathbf{I}_d$ is the $(d \times d)$ identity matrix. We first state the following standard conditions used in our analysis.

% \begin{condition}
\begin{condition} Let $\epsilon$ be a desired training error threshold and $\delta \in (0,1)$ be some failure probability.\footnote{\scriptsize{We use $\lesssim$ and $\gtrsim$ to denote inequalities that hide constants and logarithmic factors. See Appendix for exact conditions.}}
\begin{enumerate}[leftmargin=*, label=(C\arabic*)]
\itemsep-0.3em 
\item \label{assump_max_t} 
The allowed number of communication rounds $t$ is bounded by $T^* = \frac{1}{\eta} \mathrm{poly}(\epsilon^{-1},m,n,d)$.

\item \label{assump_d}
    Dimension $d$ is sufficiently large:
        $d \gtrsim \max \left\{ \tfrac{n\norm{\bmu}}{\sigma_p^{2}}, n^2 \right\}$.
\item \label{assump_m_n}
    Training set size $n$ and neural network width $m$ satisfy: $m \gtrsim \log (n/\delta), n \gtrsim \log (m/\delta)$.
\item \label{assump_sigma_0}
   Standard deviation of Gaussian initialization is sufficiently small: 
    $\sigma_0 \lesssim \min \left\{ \tfrac{\sqrt{n}}{\sigma_pd \tau}, \tfrac{1}{\normt{\bmu}} \right\}$.

\item \label{assump_norm}
    The norm of the signal satisfies: 
    $\norm{\bmu} \gtrsim \sigma_p^2$.
    
\item \label{assump_lr}
    Learning rate is sufficiently small:
    $\eta \lesssim \min \left\{ \tfrac{nm}{\sigma_{p}^{2}d}, \tfrac{1}{\normt{\bmu}^{2}}, \tfrac{1}{\sigma_p^2d} \right\}.$

\end{enumerate}
\label{assum:main_assump}
\end{condition}

The above conditions are standard and have also been made in \cite{cao2022benign, kou2023benign} for the purpose of theoretical analysis. \ref{assump_max_t} is a mild condition needed to ensure that the signal and noise coefficients remain bounded throughout the duration of training. Furthermore, we see in Theorem 1 that we only need $T = \bigO{mn \eta^{-1} \epsilon^{-1} d^{-1} \log(\tau/\epsilon)}$ rounds to reach a training error of $\epsilon$, which is well within the admissible number of rounds. \ref{assump_d} is used to bound the correlation between the noise vectors and also the correlation of the initial filter weights with the signal and noise. \ref{assump_m_n} is needed to ensure that a sufficient number of filters have non-zero activations at initialization so that the initial gradient is non-zero. \ref{assump_sigma_0} is needed to ensure that the initial weights of the CNN are not too large and that it has bounded loss for all datapoints.  \ref{assump_norm} is needed to ensure that signal learning is not too slow compared to noise memorization. Finally, a small enough learning rate in \ref{assump_lr} ensures that Local GD does not diverge. Additional discussion on these assumptions is provided in \Cref{sec:assumptions_and_results}. With this assumption we are now state our main results.

\begin{theorem}[Training Loss Convergence]
\label{thm:train_loss}
For any $\epsilon > 0$ under \Cref{assum:main_assump},  there exists a $T = \bigO{\frac{mn}{\eta\sigma_p^{2}d\tau}} + \bigO{\frac{mn \log(\tau/\epsilon)}{\eta\sigma_p^{2}d\epsilon} }$ such that \ts{FedAvg} satisfies $L(\bW^{(T)}) \leq \epsilon$ with probability $\geq 1-\delta$.
\end{theorem}

Our training error convergence consists of two stages. In the first stage consisting of $T_1 := \bigO{\tfrac{mn}{\eta\sigma_p^{2}d\tau}}$ rounds, we show that the magnitudes of the cross-entropy loss derivatives are lower bounded by a constant, i.e., $|\ell'(y_{k,i}f(\bW_{k}^{(t,s)},\bx_{k,i}))| = \bigw{1}$. Using this we can show that the signal and noise coefficients $\{\ggam^{(t)}, \gprho^{(t)}\}$ grow linearly and are $\bigtheta{1}$ by the end of this stage.
Consequently, by the end of the first stage, the model reaches a neighborhood of a global minimizer where the loss landscape is nearly convex. Then in the second stage, we can establish that the training error consistently decreases to an arbitrary error $\epsilon$ in $\bigO{\tfrac{mn \log(\tau/\epsilon)}{\eta\sigma_p^{2}d\epsilon} }$ rounds.

Note that our analysis does not require the condition $\eta \propto \nicefrac{1}{\tau}$ as is common in many works analyzing \ts{FedAvg}. Therefore, by setting $\tau$ large enough we can make the number of rounds in the first stage as small as $\bigO{1}$, thereby reducing the communication cost of FL. However, in the second stage we do not see any continued benefit of local steps; in fact the number of rounds required grows as $\log (\tau)$. This suggests an optimal strategy would be to adapt $\tau$ throughout training: start with large $\tau$ and decrease $\tau$ after some rounds, which has also been found to work well empirically \cite{wang2019adaptive}.

\begin{theorem}[Test Error Bound]
\label{thm:test_error}
Define signal-to-noise ratio $\snr := \nicefrac{\normt{\bmu}}{\sigma_p \sqrt{d}}$ and $A_j := \{r \in [m]: \inner{\bw_{j,r}^{(0)}}{j\bmu} \geq 0\}$ to be the set of aligned filters (\Cref{def:signal_alignment}) corresponding to label $j$. Then under the same conditions as \Cref{thm:train_loss}, our trained CNN achieves
\begin{enumerate}[leftmargin=*]
    \item When $ \snr^2 \lesssim \nicefrac{1}{\sqrt{nd}}$, test error $L_{\mathcal{D}}^{0-1}(\bW^{(T)}) \geq 0.1$.
    \item When $ \snr^2 \gtrsim \nicefrac{1}{\sqrt{nd}}$, test error
    \begin{align}
        L_{\mathcal{D}}^{0-1}(\bW^{(T)}) &\leq \mfrac{1}{2} \textstyle\sum_{j \in \{\pm 1\}} \exp\Big(- \mfrac{n}{d}\Big[ \tfrac{\abs{A_j}}{m} \snr^2 \nn\\
        & + \big(1 - \tfrac{\abs{A_j}}{m} \big) \snr^2 \left(h+\tfrac{1}{\tau}(1-h)\right) \Big]^2 \Big). \nn
    \end{align}
\end{enumerate}
\end{theorem}

\paragraph{Impact of SNR on harmful/benign overfitting.}
Intuitively, if the $\snr$ is too low ($\snr^2 \lesssim \nicefrac{1}{\sqrt{nd}}$), then there is simply not enough signal strength for the model to learn compared to the noise. Hence, we cannot expect the model to generalize well no matter how we train it. This generalizes the centralized training result in \citep[Theorem~4.2]{kou2023benign} (with $p=0$), which corresponds to $\tau = 1$ in \ts{FedAvg}. In this case, the model is in the regime of \textit{harmful overfitting}. 
However, if the $\snr$ is sufficiently large ($\snr^2 \gtrsim \nicefrac{1}{\sqrt{nd}}$), we enter the regime of \textit{benign overfitting}, where the model can fit the data and generalize well with the test error reducing exponentially with the global dataset size $n$. 

\begin{figure*}[t]
  \centering
  \subfloat[IID Signal Learning]{\includegraphics[width=0.3\linewidth]{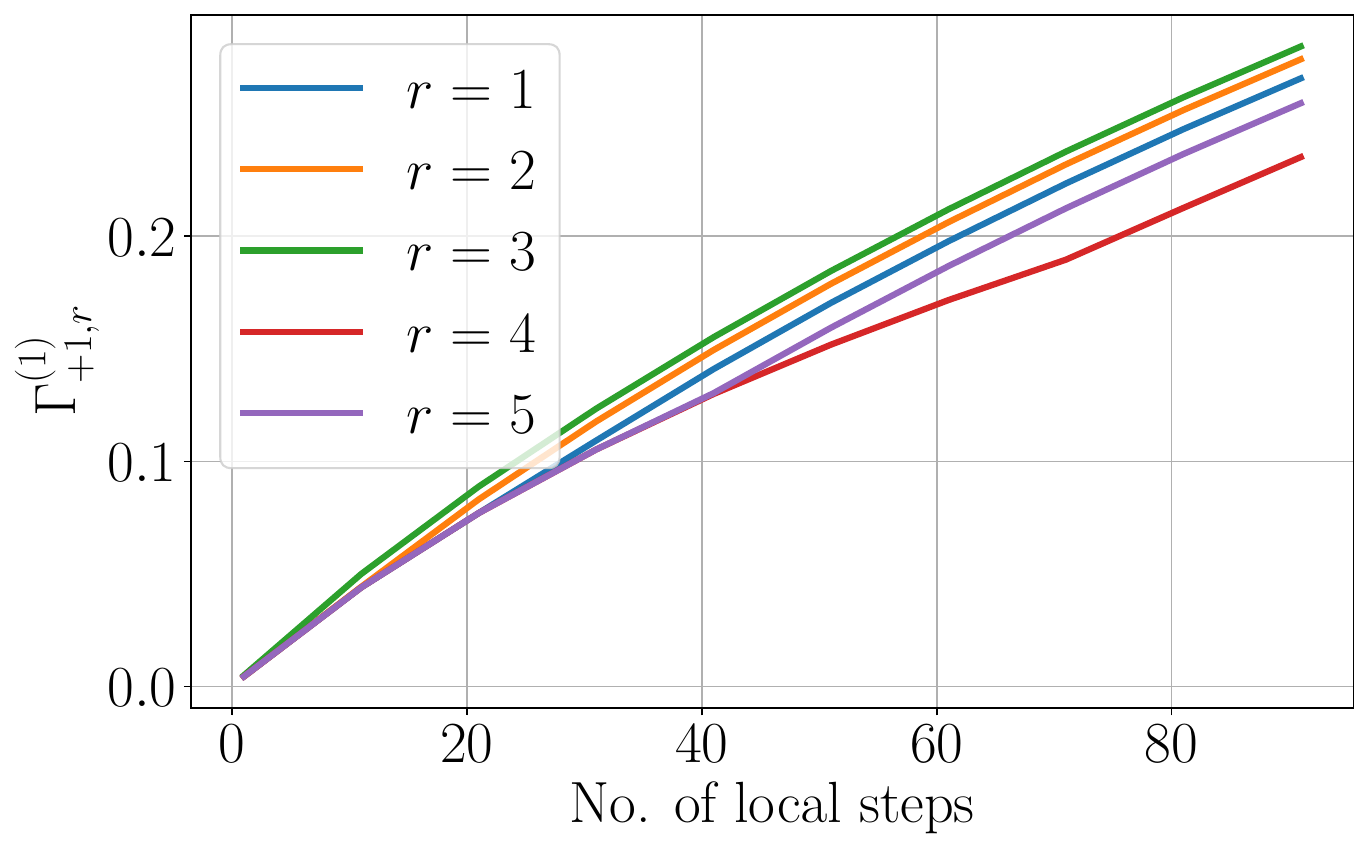}
  \label{subfig:signal_iid}}
   \subfloat[IID Noise Memorization]{\includegraphics[width=0.3\linewidth]{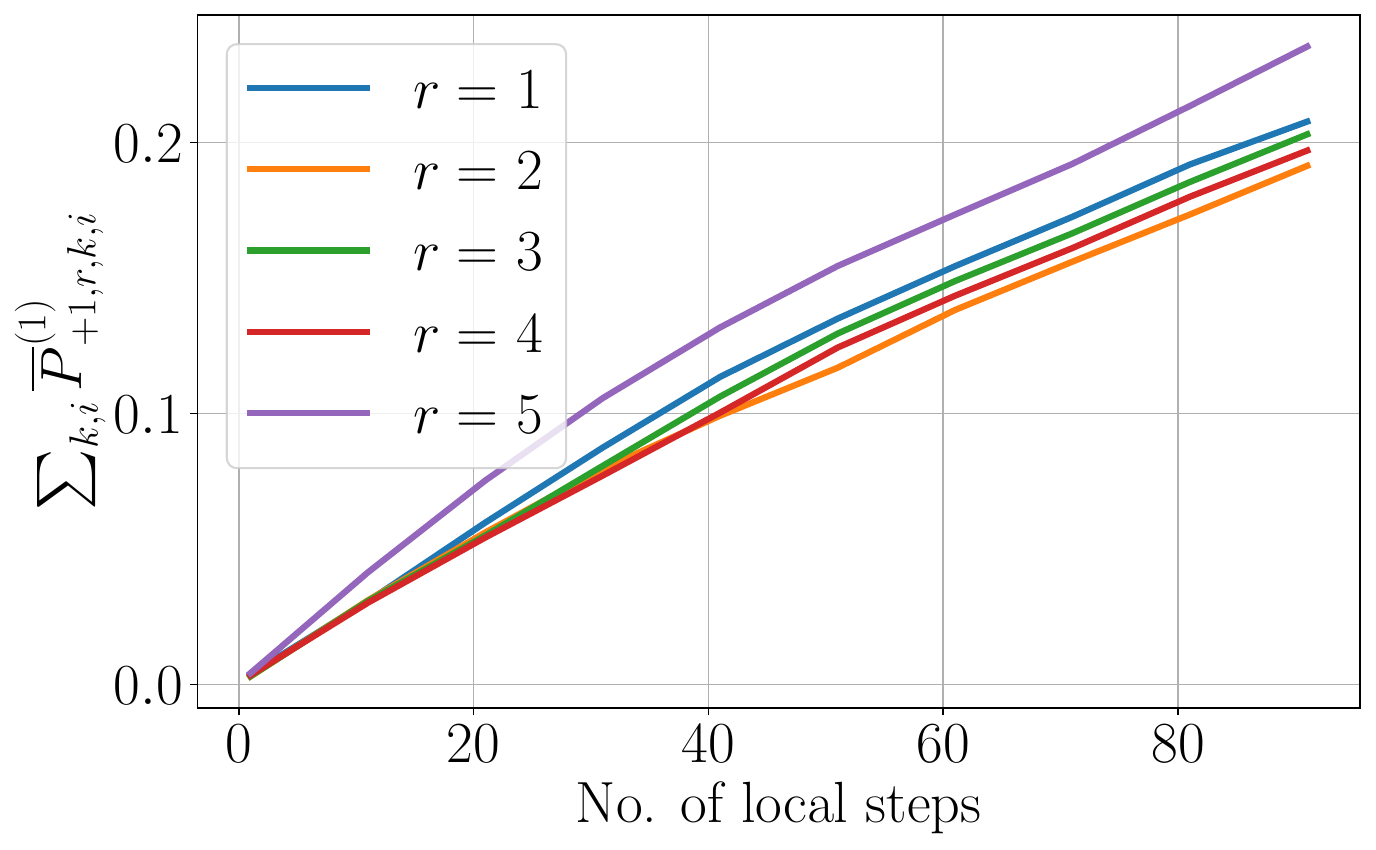}
   \label{subfig:noise_iid}}
   \subfloat[IID Sig. Learning/Noise Mem.]{\includegraphics[width=0.3\linewidth]{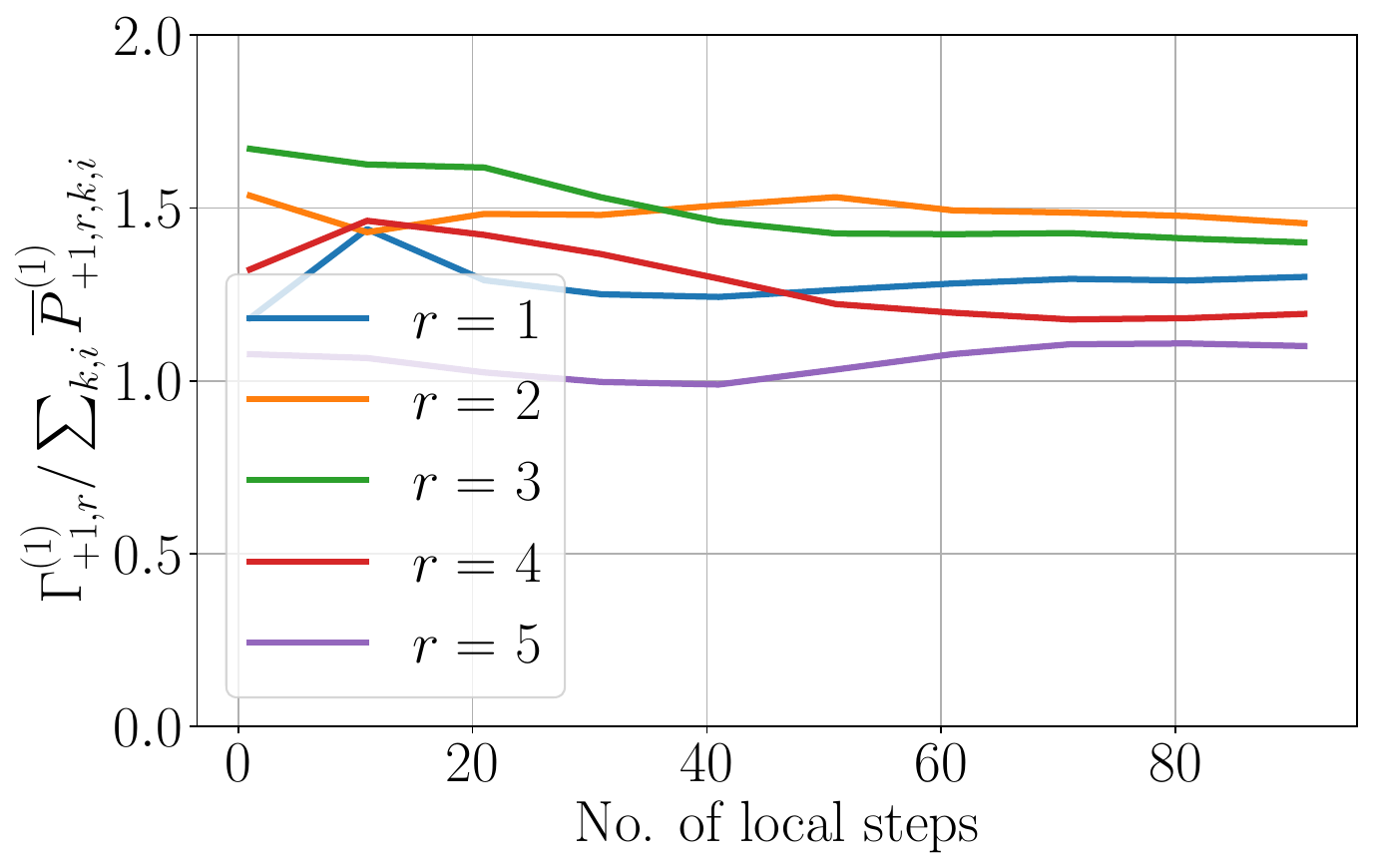}
    \label{subfig:snr_iid}}
    \\
    \subfloat[NonIID Signal Learning]{\includegraphics[width=0.3\linewidth]{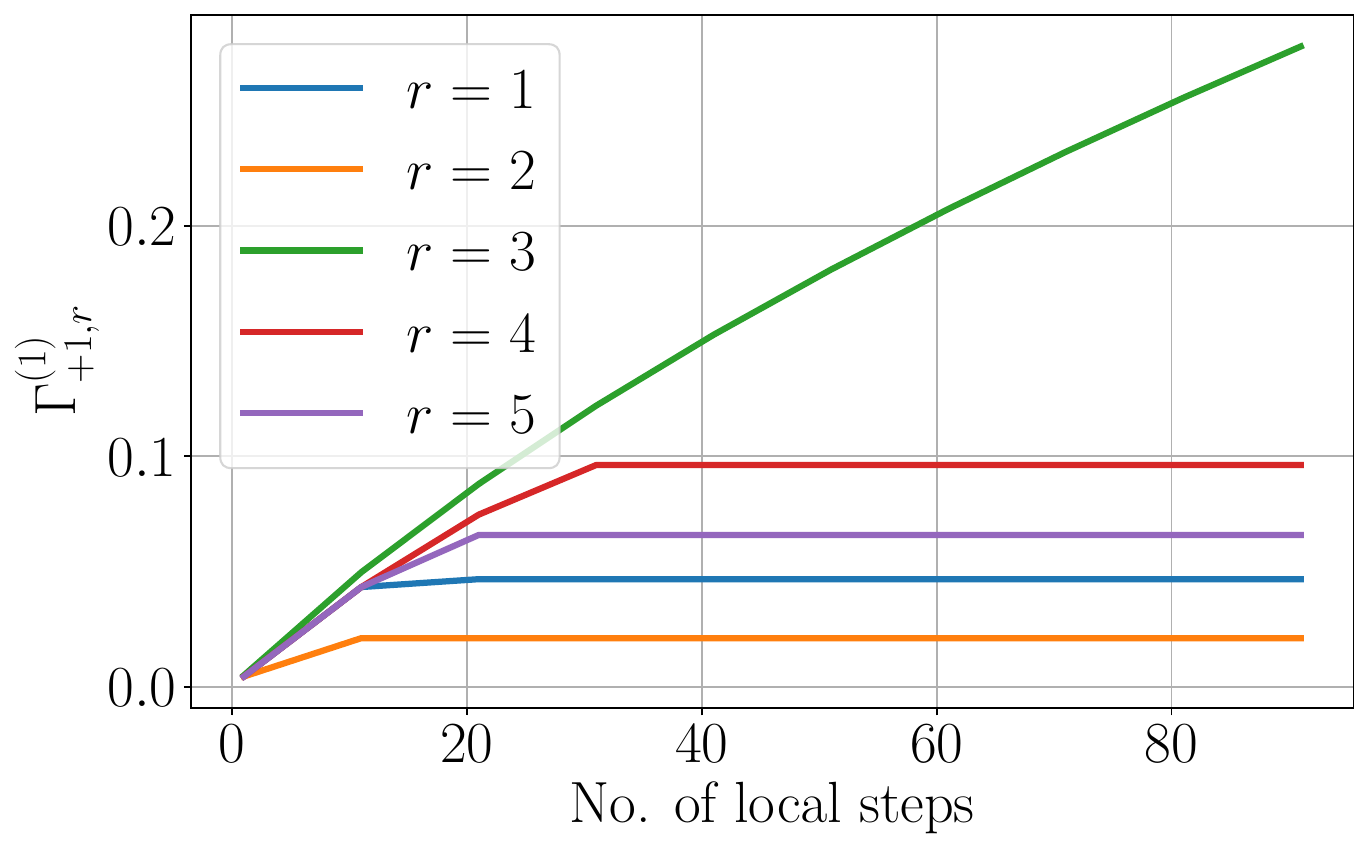}
  \label{subfig:signal_noniid}}
   \subfloat[NonIID Noise Memorization]{\includegraphics[width=0.3\linewidth]{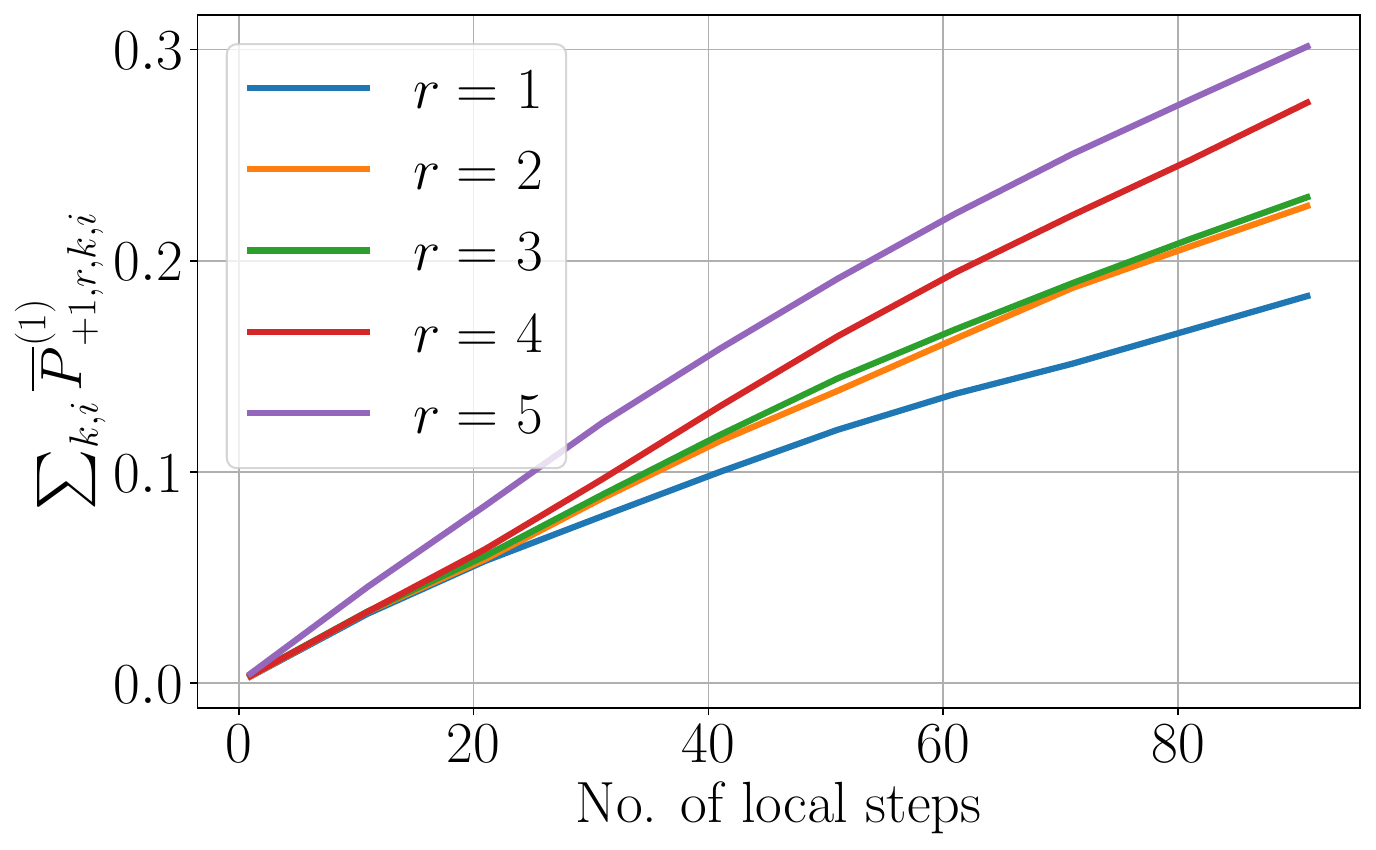}
   \label{subfig:noise_noniid}}
   \subfloat[NonIID Sig. Learning/Noise Mem.]{\includegraphics[width=0.3\linewidth]{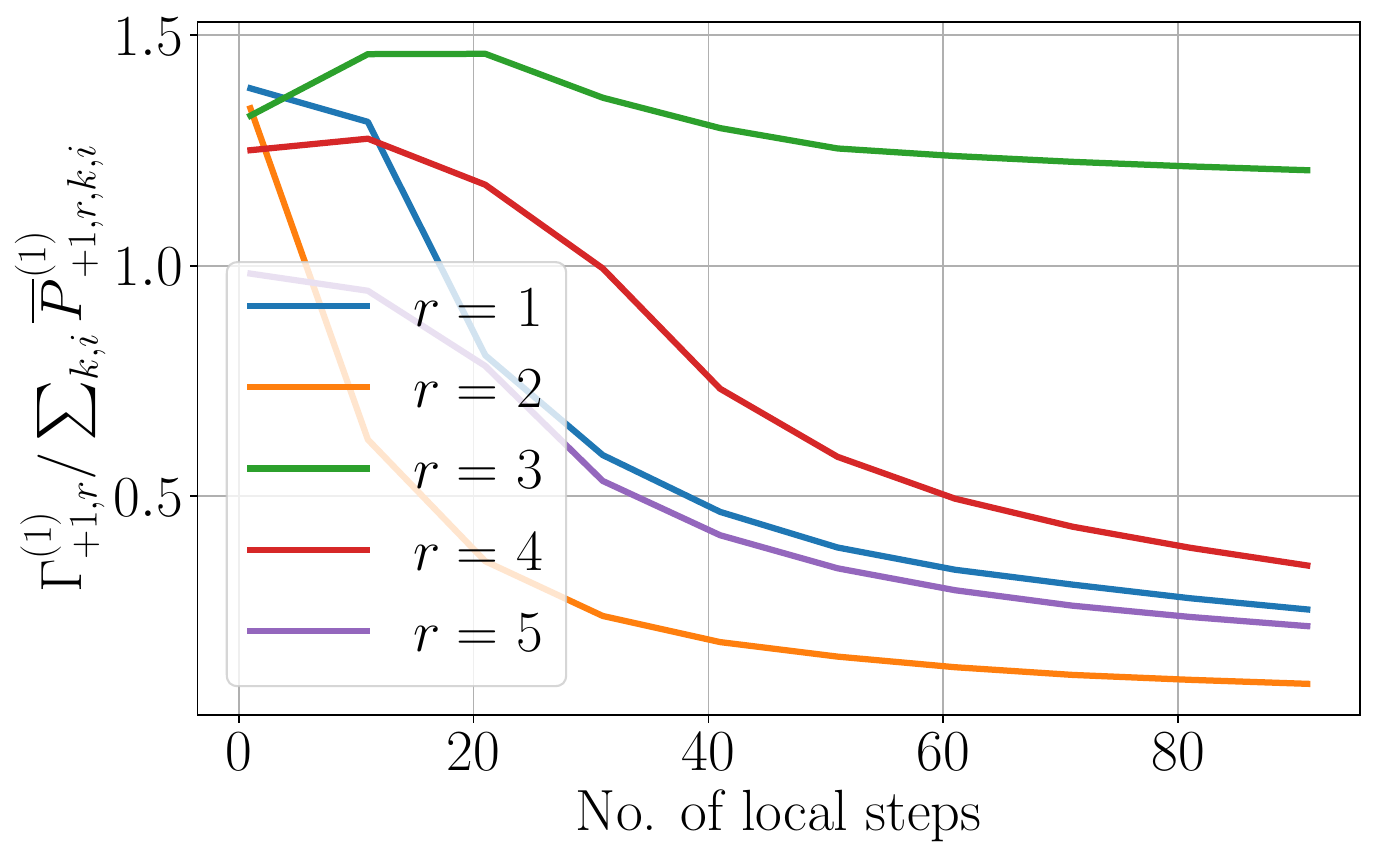}
    \label{subfig:snr_noniid}}
\caption{Signal learning and noise memorization for our CNN model in the IID $(h = \nicefrac{1}{2})$ and NonIID $(h = 0)$ setting after $1$ round. Figures \ref{subfig:signal_iid}, \ref{subfig:signal_noniid}: 
In the IID setting signal learning coefficients are similar for all the filters and increase with the number of local steps $\tau$ but in the NonIID setting they saturate (\Cref{lem:main_paper_lemma_signal_growth}) for misaligned filters ($r = 1,2,4,5$). Figures 
\ref{subfig:noise_iid}, \ref{subfig:noise_noniid}: Noise memorization is similar for all filters in both settings and grows with $\tau$ \Cref{lem:main_paper_noise_growth}. Figures \ref{subfig:snr_iid}, \ref{subfig:snr_noniid}: in the IID setting, the ratio of signal learning to noise memorization remains independent of $\tau$. But in the NonIID setting, the ratio decreases to zero as $\tau$ increases for misaligned filters ($r = 1,2,4,5$).
\label{fig:two_layer_cnn_expts}}
\vspace{-1em}
\end{figure*}

\paragraph{Empirical Verification.} 
We now provide empirical verification of the upper bound on the test error in \Cref{thm:test_error} in the benign overfitting regime. We simulate a synthetic dataset following our data-generation model in \Cref{sec:problem_setup}, with $n = 20$ datapoints, $K = 2$ clients and $m = 10$ filters. Additional experimental details can be found in \Cref{sec:expt_details}. We fix a training error threshold of $\epsilon = 0.1$ and then measure the test error of our CNN under various settings in \Cref{fig:emp_verification}. 
\Cref{subfig:filter_alignment} shows the test error as a function of the number of misaligned filters ($m-|A_j|$ in \Cref{thm:test_error}) under different data partitionings with the number of local steps fixed at $\tau = 100$. While the test error grows with the number of misaligned filters in both data settings, the rate of growth is much larger in the non-IID setting.
\Cref{subfig:local_steps} shows the test error as a function of local steps $\tau$ under different initializations for fixed $h =0$ while \Cref{subfig:heterogeneity} shows the test error as a function of heterogeneity under different initializations for fixed $\tau = 100$. As predicted by our theory, heterogeneity and the number of local steps do not affect test error when all the filters are aligned at initialization. On the other hand, the test error grows with $\tau$ and heterogeneity when the number of misaligned filters is non-zero ($m/2 = 5$) for each $j \in \{\pm 1\}$. Therefore, our empirical results strongly validate our theoretical results showing the effect of heterogeneity, number of local steps and number of misaligned filters on the test error.

\subsection{Impact of Filter Alignment and Data Heterogeneity on Signal Learning and Noise Memorization.}
\label{subsec:signal_noise_growth}
The key results in our analysis are the following lemmas which bound the growth of the signal learning and noise coefficient during the first stage of training, that is $0 \leq t \leq T_1$ (see discussion under \Cref{thm:train_loss}). Using our definition of $A_j := \{r \in [m]: \inner{\bw_{j,r}^{(0)}}{j\bmu} \geq 0\}$ as the set of aligned filters, we have the following lemma for growth of the signal learning coefficient in the first stage.
\begin{lemma}
\label{lem:main_paper_lemma_signal_growth}
    Under \Cref{assum:main_assump}, for all $0 \leq t \leq T_1$, we have $\ggam^{(t)} =  \bigw{\frac{t \eta \norm{\bmu} \tau}{m}}$ if $r \in A_j$ and $\ggam^{(t)}  = \bigw{\frac{t\eta \norm{\bmu}(1 + h(\tau-1))}{m}}$ if $r \notin A_j$.
\end{lemma}

This lemma shows that \textit{for aligned filters ($r \in A_j$), $\ggam^{(t)}$ does not depend on heterogeneity and grows linearly with the number of local steps $\tau$}. On the other hand, \textit{for misaligned filters $(r \notin A_j)$, the growth depends on the heterogeneity parameter $h$}. Furthermore, under extreme data heterogeneity ($h = 0$), for misaligned filters  $\ggam^{(t)}$ \textit{does not scale with the number of local steps $\tau$}. For the growth of noise coefficients we have the following corresponding lemma,
\begin{lemma}
\label{lem:main_paper_noise_growth}
    Under \Cref{assum:main_assump}, for all $0 \leq t \leq T_1$ we have $ \sum_{k,i}  \gprho^{(t)} = \bigtheta{{\mfrac{ t\eta \tau \sigma_p^2d}{m}}}.$
\end{lemma}
This lemma shows that \textit{noise memorization does not depend on data-heterogeneity or filter alignment and always scales linearly with the number of local steps $\tau$}. Intuitively, this can be expected because the noise vectors are independent of the label information $y$ in a datapoint following our data generation model in \Cref{sec:problem_setup} and for any given filter we can show there are $\bigw{N}$ noise vectors that are aligned with the filter at initialization for every client with high probability (see \Cref{lemma:min_size_activated_noise_filters}).

Using the above two lemmas, we have the following bound on the ratio of signal learning to noise memorization for filter $(j, r)$ at the end of the first stage of training
\begin{align}
    \mfrac{\ggam^{(T_1)}}{\sum_{k,i} \gprho^{(T_1)}} \geq
    \begin{cases}
    \snr^2, \text{ if } r \in A_j, \\ 
    \snr^2(h + \frac{1}{\tau}(1-h)), \text{ if } r \in [m] \setminus A_j.
    \end{cases}
    \label{eq:snr_lower_bound}
\end{align}

This ratio is key to bounding the generalization performance of the CNN model as we show later in the proof of \Cref{thm:test_error} in \Cref{subsec:test_error_upper_bound_proof}. For aligned filters ($r \in A_j$), the ratio is unaffected by data heterogeneity $h$ and the number of local steps $\tau$. However, for misaligned filters ($r \in [m] \setminus A_j$), the ratio becomes smaller as heterogeneity increases ($h$ becomes smaller) or $\tau$ increases. Thus, for misaligned filters we see a corresponding dependence on heterogeneity and local steps in our upper bound on test error in \Cref{thm:test_error}. Note that in centralized training with $\tau = 1$, we have $(h + \frac{1}{\tau} (1-h)) = 1$ and thus we do not see any impact of heterogeneity at misaligned filters.
Therefore, we recover the bound $L_{\mathcal{D}}^{0-1}(\bW^{(T)}) \leq \exp(- n \snr^2/d)$ in \citep[Theorem~4.2]{kou2023benign}.
\textit{It is only in FL training with $\tau > 1$ local steps that we encounter the adverse effect of data heterogeneity at the misaligned filters.}

\paragraph{Empirical Verification.} We empirically verify the results above in the IID $(h = 1/2)$ and Non-IID $(h = 0)$ setting following the same simulation setup as done in \Cref{fig:emp_verification}. Figure \Cref{subfig:signal_iid} shows that in the IID setting signal learning coefficients are similar for all the filters and increases with the number of local step. However, as shown by \Cref{subfig:signal_noniid}, in the NonIID setting signal learning saturates 
for misaligned filters. 
Figures \ref{subfig:noise_iid} and \ref{subfig:noise_noniid} show that the growth of noise coefficients for all the filters is similar in the IID and non-IID case. 
\captionsetup{font=scriptsize}
\begin{wrapfigure}[8]{r}{0.2\textwidth}
\scriptsize

\centering 
\includegraphics[width=0.2\textwidth]{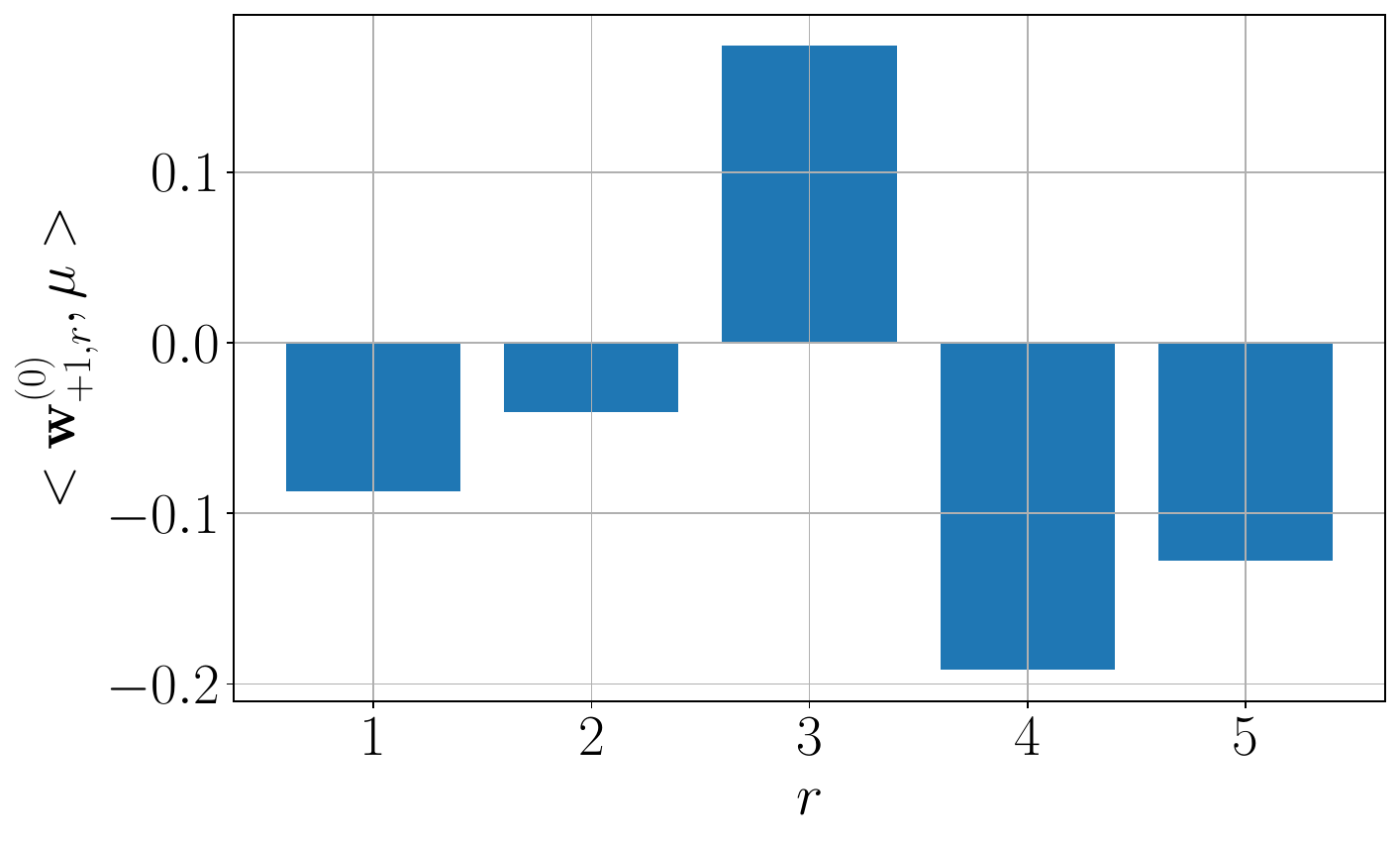}\vspace{-0.7em}\caption{Initial alignment of the filters in \Cref{fig:two_layer_cnn_expts}.}
\label{fig:init_filt_alignment}
\end{wrapfigure}
\captionsetup{font=small}In \Cref{subfig:snr_iid} we see that ratio of signal learning to noise memorization is lower bounded by a constant for all filters in the IID setting whereas in the Non-IID setting it decays as $\tau$ increases for misaligned filters (\Cref{subfig:snr_noniid}), thus verifying our theoretical analysis.

\subsection{Impact of Pre-Training on Federated Learning} \label{subsec:unraveling_pretrain}

Given the result in \Cref{thm:test_error}, we return to our question in \Cref{sec:intro}, about the \textit{effect of pre-trained initialization on improving generalization performance in FL}. We focus on centralized pre-training but our discussion here can be extended to federated pre-training as well (see \Cref{lemma:filter_aligned_after_T1} which states a federated counterpart of the lemma below). 

Suppose we pre-train a CNN model in a centralized manner on a dataset with signal $\bmu^{(\text{pre})}$ generated according to the data model described in \Cref{sec:problem_setup}.  Now if we train for sufficient number of iterations, then we can show that \textit{all} filters will be correctly aligned with the pre-training signal. 

\begin{lemma}[All Filters Aligned After Sufficient Training]
\label{lemma:pretrain}
There exists $T_1 = \bigO{\frac{mn}{\eta\sigma_p^{2}d}}$ such that for all $t \geq T_{1}, j \in \{ \pm 1\}, r \in [m]$ we have $\inner{\bw^{(\text{pre},t)}_{j,r}}{j\bmu^{(\text{pre})}}
\geq 0$. 
\end{lemma}

Now suppose we pre-train for $t \geq T_1$ iterations to get a model $\bW^{(\text{pre}, *)}$ and use this model to initialize for downstream federated training (i.e., $\bW^{(0)} = \bW^{(\text{pre, *})}$) with signal vector $\bmu$. Then for all $j,r$ filters, we have
$ \inner{\bw_{j,r}^{(0)}}{j\bmu} = \inner{\bw^{(\text{pre,*})}_{j,r}}{j\bmu^{(\text{pre})}} + \inner{\bw^{(\text{pre,*})}_{j,r}}{j(\bmu - \bmu^{(\text{pre})})}$. We also know that $\inner{\bw^{(\text{pre,*})}_{j,r}}{j\bmu^{(\text{pre})}} \geq 0$ using \Cref{lemma:pretrain}. 
Therefore, if $\|\bmu - \bmu^{(\text{pre})}\|_2$ is small,
% $\inner{\bw_{j,r}^{(0)}}{j\bmu} \geq 0$ for all $j,r$, and 
all the filters $\{\bw_{j,r}^{(0)}\}$ are correctly aligned with the signal $j \bmu$. As a result, in \Cref{thm:test_error} $A_j = [m]$ for $j \in \{ \pm 1 \}$ and in the benign overfitting regime ($\snr^2 \gtrsim \nicefrac{1}{\sqrt{nd}}$), we recover the centralized result $L_{\mathcal{D}}^{0-1}(\bW^{(T)}) \leq \exp(- n \snr^2/d)$ \citep[Theorem~4.2]{kou2023benign}. Hence, the adverse effects of cross-client heterogeneity are mitigated with pre-trained initialization. 

\section{Experiments}
\label{sec:expts}
In this section we provide empirical results showing how our insights from \Cref{sec:main_results} extend to practical FL training on real world datasets with deep CNN models. Unless specified otherwise, we use the ResNet18 model \cite{DBLP:conf/cvpr/HeZRS16} in all our experiments and split the data across $20$ clients using the Dirichlet sampling scheme \cite{hsu2019noniid} with non-iid parameter $\alpha = 0.3$. For pre-training, we use a ResNet18 pre-trained on ImageNet \cite{russakovsky2015imagenet}, available in PyTorch \cite{paszke2019pytorch}. Additional experimental details can be found in \Cref{sec:expt_details}.

\begin{figure}[h]
  \centering
  \subfloat[]
  {\raisebox{0.5em}{\includegraphics[width=0.49\linewidth]{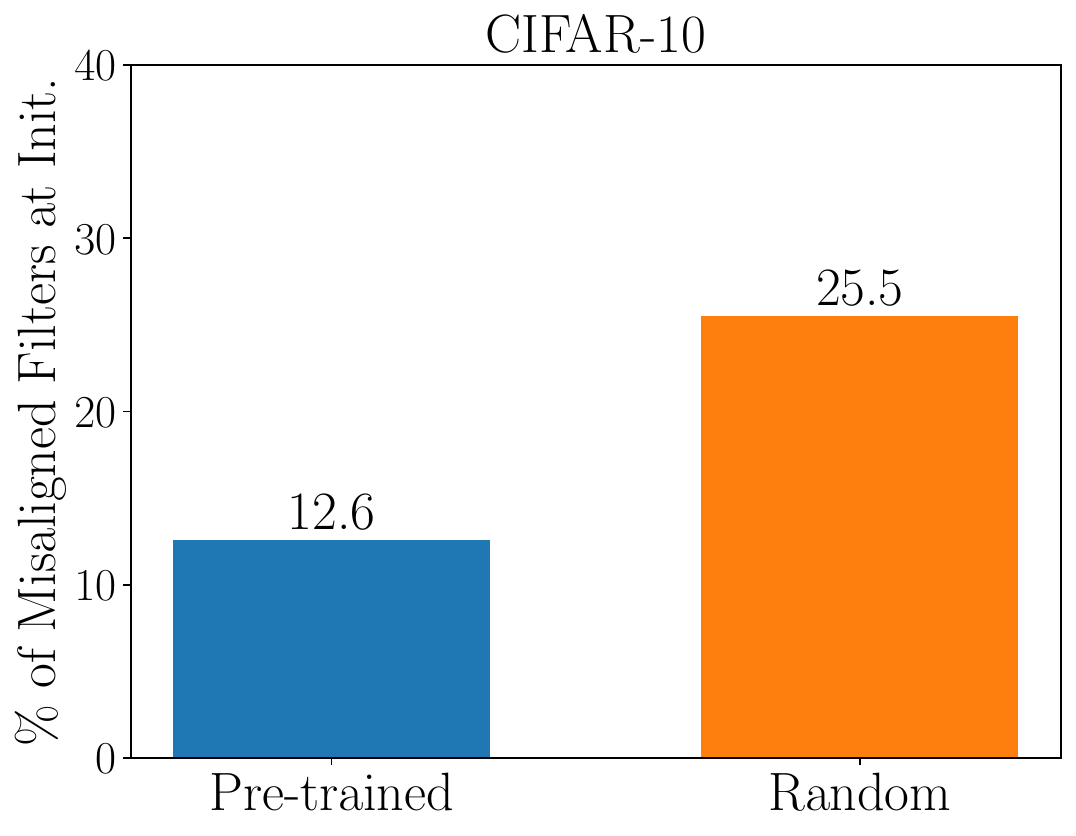}}
  \label{subfig:filter_misalign_cifar10_0.3}}
  \subfloat[]
  {\includegraphics[width=0.49\linewidth]{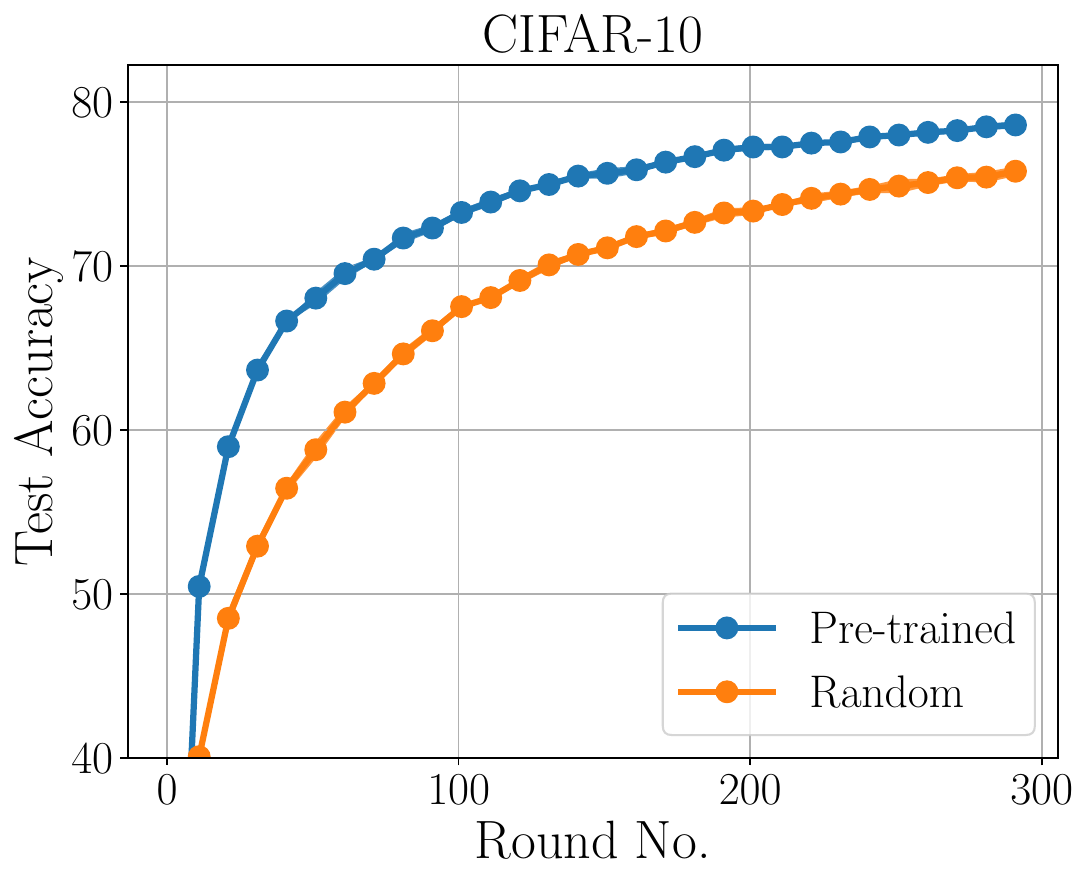}
  \label{subfig:test_acc_cifar10_0.3}}
    \\
    \subfloat[]{\raisebox{0.5em}{\includegraphics[width=0.49\linewidth]{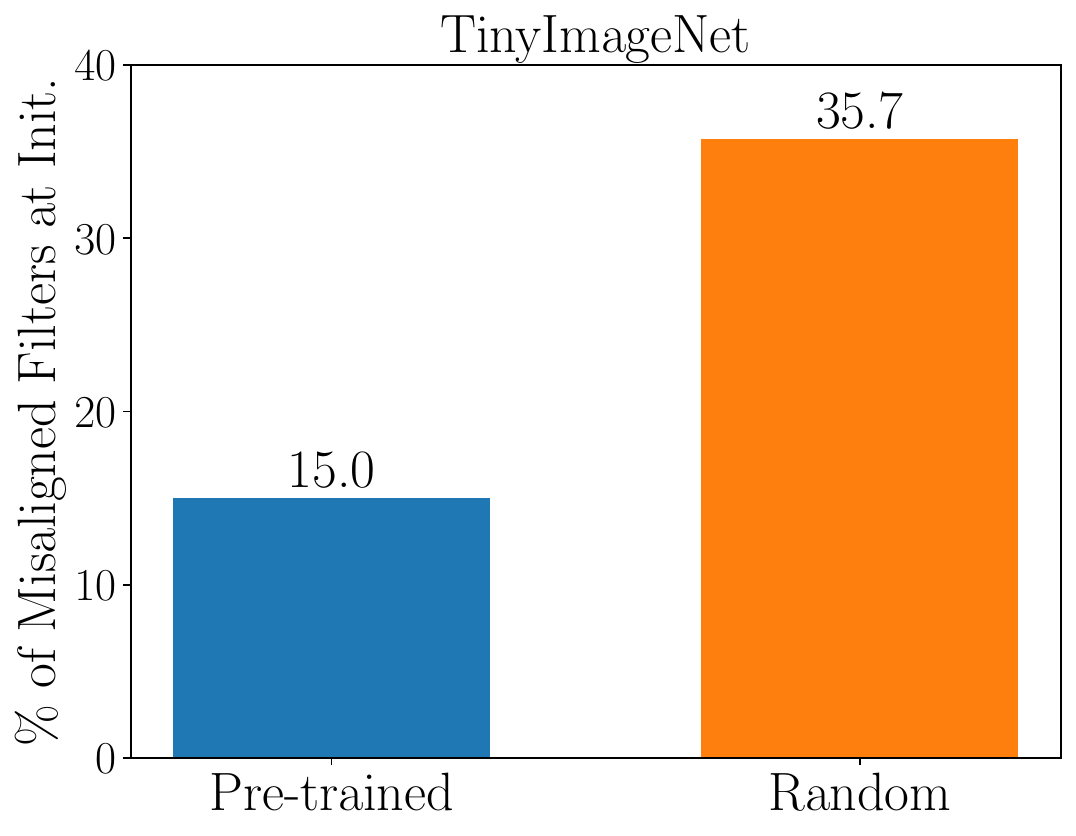}}
\label{subfig:filter_misalign_tinyimagenet}}
     \subfloat[]{\includegraphics[width=0.49\linewidth]{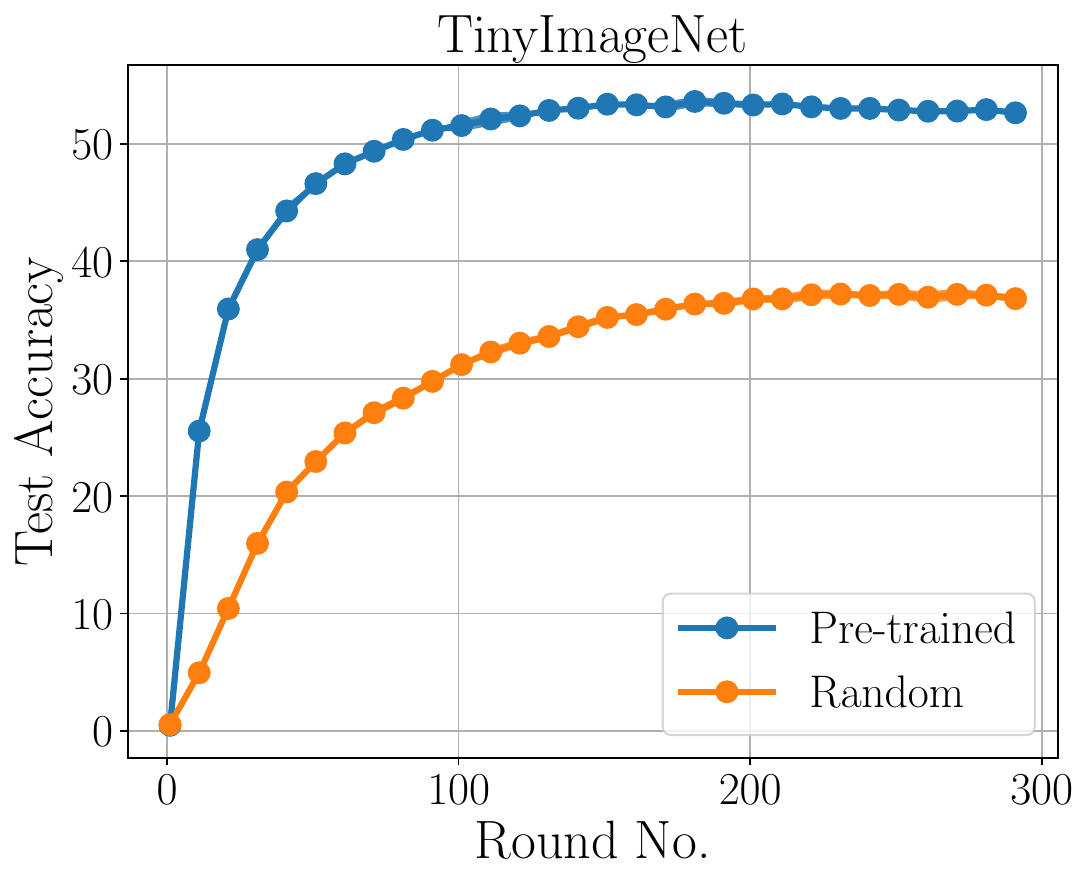}
    \label{subfig:test_accuracy_tinyimageent}}
\caption{The percentage of misaligned filters (see \Cref{eq:empirical_alignment} and test accuracy for different initializations on CIFAR-10 (\Cref{subfig:filter_misalign_cifar10_0.3} and \Cref{subfig:test_acc_cifar10_0.3}) and TinyImageNet (\Cref{subfig:filter_misalign_tinyimagenet} and \Cref{subfig:test_accuracy_tinyimageent}). As the complexity of the signal information in the data grows from CIFAR-10 to TinyImageNet, we see a sharp increase in the ratio of misaligned filters for random initialization, explaining why pre-trained initialization offers larger improvements for TinyImageNet.}
\label{fig:real_world_misalignment}
\end{figure}

\begin{figure}[h]
  \centering
  \subfloat[]{
  \raisebox{0.5em}{\includegraphics[width=0.49\linewidth]{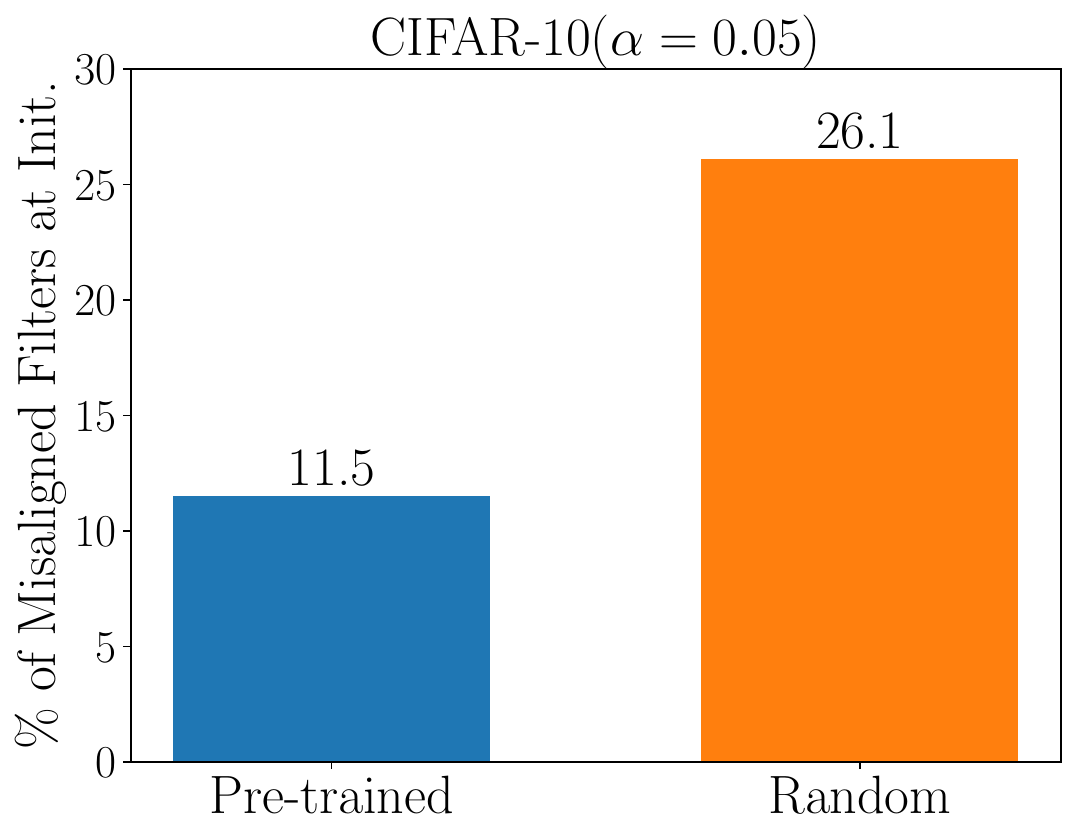}}
  \label{subfig:filter_misalign_cifar10_0.05}}
   \subfloat[]{\includegraphics[width=0.49\linewidth]{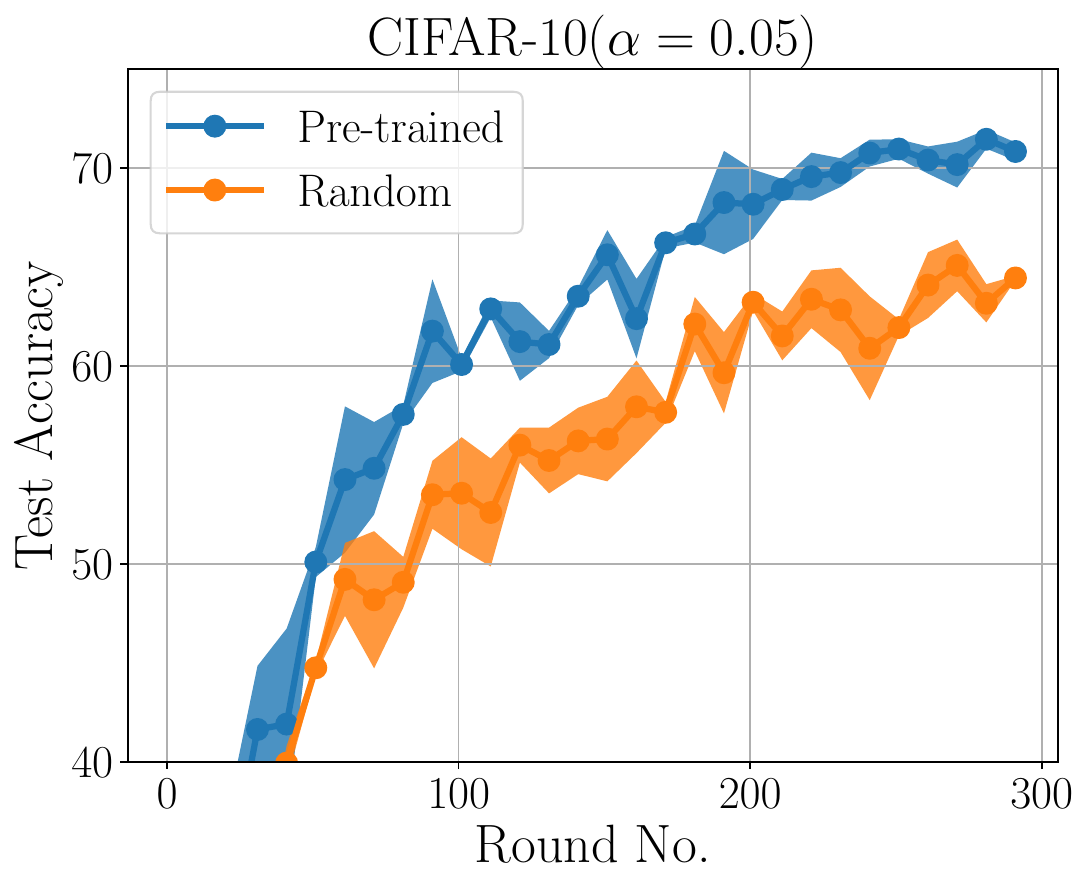}
   \label{subfig:test_acc_cifar10_0.05}}
    \\
    \subfloat[]{\raisebox{0.5em}{\includegraphics[width=0.49\linewidth]{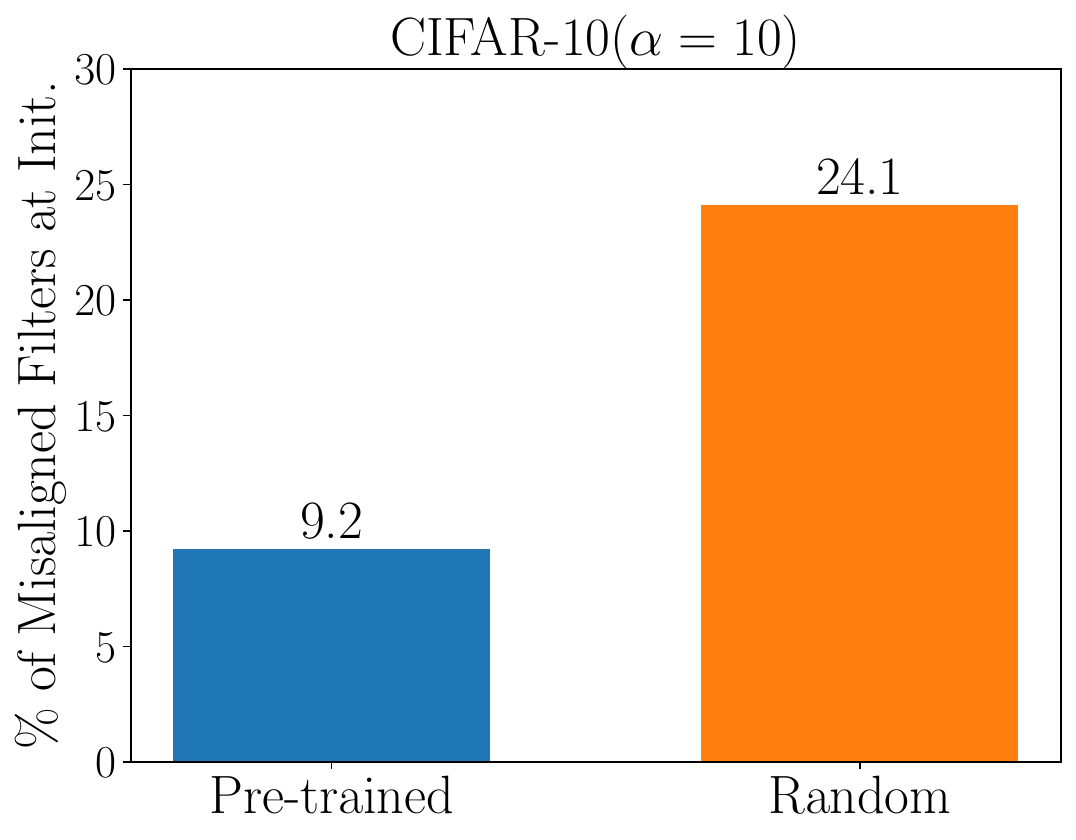}}
  \label{subfig:filter_misalign_cifar10_10}}
   \subfloat[]{\includegraphics[width=0.49\linewidth]{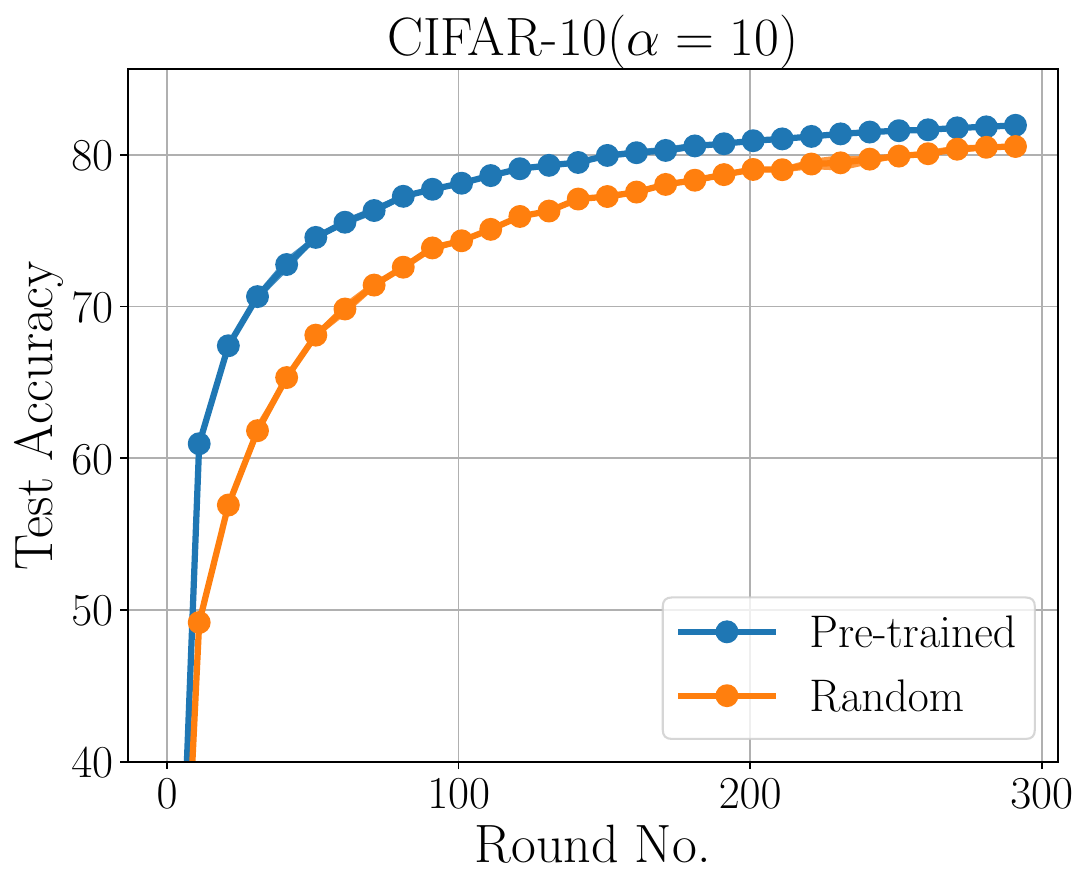}
   \label{subfig:test_acc_cifar10_10}}
\caption{The percentage of misaligned filters (see \Cref{eq:empirical_alignment}) and test accuracy for different initializations on CIFAR-10  with $\alpha = 0.05$ heterogeneity (\Cref{subfig:filter_misalign_cifar10_0.05} and \Cref{subfig:test_acc_cifar10_0.05}) and $\alpha = 10$ heterogeneity (\Cref{subfig:filter_misalign_cifar10_10} and \Cref{subfig:test_acc_cifar10_10}). Although the percentage of misaligned filters does not vary significantly across the two settings for both initializations (signal information is the same in both settings), pre-training offers more improvement in the higher heterogeneity setting ($\alpha = 0.05)$, as suggested by our theoretical analysis.} 
\label{fig:varying_heterogeneity}
\end{figure}

\paragraph{Empirical Measure of Misalignment.} Measuring filter alignment for deep CNNs is challenging since we cannot explicitly characterize the signal information present in real world datasets and furthermore different layers will learn the signal at different levels of granularity. Nonetheless, our theoretical findings suggest that given sufficient number of training rounds, filters will be aligned with the signal (see \Cref{sec:main_results}) and once a filter is aligned, the sign of the output produced by the filter with respect to the signal does not change, i.e, if $\langle \bw_{j,r}^{(t)}, j \bmu \rangle > 0$ then $\mathrm{sign}(\langle \bw_{j,r}^{(t')}, \bmu \rangle) = \mathrm{sign}(\langle \bw_{j,r}^{(t)}, \bmu \rangle)$, for all $t' \geq t$. Therefore, we propose to use the sign of the output produced by a filter at the end of training as a reference for alignment at any given round. Formally, let $\bW^{(0)}, \bW^{(1)} \cdots \bW^{(T)}$ be the sequence of iterates produced by federated training and let $\mathcal{F}(\bw, \bx) = [\langle \bw, \bx(1) \rangle, \langle \bw, \bx(2) \rangle, \dots \langle \bw, \bx(p) \rangle] \in \mathbb{R}^{p}$ be the feature map vector generated by filter $\bw$ for input $\bx$. For a given batch of data $\mathcal{B}$, we define the empirical measure of alignment of filter $\bw^{(t)}$ relative to $\bw^{(T)}$ as follows:
\begin{align}
\label{eq:empirical_alignment}
    \mathcal{A}(\bw^{(t)}) := \sum_{x \in \mathcal{B}, l \in [p]} \mathrm{sign}(\mathcal{F}_l(\bw^{(t)}, \bx))\mathrm{sign}(\mathcal{F}_l(\bw^{(T)}, \bx)).
\end{align} 

We say that the weight $\bw^{(t)}$ at round $t$ is misaligned if $\mathcal{A}(\bw^{(t)}) < 0$, because this implies that the sign of the output produced by the filter $\bw$ at round $t$ eventually changed for a majority of the inputs, hence indicating that the filter was misaligned at round $t$. We compute this measure over a batch of data to account for signal information coming from different classes of data as well as reduce the impact of noise in the data.

\paragraph{Measuring Misalignment on Real World Datasets with Varying Signal Information.} In this experiment our goal is to empirically demonstrate that (a) pre-trained initialization leads to much fewer number of misaligned filters than random initialization and (b) the number of misaligned filters for random initialization increases as we increase the complexity of the signal. To demonstrate this, we consider federated training on the \textbf{1. CIFAR-10} \cite{krizhevsky2009learning} and \textbf{2. TinyImageNet} \cite{le2015tiny} datasets. \Cref{fig:real_world_misalignment} shows the test accuracy and percentage of misaligned filter across training rounds for both datasets with pre-trained and random initialization. Firstly, we see that the percentage of misaligned filters is $2-3 \times$ smaller when starting from a pre-trained initialization compared to a random initialization. Furthermore, as the complexity of the signal information in the dataset increases (CIFAR-10 < TinyImageNet), we see a sharp increase in the percentage of misaligned filters ($25\%$ to $40\%$) for random initialization. In contrast, with pre-trained initialization, the percentage of misaligned filters remains less than $15\%$ across datasets leading to a larger improvement in test accuracy for TinyImageNet. These results align with our theoretical findings: as the ratio of misaligned filters increases, the benefits of pre-training become more pronounced.

\paragraph{Measuring Misalignment with Varying Heterogeneity Levels.} We extend the experiment in \Cref{fig:real_world_misalignment} conducted on CIFAR-10 with $\alpha = 0.3$ Dirichlet heterogeneity to other levels of heterogeneity \textbf{1. $\alpha = 0.05$} which is an extreme non-IID split and \textbf{2. $\alpha = 10$} which can be thought of as close to IID split. \Cref{fig:varying_heterogeneity} shows the test accuracy and percentage of misaligned filters plots for these two heterogeneity levels with pre-trained and random initialization. We observe that in both cases the percentage of misaligned filters remains approximately $25\%$ with random initialization and $10\%$ with pre-trained initialization, regardless of the level of heterogeneity. However, as heterogeneity increases, the improvement in test accuracy provided by pre-trained initialization becomes more pronounced. This trend is consistent with our theoretical analysis in Theorem 2, which suggests that the percentage of misaligned filters will have a greater impact on test performance as data heterogeneity increases.

\section{Conclusion and Future Work}

In this work we provide a deeper theoretical explanation for why pre-training can drastically reduce the adverse effects of non-IID data in FL by studying the class of two layer CNN models under a signal-noise data model. Our analysis shows that the reduction in test accuracy seen in non-IID FL compared to IID FL is only caused by filters that are misaligned at initialization. When starting from a pre-trained model we expect most of the filters to be already aligned with the signal thereby reducing the effect of heterogeneity and leading to a higher ratio of signal learning to noise memorization. This is corroborated by experiments on synthetic setup as well as more practical FL training tasks. Our work also opens up several avenues for future work. These including extending the analysis to deeper and more practical neural networks and also incorporating multi-class classification with more than two labels. Another interesting direction is to see how pre-training affects other federated algorithms such as those that explicitly incorporate heterogeneity reducing mechanisms. 

\section*{Acknowledgments}
This work was supported in part by NSF grants CCF 2045694, CNS-2112471, CPS-2111751, and SHF-2107024, ONR grant N00014-23-1-2149, and a Google Research Scholar Award.

% \section*{Impact Statement}

% This paper presents work whose goal is to advance the field of Machine Learning. There are many potential societal consequences of our work, none which we feel must be specifically highlighted here.

\bibliography{example_paper}

\begin{thebibliography}{83}
\providecommand{\natexlab}[1]{#1}
\providecommand{\url}[1]{\texttt{#1}}
\expandafter\ifx\csname urlstyle\endcsname\relax
  \providecommand{\doi}[1]{doi: #1}\else
  \providecommand{\doi}{doi: \begingroup \urlstyle{rm}\Url}\fi

\bibitem[Acar et~al.(2021)Acar, Zhao, Matas, Mattina, Whatmough, and Saligrama]{acar2021federated}
Acar, D. A.~E., Zhao, Y., Matas, R., Mattina, M., Whatmough, P., and Saligrama, V.
\newblock Federated learning based on dynamic regularization.
\newblock In \emph{International Conference on Learning Representations}, 2021.

\bibitem[Allen-Zhu \& Li(2023)Allen-Zhu and Li]{allen2020towards}
Allen-Zhu, Z. and Li, Y.
\newblock Towards understanding ensemble, knowledge distillation and self-distillation in deep learning.
\newblock \emph{International Conference on Learning Representations}, 2023.

\bibitem[Bao et~al.(2024)Bao, Crawshaw, and Liu]{baoprovable}
Bao, Y., Crawshaw, M., and Liu, M.
\newblock Provable benefits of local steps in heterogeneous federated learning for neural networks: A feature learning perspective.
\newblock In \emph{Forty-first International Conference on Machine Learning}, 2024.

\bibitem[Barnes et~al.(2022)Barnes, Dytso, and Poor]{barnes2022improved}
Barnes, L.~P., Dytso, A., and Poor, H.~V.
\newblock Improved information theoretic generalization bounds for distributed and federated learning.
\newblock In \emph{2022 IEEE International Symposium on Information Theory (ISIT)}, pp.\  1465--1470. IEEE, 2022.

\bibitem[Cao et~al.(2022)Cao, Chen, Belkin, and Gu]{cao2022benign}
Cao, Y., Chen, Z., Belkin, M., and Gu, Q.
\newblock Benign overfitting in two-layer convolutional neural networks.
\newblock \emph{Advances in Neural Information Processing Systems}, 35:\penalty0 25237--25250, 2022.

\bibitem[Chen et~al.(2022)Chen, Tu, Li, Shen, and Chao]{chen2022importance}
Chen, H.-Y., Tu, C.-H., Li, Z., Shen, H.-W., and Chao, W.-L.
\newblock On the importance and applicability of pre-training for federated learning.
\newblock \emph{International Conference on Learning Representations}, 2022.

\bibitem[Chen et~al.(2021)Chen, Zheng, Long, and Su]{chen2021theorem}
Chen, S., Zheng, Q., Long, Q., and Su, W.~J.
\newblock A theorem of the alternative for personalized federated learning.
\newblock \emph{arXiv preprint arXiv:2103.01901}, 2021.

\bibitem[Cheng et~al.(2021)Cheng, Chadha, and Duchi]{cheng2021fine}
Cheng, G., Chadha, K., and Duchi, J.
\newblock Fine-tuning is fine in federated learning.
\newblock \emph{arXiv preprint arXiv:2108.07313}, 3, 2021.

\bibitem[Collins et~al.(2022)Collins, Hassani, Mokhtari, and Shakkottai]{collins2022fedavg}
Collins, L., Hassani, H., Mokhtari, A., and Shakkottai, S.
\newblock Fedavg with fine tuning: Local updates lead to representation learning.
\newblock \emph{Advances in Neural Information Processing Systems}, 35:\penalty0 10572--10586, 2022.

\bibitem[De et~al.(2022)De, Berrada, Hayes, Smith, and Balle]{de2022unlocking}
De, S., Berrada, L., Hayes, J., Smith, S.~L., and Balle, B.
\newblock Unlocking high-accuracy differentially private image classification through scale.
\newblock \emph{arXiv preprint arXiv:2204.13650}, 2022.

\bibitem[Devlin et~al.(2019)Devlin, Chang, Lee, and Toutanova]{Devlin2019BERTPO}
Devlin, J., Chang, M.-W., Lee, K., and Toutanova, K.
\newblock Bert: Pre-training of deep bidirectional transformers for language understanding.
\newblock In \emph{North American Chapter of the Association for Computational Linguistics}, 2019.

\bibitem[Devroye et~al.(2018)Devroye, Mehrabian, and Reddad]{devroye2018total}
Devroye, L., Mehrabian, A., and Reddad, T.
\newblock The total variation distance between high-dimensional gaussians with the same mean.
\newblock \emph{arXiv preprint arXiv:1810.08693}, 2018.

\bibitem[Dosovitskiy et~al.(2021)Dosovitskiy, Beyer, Kolesnikov, Weissenborn, Zhai, Unterthiner, Dehghani, Minderer, Heigold, Gelly, et~al.]{dosovitskiy2020image}
Dosovitskiy, A., Beyer, L., Kolesnikov, A., Weissenborn, D., Zhai, X., Unterthiner, T., Dehghani, M., Minderer, M., Heigold, G., Gelly, S., et~al.
\newblock An image is worth 16x16 words: Transformers for image recognition at scale.
\newblock \emph{International Conference on Learning Representations}, 2021.

\bibitem[Du et~al.(2018)Du, Lee, Tian, Singh, and Poczos]{du2018gradient}
Du, S., Lee, J., Tian, Y., Singh, A., and Poczos, B.
\newblock Gradient descent learns one-hidden-layer cnn: Don’t be afraid of spurious local minima.
\newblock In \emph{International Conference on Machine Learning}, pp.\  1339--1348. PMLR, 2018.

\bibitem[Dwork et~al.(2006)Dwork, McSherry, Nissim, and Smith]{dwork2006calibrating}
Dwork, C., McSherry, F., Nissim, K., and Smith, A.
\newblock Calibrating noise to sensitivity in private data analysis.
\newblock In \emph{Theory of Cryptography: Third Theory of Cryptography Conference, TCC 2006, New York, NY, USA, March 4-7, 2006. Proceedings 3}, pp.\  265--284. Springer, 2006.

\bibitem[Fallah et~al.(2021)Fallah, Mokhtari, and Ozdaglar]{fallah2021generalization}
Fallah, A., Mokhtari, A., and Ozdaglar, A.
\newblock Generalization of model-agnostic meta-learning algorithms: Recurring and unseen tasks.
\newblock \emph{Advances in Neural Information Processing Systems}, 34:\penalty0 5469--5480, 2021.

\bibitem[Fan{\`\i} et~al.(2023)Fan{\`\i}, Camoriano, Caputo, and Ciccone]{fani2023fed3r}
Fan{\`\i}, E., Camoriano, R., Caputo, B., and Ciccone, M.
\newblock Fed3r: Recursive ridge regression for federated learning with strong pre-trained models.
\newblock In \emph{International Workshop on Federated Learning in the Age of Foundation Models in Conjunction with NeurIPS 2023}, 2023.

\bibitem[Ganesh et~al.(2023)Ganesh, Haghifam, Nasr, Oh, Steinke, Thakkar, Thakurta, and Wang]{ganesh2023public}
Ganesh, A., Haghifam, M., Nasr, M., Oh, S., Steinke, T., Thakkar, O., Thakurta, A.~G., and Wang, L.
\newblock Why is public pretraining necessary for private model training?
\newblock In \emph{International Conference on Machine Learning}, pp.\  10611--10627. PMLR, 2023.

\bibitem[Gao et~al.(2020)Gao, Biderman, Black, Golding, Hoppe, Foster, Phang, He, Thite, Nabeshima, et~al.]{gao2020pile}
Gao, L., Biderman, S., Black, S., Golding, L., Hoppe, T., Foster, C., Phang, J., He, H., Thite, A., Nabeshima, N., et~al.
\newblock The pile: An 800gb dataset of diverse text for language modeling.
\newblock \emph{arXiv preprint arXiv:2101.00027}, 2020.

\bibitem[Gholami \& Seferoglu(2024)Gholami and Seferoglu]{gholami2024improved}
Gholami, P. and Seferoglu, H.
\newblock Improved generalization bounds for communication efficient federated learning.
\newblock \emph{arXiv preprint arXiv:2404.11754}, 2024.

\bibitem[Glorot \& Bengio(2010)Glorot and Bengio]{DBLP:journals/jmlr/GlorotB10}
Glorot, X. and Bengio, Y.
\newblock Understanding the difficulty of training deep feedforward neural networks.
\newblock In Teh, Y.~W. and Titterington, D.~M. (eds.), \emph{Proceedings of the Thirteenth International Conference on Artificial Intelligence and Statistics, {AISTATS} 2010, Chia Laguna Resort, Sardinia, Italy, May 13-15, 2010}, volume~9 of \emph{{JMLR} Proceedings}, pp.\  249--256. JMLR.org, 2010.
\newblock URL \url{http://proceedings.mlr.press/v9/glorot10a.html}.

\bibitem[Gupta et~al.(2022)Gupta, Huang, Zhong, Gao, Li, and Chen]{gupta2022recovering}
Gupta, S., Huang, Y., Zhong, Z., Gao, T., Li, K., and Chen, D.
\newblock Recovering private text in federated learning of language models.
\newblock \emph{Advances in Neural Information Processing Systems}, 35:\penalty0 8130--8143, 2022.

\bibitem[He et~al.(2015)He, Zhang, Ren, and Sun]{DBLP:conf/iccv/HeZRS15}
He, K., Zhang, X., Ren, S., and Sun, J.
\newblock Delving deep into rectifiers: Surpassing human-level performance on imagenet classification.
\newblock In \emph{2015 {IEEE} International Conference on Computer Vision, {ICCV} 2015, Santiago, Chile, December 7-13, 2015}, pp.\  1026--1034. {IEEE} Computer Society, 2015.
\newblock \doi{10.1109/ICCV.2015.123}.
\newblock URL \url{https://doi.org/10.1109/ICCV.2015.123}.

\bibitem[He et~al.(2016)He, Zhang, Ren, and Sun]{DBLP:conf/cvpr/HeZRS16}
He, K., Zhang, X., Ren, S., and Sun, J.
\newblock Deep residual learning for image recognition.
\newblock In \emph{2016 {IEEE} Conference on Computer Vision and Pattern Recognition, {CVPR} 2016, Las Vegas, NV, USA, June 27-30, 2016}, pp.\  770--778. {IEEE} Computer Society, 2016.
\newblock \doi{10.1109/CVPR.2016.90}.
\newblock URL \url{https://doi.org/10.1109/CVPR.2016.90}.

\bibitem[He et~al.(2019)He, Girshick, and Doll{\'a}r]{he2019rethinking}
He, K., Girshick, R., and Doll{\'a}r, P.
\newblock Rethinking imagenet pre-training.
\newblock In \emph{Proceedings of the IEEE/CVF International Conference on Computer Vision}, pp.\  4918--4927, 2019.

\bibitem[Hou et~al.(2024)Hou, Shrivastava, Zhan, Conway, Le, Sagar, Fanti, and Lazar]{hou2024pre}
Hou, C., Shrivastava, A., Zhan, H., Conway, R., Le, T., Sagar, A., Fanti, G., and Lazar, D.
\newblock Pre-text: Training language models on private federated data in the age of llms.
\newblock \emph{arXiv preprint arXiv:2406.02958}, 2024.

\bibitem[Hsu et~al.(2019)Hsu, Qi, and Brown]{hsu2019noniid}
Hsu, T.-M.~H., Qi, H., and Brown, M.
\newblock Measuring the effects of non-identical data distribution for federated visual classification.
\newblock In \emph{International Workshop on Federated Learning for User Privacy and Data Confidentiality in Conjunction with NeurIPS 2019 (FL-NeurIPS'19)}, December 2019.

\bibitem[Hu et~al.(2022)Hu, Li, and Liu]{hu2022generalization}
Hu, X., Li, S., and Liu, Y.
\newblock Generalization bounds for federated learning: Fast rates, unparticipating clients and unbounded losses.
\newblock In \emph{The Eleventh International Conference on Learning Representations}, 2022.

\bibitem[Huang et~al.(2021)Huang, Li, Song, and Yang]{huang2021fl}
Huang, B., Li, X., Song, Z., and Yang, X.
\newblock Fl-ntk: A neural tangent kernel-based framework for federated learning analysis.
\newblock In \emph{International Conference on Machine Learning}, pp.\  4423--4434. PMLR, 2021.

\bibitem[Huang et~al.(2023)Huang, Shi, Cai, and Suzuki]{huang2023understanding}
Huang, W., Shi, Y., Cai, Z., and Suzuki, T.
\newblock Understanding convergence and generalization in federated learning through feature learning theory.
\newblock In \emph{The Twelfth International Conference on Learning Representations}, 2023.

\bibitem[Iandola et~al.(2016)Iandola, Han, Moskewicz, Ashraf, Dally, and Keutzer]{iandola2016squeezenet}
Iandola, F.~N., Han, S., Moskewicz, M.~W., Ashraf, K., Dally, W.~J., and Keutzer, K.
\newblock Squeezenet: Alexnet-level accuracy with 50x fewer parameters and< 0.5 mb model size.
\newblock \emph{arXiv preprint arXiv:1602.07360}, 2016.

\bibitem[Jelassi \& Li(2022)Jelassi and Li]{jelassi2022towards}
Jelassi, S. and Li, Y.
\newblock Towards understanding how momentum improves generalization in deep learning.
\newblock In \emph{International Conference on Machine Learning}, pp.\  9965--10040. PMLR, 2022.

\bibitem[Kairouz et~al.(2021)Kairouz, McMahan, Avent, Bellet, Bennis, Bhagoji, Bonawitz, Charles, Cormode, Cummings, et~al.]{kairouz2021advances}
Kairouz, P., McMahan, H.~B., Avent, B., Bellet, A., Bennis, M., Bhagoji, A.~N., Bonawitz, K., Charles, Z., Cormode, G., Cummings, R., et~al.
\newblock Advances and open problems in federated learning.
\newblock \emph{Foundations and Trends{\textregistered} in Machine Learning}, 14\penalty0 (1--2):\penalty0 1--210, 2021.

\bibitem[Karimireddy et~al.(2020)Karimireddy, Kale, Mohri, Reddi, Stich, and Suresh]{karimireddy2020scaffold}
Karimireddy, S.~P., Kale, S., Mohri, M., Reddi, S., Stich, S., and Suresh, A.~T.
\newblock Scaffold: Stochastic controlled averaging for federated learning.
\newblock In \emph{International Conference on Machine Learning}, pp.\  5132--5143. PMLR, 2020.

\bibitem[Karimireddy et~al.(2021)Karimireddy, Jaggi, Kale, Mohri, Reddi, Stich, and Suresh]{karimireddy2021breaking}
Karimireddy, S.~P., Jaggi, M., Kale, S., Mohri, M., Reddi, S., Stich, S.~U., and Suresh, A.~T.
\newblock Breaking the centralized barrier for cross-device federated learning.
\newblock \emph{Advances in Neural Information Processing Systems}, 34:\penalty0 28663--28676, 2021.

\bibitem[Kou et~al.(2023)Kou, Chen, Chen, and Gu]{kou2023benign}
Kou, Y., Chen, Z., Chen, Y., and Gu, Q.
\newblock Benign overfitting in two-layer relu convolutional neural networks.
\newblock In \emph{International Conference on Machine Learning}, pp.\  17615--17659. PMLR, 2023.

\bibitem[Krizhevsky(2009)]{krizhevsky2009learning}
Krizhevsky, A.
\newblock Learning multiple layers of features from tiny images.
\newblock Technical report, University of Toronto, 2009.
\newblock URL \url{https://www.cs.toronto.edu/~kriz/learning-features-2009-TR.pdf}.

\bibitem[Kumar et~al.(2022)Kumar, Raghunathan, Jones, Ma, and Liang]{kumar2022fine}
Kumar, A., Raghunathan, A., Jones, R., Ma, T., and Liang, P.
\newblock Fine-tuning can distort pretrained features and underperform out-of-distribution.
\newblock \emph{International Conference on Learning Representations}, 2022.

\bibitem[Le \& Yang(2015)Le and Yang]{le2015tiny}
Le, Y. and Yang, X.
\newblock Tiny imagenet visual recognition challenge.
\newblock \emph{CS 231N}, 7\penalty0 (7):\penalty0 3, 2015.

\bibitem[LeCun et~al.(2012)LeCun, Bottou, Orr, and M{\"{u}}ller]{DBLP:series/lncs/LeCunBOM12}
LeCun, Y., Bottou, L., Orr, G.~B., and M{\"{u}}ller, K.
\newblock Efficient backprop.
\newblock In Montavon, G., Orr, G.~B., and M{\"{u}}ller, K. (eds.), \emph{Neural Networks: Tricks of the Trade - Second Edition}, volume 7700 of \emph{Lecture Notes in Computer Science}, pp.\  9--48. Springer, 2012.
\newblock \doi{10.1007/978-3-642-35289-8\_3}.
\newblock URL \url{https://doi.org/10.1007/978-3-642-35289-8\_3}.

\bibitem[Legate et~al.(2024)Legate, Bernier, Page-Caccia, Oyallon, and Belilovsky]{legate2024guiding}
Legate, G., Bernier, N., Page-Caccia, L., Oyallon, E., and Belilovsky, E.
\newblock Guiding the last layer in federated learning with pre-trained models.
\newblock \emph{Advances in Neural Information Processing Systems}, 36, 2024.

\bibitem[Li et~al.(2021)Li, He, and Song]{li2021model}
Li, Q., He, B., and Song, D.
\newblock Model-contrastive federated learning.
\newblock In \emph{Proceedings of the IEEE/CVF Conference on Computer Vision and Pattern Recognition}, pp.\  10713--10722, 2021.

\bibitem[Li et~al.(2020)Li, Sahu, Talwalkar, and Smith]{li2020federatedchallenge}
Li, T., Sahu, A.~K., Talwalkar, A., and Smith, V.
\newblock Federated learning: Challenges, methods, and future directions.
\newblock \emph{IEEE Signal Processing Magazine}, 37\penalty0 (3):\penalty0 50--60, 2020.

\bibitem[Li et~al.(2022{\natexlab{a}})Li, Liu, Hashimoto, Inan, Kulkarni, Lee, and Guha~Thakurta]{li2022does}
Li, X., Liu, D., Hashimoto, T.~B., Inan, H.~A., Kulkarni, J., Lee, Y.-T., and Guha~Thakurta, A.
\newblock When does differentially private learning not suffer in high dimensions?
\newblock \emph{Advances in Neural Information Processing Systems}, 35:\penalty0 28616--28630, 2022{\natexlab{a}}.

\bibitem[Li et~al.(2022{\natexlab{b}})Li, Tramer, Liang, and Hashimoto]{li2021large}
Li, X., Tramer, F., Liang, P., and Hashimoto, T.
\newblock Large language models can be strong differentially private learners.
\newblock \emph{International Conference on Learning Representations}, 2022{\natexlab{b}}.

\bibitem[Lin et~al.(2013)Lin, Chen, and Yan]{lin2013network}
Lin, M., Chen, Q., and Yan, S.
\newblock Network in network.
\newblock \emph{arXiv preprint arXiv:1312.4400}, 2013.

\bibitem[Lin et~al.(2020)Lin, Kong, Stich, and Jaggi]{lin2020ensemble}
Lin, T., Kong, L., Stich, S.~U., and Jaggi, M.
\newblock Ensemble distillation for robust model fusion in federated learning.
\newblock \emph{Advances in Neural Information Processing Systems}, 33:\penalty0 2351--2363, 2020.

\bibitem[Liu \& Miller(2020)Liu and Miller]{liu2020federated}
Liu, D. and Miller, T.
\newblock Federated pretraining and fine tuning of bert using clinical notes from multiple silos.
\newblock \emph{arXiv preprint arXiv:2002.08562}, 2020.

\bibitem[Mahajan et~al.(2018)Mahajan, Girshick, Ramanathan, He, Paluri, Li, Bharambe, and Van Der~Maaten]{mahajan2018exploring}
Mahajan, D., Girshick, R., Ramanathan, V., He, K., Paluri, M., Li, Y., Bharambe, A., and Van Der~Maaten, L.
\newblock Exploring the limits of weakly supervised pretraining.
\newblock In \emph{Proceedings of the European Conference on Computer Vision (ECCV)}, pp.\  181--196, 2018.

\bibitem[McMahan et~al.(2017)McMahan, Moore, Ramage, Hampson, and y~Arcas]{mcmahan2017communication}
McMahan, B., Moore, E., Ramage, D., Hampson, S., and y~Arcas, B.~A.
\newblock Communication-efficient learning of deep networks from decentralized data.
\newblock In \emph{Artificial Intelligence and Statistics}, pp.\  1273--1282. PMLR, 2017.

\bibitem[Mohri et~al.(2019)Mohri, Sivek, and Suresh]{mohri2019agnostic}
Mohri, M., Sivek, G., and Suresh, A.~T.
\newblock Agnostic federated learning.
\newblock In \emph{International Conference on Machine Learning}, pp.\  4615--4625. PMLR, 2019.

\bibitem[Nguyen et~al.(2022)Nguyen, Wang, Malik, Sanjabi, and Rabbat]{nguyen2022begin}
Nguyen, J., Wang, J., Malik, K., Sanjabi, M., and Rabbat, M.
\newblock Where to begin? on the impact of pre-training and initialization in federated learning.
\newblock \emph{International Conference on Learning Representations}, 2022.

\bibitem[Oh \& Yun(2024)Oh and Yun]{oh2024provable}
Oh, J. and Yun, C.
\newblock Provable benefit of cutout and cutmix for feature learning.
\newblock In \emph{High-dimensional Learning Dynamics 2024: The Emergence of Structure and Reasoning}, 2024.

\bibitem[Paszke et~al.(2019)Paszke, Gross, Massa, Lerer, Bradbury, Chanan, Killeen, Lin, Gimelshein, Antiga, et~al.]{paszke2019pytorch}
Paszke, A., Gross, S., Massa, F., Lerer, A., Bradbury, J., Chanan, G., Killeen, T., Lin, Z., Gimelshein, N., Antiga, L., et~al.
\newblock Pytorch: An imperative style, high-performance deep learning library.
\newblock \emph{Advances in Neural Information Processing Systems}, 32, 2019.

\bibitem[Radford et~al.(2019)Radford, Wu, Child, Luan, Amodei, Sutskever, et~al.]{radford2019language}
Radford, A., Wu, J., Child, R., Luan, D., Amodei, D., Sutskever, I., et~al.
\newblock Language models are unsupervised multitask learners.
\newblock \emph{OpenAI blog}, 1\penalty0 (8):\penalty0 9, 2019.

\bibitem[Raffel et~al.(2020)Raffel, Shazeer, Roberts, Lee, Narang, Matena, Zhou, Li, and Liu]{raffel2020exploring}
Raffel, C., Shazeer, N., Roberts, A., Lee, K., Narang, S., Matena, M., Zhou, Y., Li, W., and Liu, P.~J.
\newblock Exploring the limits of transfer learning with a unified text-to-text transformer.
\newblock \emph{Journal of Machine Learning Research}, 21\penalty0 (140):\penalty0 1--67, 2020.

\bibitem[Reddi et~al.(2021)Reddi, Charles, Zaheer, Garrett, Rush, Kone{\v{c}}n{\`y}, Kumar, and McMahan]{reddi2020adaptive}
Reddi, S., Charles, Z., Zaheer, M., Garrett, Z., Rush, K., Kone{\v{c}}n{\`y}, J., Kumar, S., and McMahan, H.~B.
\newblock Adaptive federated optimization.
\newblock \emph{International Conference on Learning Representations}, 2021.

\bibitem[Russakovsky et~al.(2015)Russakovsky, Deng, Su, Krause, Satheesh, Ma, Huang, Karpathy, Khosla, Bernstein, et~al.]{russakovsky2015imagenet}
Russakovsky, O., Deng, J., Su, H., Krause, J., Satheesh, S., Ma, S., Huang, Z., Karpathy, A., Khosla, A., Bernstein, M., et~al.
\newblock Imagenet large scale visual recognition challenge.
\newblock \emph{International Journal of Computer Vision}, 115:\penalty0 211--252, 2015.

\bibitem[Schuhmann et~al.(2022)Schuhmann, Beaumont, Vencu, Gordon, Wightman, Cherti, Coombes, Katta, Mullis, Wortsman, et~al.]{schuhmann2022laion}
Schuhmann, C., Beaumont, R., Vencu, R., Gordon, C., Wightman, R., Cherti, M., Coombes, T., Katta, A., Mullis, C., Wortsman, M., et~al.
\newblock Laion-5b: An open large-scale dataset for training next generation image-text models.
\newblock \emph{Advances in Neural Information Processing Systems}, 35:\penalty0 25278--25294, 2022.

\bibitem[Sefidgaran et~al.(2022)Sefidgaran, Chor, and Zaidi]{sefidgaran2022rate}
Sefidgaran, M., Chor, R., and Zaidi, A.
\newblock Rate-distortion theoretic bounds on generalization error for distributed learning.
\newblock \emph{Advances in Neural Information Processing Systems}, 35:\penalty0 19687--19702, 2022.

\bibitem[Sun et~al.(2017)Sun, Shrivastava, Singh, and Gupta]{sun2017revisiting}
Sun, C., Shrivastava, A., Singh, S., and Gupta, A.
\newblock Revisiting unreasonable effectiveness of data in deep learning era.
\newblock In \emph{Proceedings of the IEEE International Conference on Computer Vision}, pp.\  843--852, 2017.

\bibitem[Sun \& Wei(2022)Sun and Wei]{sun2022communication}
Sun, Z. and Wei, E.
\newblock A communication-efficient algorithm with linear convergence for federated minimax learning.
\newblock \emph{Advances in Neural Information Processing Systems}, 35:\penalty0 6060--6073, 2022.

\bibitem[Sun et~al.(2024)Sun, Niu, and Wei]{sun2024understanding}
Sun, Z., Niu, X., and Wei, E.
\newblock Understanding generalization of federated learning via stability: Heterogeneity matters.
\newblock In \emph{International Conference on Artificial Intelligence and Statistics}, pp.\  676--684. PMLR, 2024.

\bibitem[Tan et~al.(2022)Tan, Long, Ma, Liu, Zhou, and Jiang]{tan2022federated}
Tan, Y., Long, G., Ma, J., Liu, L., Zhou, T., and Jiang, J.
\newblock Federated learning from pre-trained models: A contrastive learning approach.
\newblock \emph{Advances in Neural Information Processing Systems}, 35:\penalty0 19332--19344, 2022.

\bibitem[Thomee et~al.(2016)Thomee, Shamma, Friedland, Elizalde, Ni, Poland, Borth, and Li]{thomee2016yfcc100m}
Thomee, B., Shamma, D.~A., Friedland, G., Elizalde, B., Ni, K., Poland, D., Borth, D., and Li, L.-J.
\newblock Yfcc100m: The new data in multimedia research.
\newblock \emph{Communications of the ACM}, 59\penalty0 (2):\penalty0 64--73, 2016.

\bibitem[Tian et~al.(2022)Tian, Wan, Lyu, Yao, Jin, and Sun]{tian2022fedbert}
Tian, Y., Wan, Y., Lyu, L., Yao, D., Jin, H., and Sun, L.
\newblock Fedbert: When federated learning meets pre-training.
\newblock \emph{ACM Transactions on Intelligent Systems and Technology (TIST)}, 13\penalty0 (4):\penalty0 1--26, 2022.

\bibitem[Wang et~al.(2018)Wang, Singh, Michael, Hill, Levy, and Bowman]{wang2018glue}
Wang, A., Singh, A., Michael, J., Hill, F., Levy, O., and Bowman, S.~R.
\newblock Glue: A multi-task benchmark and analysis platform for natural language understanding.
\newblock \emph{arXiv preprint arXiv:1804.07461}, 2018.

\bibitem[Wang et~al.(2023)Wang, Zhang, Cao, Li, McMahan, Oh, Xu, and Zaheer]{wang2023can}
Wang, B., Zhang, Y.~J., Cao, Y., Li, B., McMahan, H.~B., Oh, S., Xu, Z., and Zaheer, M.
\newblock Can public large language models help private cross-device federated learning?
\newblock \emph{arXiv preprint arXiv:2305.12132}, 2023.

\bibitem[Wang et~al.(2020)Wang, Yurochkin, Sun, Papailiopoulos, and Khazaeni]{wang2020federated}
Wang, H., Yurochkin, M., Sun, Y., Papailiopoulos, D.~S., and Khazaeni, Y.
\newblock Federated learning with matched averaging.
\newblock In \emph{International Conference on Learning Representations}, 2020.

\bibitem[Wang \& Joshi(2019)Wang and Joshi]{wang2019adaptive}
Wang, J. and Joshi, G.
\newblock Adaptive communication strategies to achieve the best error-runtime trade-off in local-update sgd.
\newblock \emph{Proceedings of Machine Learning and Systems}, 1:\penalty0 212--229, 2019.

\bibitem[Wu et~al.(2024)Wu, Xu, Zhang, Zhang, and Ramage]{wu2024prompt}
Wu, S., Xu, Z., Zhang, Y., Zhang, Y., and Ramage, D.
\newblock Prompt public large language models to synthesize data for private on-device applications.
\newblock \emph{COLM}, 2024.

\bibitem[Wu \& He(2018)Wu and He]{wu2018group}
Wu, Y. and He, K.
\newblock Group normalization.
\newblock In \emph{Proceedings of the European conference on computer vision (ECCV)}, pp.\  3--19, 2018.

\bibitem[Xu et~al.(2023{\natexlab{a}})Xu, Collins, Wang, Panait, Oh, Augenstein, Liu, Schroff, and McMahan]{xu2023learning}
Xu, Z., Collins, M., Wang, Y., Panait, L., Oh, S., Augenstein, S., Liu, T., Schroff, F., and McMahan, H.~B.
\newblock Learning to generate image embeddings with user-level differential privacy.
\newblock In \emph{Proceedings of the IEEE/CVF Conference on Computer Vision and Pattern Recognition}, pp.\  7969--7980, 2023{\natexlab{a}}.

\bibitem[Xu et~al.(2023{\natexlab{b}})Xu, Zhang, Andrew, Choquette, Kairouz, Mcmahan, Rosenstock, and Zhang]{xu2023federated}
Xu, Z., Zhang, Y., Andrew, G., Choquette, C., Kairouz, P., Mcmahan, B., Rosenstock, J., and Zhang, Y.
\newblock Federated learning of gboard language models with differential privacy.
\newblock In Sitaram, S., Beigman~Klebanov, B., and Williams, J.~D. (eds.), \emph{Proceedings of the 61st Annual Meeting of the Association for Computational Linguistics (Volume 5: Industry Track)}, pp.\  629--639, Toronto, Canada, July 2023{\natexlab{b}}. Association for Computational Linguistics.
\newblock \doi{10.18653/v1/2023.acl-industry.60}.
\newblock URL \url{https://aclanthology.org/2023.acl-industry.60/}.

\bibitem[Yang et~al.(2021{\natexlab{a}})Yang, Wang, Xu, Chen, Bian, Liu, and Liu]{yang2021characterizing}
Yang, C., Wang, Q., Xu, M., Chen, Z., Bian, K., Liu, Y., and Liu, X.
\newblock Characterizing impacts of heterogeneity in federated learning upon large-scale smartphone data.
\newblock In \emph{Proceedings of the Web Conference 2021}, pp.\  935--946, 2021{\natexlab{a}}.

\bibitem[Yang et~al.(2021{\natexlab{b}})Yang, Fang, and Liu]{yang2021achieving}
Yang, H., Fang, M., and Liu, J.
\newblock Achieving linear speedup with partial worker participation in non-iid federated learning.
\newblock \emph{International Conference on Learning Representations}, 2021{\natexlab{b}}.

\bibitem[Ye et~al.(2023)Ye, Zhu, Liu, Shokri, and Cevher]{DBLP:conf/nips/init_matters_2}
Ye, J., Zhu, Z., Liu, F., Shokri, R., and Cevher, V.
\newblock Initialization matters: Privacy-utility analysis of overparameterized neural networks.
\newblock In Oh, A., Naumann, T., Globerson, A., Saenko, K., Hardt, M., and Levine, S. (eds.), \emph{Advances in Neural Information Processing Systems 36: Annual Conference on Neural Information Processing Systems 2023, NeurIPS 2023, New Orleans, LA, USA, December 10 - 16, 2023}, 2023.
\newblock URL \url{http://papers.nips.cc/paper\_files/paper/2023/hash/1165af8b913fb836c6280b42d6e0084f-Abstract-Conference.html}.

\bibitem[Yu et~al.(2014)Yu, Wang, Chen, and Wei]{yu2014mixed}
Yu, D., Wang, H., Chen, P., and Wei, Z.
\newblock Mixed pooling for convolutional neural networks.
\newblock In \emph{Rough Sets and Knowledge Technology: 9th International Conference, RSKT 2014, Shanghai, China, October 24-26, 2014, Proceedings 9}, pp.\  364--375. Springer, 2014.

\bibitem[Yu et~al.(2022)Yu, Naik, Backurs, Gopi, Inan, Kamath, Kulkarni, Lee, Manoel, Wutschitz, Yekhanin, and Zhang]{yu2022differentially}
Yu, D., Naik, S., Backurs, A., Gopi, S., Inan, H.~A., Kamath, G., Kulkarni, J., Lee, Y.~T., Manoel, A., Wutschitz, L., Yekhanin, S., and Zhang, H.
\newblock Differentially private fine-tuning of language models.
\newblock In \emph{International Conference on Learning Representations (ICLR)}, 2022.

\bibitem[Yuan et~al.(2021)Yuan, Morningstar, Ning, and Singhal]{yuan2021we}
Yuan, H., Morningstar, W., Ning, L., and Singhal, K.
\newblock What do we mean by generalization in federated learning?
\newblock \emph{arXiv preprint arXiv:2110.14216}, 2021.

\bibitem[Zhang et~al.(2023)Zhang, Feng, Alam, Dimitriadis, Zhang, Narayanan, and Avestimehr]{zhang2023gpt}
Zhang, T., Feng, T., Alam, S., Dimitriadis, D., Zhang, M., Narayanan, S.~S., and Avestimehr, S.
\newblock Gpt-fl: Generative pre-trained model-assisted federated learning.
\newblock \emph{arXiv preprint arXiv:2306.02210}, 2023.

\bibitem[Zhuang et~al.(2023)Zhuang, Chen, and Lyu]{zhuang2023foundation}
Zhuang, W., Chen, C., and Lyu, L.
\newblock When foundation model meets federated learning: Motivations, challenges, and future directions.
\newblock \emph{arXiv preprint arXiv:2306.15546}, 2023.

\bibitem[Zou et~al.(2023)Zou, Cao, Li, and Gu]{zou2021understanding}
Zou, D., Cao, Y., Li, Y., and Gu, Q.
\newblock Understanding the generalization of adam in learning neural networks with proper regularization.
\newblock \emph{International Conference on Learning Representations}, 2023.

\end{thebibliography}
\bibliographystyle{icml2025}

%%%%%%%%%%%%%%%%%%%%%%%%%%%%%%%%%%%%%%%%%%%%%%%%%%%%%%%%%%%%%%%%%%%%%%%%%%%%%%%
%%%%%%%%%%%%%%%%%%%%%%%%%%%%%%%%%%%%%%%%%%%%%%%%%%%%%%%%%%%%%%%%%%%%%%%%%%%%%%%
% APPENDIX
%%%%%%%%%%%%%%%%%%%%%%%%%%%%%%%%%%%%%%%%%%%%%%%%%%%%%%%%%%%%%%%%%%%%%%%%%%%%%%%
%%%%%%%%%%%%%%%%%%%%%%%%%%%%%%%%%%%%%%%%%%%%%%%%%%%%%%%%%%%%%%%%%%%%%%%%%%%%%%%
\newpage
\appendix
\onecolumn

\startcontents[sections]
\printcontents[sections]{l}{1}{\setcounter{tocdepth}{2}}

\clearpage

\section{Additional Related Work}

\paragraph{Use of Pre-Trained Models in Federated Learning.} \cite{tan2022federated}
explore the benefit of using pre-trained models in FL by proposing to use multiple fixed pre-trained backbones as the encoder model at each client and using contrastive learning to extract useful shared representations. \cite{zhuang2023foundation} discuss the opportunities and challenges of using large foundation models for FL including the high communication and computation cost. One solution to this as proposed by \cite{legate2024guiding} is that instead of full fine-tuning as done in \cite{chen2022importance, nguyen2022begin}, we can just fine-tune the last layer. Specifically \cite{legate2024guiding} proposes a two-stage approach to federated fine-tuning by first fine-tuning the head and then doing a full-finetuning. This approach is inspired by results in the centralized setting \cite{kumar2022fine} which show that in some case fine-tuning can distort the pre-trained features. \cite{fani2023fed3r} also study the problem of fine-tuning just the last layer in a federated setting by replacing the softmax classifier with a ridge-regression classifier which enables them to compute a closed form expression for the last layer weights.

There has also been some recent work on exploring the benefit of pre-training for federated natural language processing tasks including the use of Large Language Models (LLMs). \cite{wang2023can} discuss how to leverage the power of pre-trained LLMs for private on-device fine-tuning of language models. Specifically, \cite{wang2023can} proposes a distribution matching approach to select public data that is closest to private data and then use this selected public data to train the on-device language model. \cite{zhang2023gpt} propose to first pre-train on synthetic data to construct the initialization point followed by federated fine-tuning. \cite{hou2024pre} propose that clients send DP information to the server which then uses this information to generate synthetic data and fine-tune centrally on this synthetic data.
\cite{liu2020federated} discuss the challenges of pre-training and fine-tuning BERT in federated manner using clinical notes from multiple silos without data transfer. \cite{tian2022fedbert} propose to pre-train a BERT model in a federated manner in a more general setting and show that their pre-trained model can retain accuracy on the GLUE \citep{wang2018glue} dataset without sacrificing client privacy. \citet{xu2023federated} pretrain production on-device language models on public web data before fine-tuning in federated learning with differential privacy, and \citet{wu2024prompt} later replace the pretraining data with data synthesized by LLMs. \cite{gupta2022recovering} propose a defense using pre-trained models to prevent an attacker from recovering multiple sentences from gradients in the federated
training of the language modeling task.

\paragraph{Importance of Initialization for Private Optimization.} We note that an orthogonal line of work has explored the benefits of starting from a pre-trained model when doing differentially private optimization \cite{dwork2006calibrating} and seen similar striking improvement in accuracy \cite{de2022unlocking, li2021large,yu2022differentially,xu2023learning}, as we see in the heterogeneous FL setting.  \cite{ganesh2023public} study this phenomenon for a stylized mean estimation problem and show that public pre-training can help the model start from a good loss basin which is otherwise hard to achieve with private noisy optimization. \cite{li2022does} study differentially private convex optimization and show that starting from a pre-trained model can leads to dimension independent convergence guarantees. Specifically \cite{li2022does} define the notion of restricted Lipschitz continuity and show that when gradients are low rank most of the restricted Lispchitz coefficients will be zero. \cite{DBLP:conf/nips/init_matters_2} studies the impact of different random initializations on the privacy bound when training overparameterized neural networks and shows that for some initializations (LeCun \cite{DBLP:series/lncs/LeCunBOM12}, Xavier \cite{DBLP:journals/jmlr/GlorotB10}) the privacy bound improves with increasing depth while for other initializations (He \cite{DBLP:conf/iccv/HeZRS15}, NTK \cite{allen2020towards}) it degrades with increasing depth.

\paragraph{Generalization performance in Federated Learning.}
Several existing works have studied the generalization performance of FL in different settings \cite{cheng2021fine, gholami2024improved, huang2023understanding, yuan2021we}. Some of the initial works either provide results independent of the algorithm being used \cite{mohri2019agnostic, hu2022generalization, sun2022communication}, or only study convex losses \cite{chen2021theorem, fallah2021generalization}.
\cite{barnes2022improved, sefidgaran2022rate} derive information-theoretic bounds, but these bounds require specific forms of loss functions and cannot capture effects of heterogeneity.
\cite{huang2021fl} study the generalization of FedAvg on wide two-layer ReLU networks with homogeneous data.
\cite{collins2022fedavg} studies FedAvg under multi-task linear representation learning setting. In \cite{sun2024understanding}, the authors have demonstrated the impact of data heterogeneity on the generalization performance of some popular FL algorithms.

\section{Theory Notation and Preliminaries}
\label{sec:appendix_prelim}

We follow a similar notation as \cite{kou2023benign} in most of the analysis.

\captionsetup[table]{labelfont={color=black},font={color=black}}

\begin{table}[h!]
\centering
\caption{Summary of notation}
\color{black}
\begin{tabular}{c|l}
\hline \\
Symbol & Description\\ [0.05cm]
\hline\\
$j \in \{-1, 1\}$ & Layer index\\
$m$ & Number of filters \\
$d$ & Dimension of filter \\
$r \in [m]$ & Filter Index\\
$K$ & Number of clients \\
$k \in [K]$ & Client index \\
$N$ & Number of datapoints at each client \\
$i \in [N]$ & Datapoint index \\
$n = KN$ & Global dataset size \\
$y_{k,i} \in \{1,-1\}$ & Label of $i$-th datapoint at $k$-th client \\
$\bmu$ & Signal vector \\
$\sigma_p^2$ & Variance of Gaussian noise \\
$\bxi_{k,i}$ & Noise vector for $k$-th client and $i$-th datapoint \\
$\eta$ & Local learning rate \\
$\tau$ & Number of local steps \\
$\ell(z) = \log(1+\exp(-z))$ & Cross-entropy loss function \\
$\sigma(z) = \max(0,z)$ & ReLU function \\
$\sigma'(z) = \ind{z \geq 0}$ & Derivative of ReLU function \\
$t$ & Round index \\
$s$ & Iteration index \\
$h$ & Heterogeneity parameter \\
$\snr := \nicefrac{\normt{\bmu}}{\sigma_p \sqrt{d}}$ & Signal to Noise Ratio \\
\midrule
$\lm^{(\cdot,\cdot)}$ & Parameterized weights of the $k$-th client \\
$\lmweight^{(\cdot,\cdot)}$ & $(j,r)$-th filter weight of the $k$-th client\\
$\lgam^{(\cdot,\cdot)}$ & Local signal co-efficient for $k$-th client \\
$\lrho^{(\cdot,\cdot)}$ & Local noise coefficient for $k$-th client and $i$-th datapoint \\
$\lprho^{(\cdot,\cdot)}$ & Positive local noise coefficient for $k$-th client and $i$-th datapoint \\
$\lnrho^{(t,s)}$ & Negative local noise coefficient for $k$-th client and $i$-th datapoint \\
$\lderiv^{(\cdot,\cdot)}$ & Shorthand for $-1/\left(1+\exp(y_{k,i}f(\lm^{(\cdot,\cdot)},\bx_{k,i})\right)$ which is the \\
& derivative of cross-entropy loss for $i$-th datapoint at $k$-th client \\
$\gm^{(\cdot)}$ & Parameterized weight vector of the global model\\
$\gmweight^{(\cdot)}$ & $j,r$-th filter weight of the global model \\
$\ggam^{(\cdot)}$ & Global signal co-efficient \\
$\grho^{(\cdot)}$ & Global noise coefficient for $(k,i)$-th datapoint \\
$\gprho^{(\cdot)}$ & Positive global noise coefficient for $(k,i)$-th datapoint \\
$\gnrho^{(\cdot)}$ & Negative global noise coefficient for $(k,i)$-th client datapoint \\
\bottomrule
\end{tabular}
\end{table}

\subsection{Local Model Update}
Using local GD updates in \eqref{eq:localGD} to minimize the local loss function in \eqref{eq:global_obj}, the local model update for the $(j,r)$ filter at client $k$ in round $t$ can be written as,

\begin{align}
    \lmweight^{(t,\tau)} &= \gmweight^{(t)} -\frac{\eta}{Nm}\sum_{s=0}^{\tau-1}\sum_{i \in [N]} \lderiv^{(t,s)}\cdot\noisederiv{\lmweight^{(t,s)}}\cdot j y_{k,i}\bxi_{k,i} \nonumber \\
    & \hspace{10pt} - \frac{\eta}{Nm}\sum_{s=0}^{\tau-1}\sum_{i \in [N]} \lderiv^{(t,s)}\cdot\signalderiv{\lmweight^{(t,s)}}\cdot j \bmu \nonumber \\
    & = \gmweight^{(t)} + j\lgam^{(t,\tau)}\cdot\normt{\bmu}^{-2}\cdot\bmu + \sum_{i \in [N]} \lrho^{(t,\tau)} \cdot \normt{\bxi_{k,i}}^{-2}\cdot \bxi_{k,i} \label{eq:w_client_decomp_sig_noise}
\end{align}
where, we use $\lmweight^{(t,0)} \triangleq \gmweight^{(t)}$. Further, we define
\begin{align}
    \lgam^{(t,\tau)} & \triangleq -\frac{\eta}{Nm}\sum_{s=0}^{\tau-1}\sum_{i \in [N]} \lderiv^{(t,s)}\cdot\signalderiv{\lmweight^{(t,s)}}\cdot\norm{\bmu}, 
    \label{eq:gamma_jrk} \\
    \lrho^{(t,\tau)} & \triangleq -\frac{\eta}{Nm}\sum_{s=0}^{\tau-1}\lderiv^{(t,s)}\cdot\noisederiv{\lmweight^{(t,s)}}\cdot\norm{\bxi_{k,i}}\cdot jy_{k,i}. \label{eq:rho_jrki}
\end{align}
which respectively, denote the local signal $(\lgam^{(t,\tau)})$ and local noise $(\{\lrho^{(t,\tau)}\}_i)$ components of $\lmweight^{(t,\tau)}$.
We also define $\lprho^{(t,\tau)} = \lrho^{(t,\tau)} \ind{\lrho^{(t,\tau)} \geq 0}$ and $\lnrho^{(t,\tau)} = \lrho^{(t,\tau)} \ind{\lrho^{(t,\tau)} < 0}$, where $\ind{\cdot}$ denotes the indicator function, and which can alternatively be written as

\begin{align}
    \lprho^{(t,\tau)} =  -\frac{\eta}{Nm}\sum_{s=0}^{\tau-1}\lderiv^{(t,s)}\cdot\noisederiv{\lmweight^{(t,s)}}\cdot\norm{\bxi_{k,i}}\cdot \ind{y_{k,i} = j}, \label{eq:overbar_rho_jrki} \\
    \lnrho^{(t,\tau)} =  \frac{\eta}{Nm}\sum_{s=0}^{\tau-1}\lderiv^{(t,s)}\cdot\noisederiv{\lmweight^{(t,s)}}\cdot\norm{\bxi_{k,i}}\cdot \ind{y_{k,i} = -j}. \label{eq:underbar_rho_jrki}
\end{align}

\subsection{\texorpdfstring{Proof of \Cref{prop:decomposition}}{Proof of Proposition 1}}

The global model update at round $t+1$ can be written as
\begin{align}
    \gmweight^{(t+1)} &= \sum_{k=1}^K \frac{1}{K} \lmweight^{(t,\tau)} \nonumber\\
    & = \gmweight^{(t)} + \frac{j}{K} \sum_{k=1}^K \lgam^{(t,\tau)} \cdot\normt{\bmu}^{-2}\cdot\bmu + \sum_{k=1}^K \sum_{i \in [N]} \frac{1}{K} \lrho^{(t,\tau)} \cdot \normt{\bxi_{k,i}}^{-2}\cdot \bxi_{k,i}. \label{eq:w_server_decomp_sig_noise_1}
\end{align}
Mimicking the signal-noise decomposition in \eqref{eq:w_client_decomp_sig_noise}, we can define a similar decomposition for the global model as follows. 
\begin{align}
    \gmweight^{(t)} = \gmweight^{(0)} + j\ggam^{(t)}\cdot\normt{\bmu}^{-2}\cdot \bmu + \sum_{k=1}^K\sum_{i \in [N]}\grho^{(t)}\cdot\normt{\bxi_{k,i}}^{-2}\cdot\bxi_{k,i}.
\label{eq:gmweight_decomp}
\end{align}

\subsection{Co-efficient Update Equations}

Comparing with \eqref{eq:w_server_decomp_sig_noise_1}, we have the following recursive update for the global signal and noise coefficients using $n=KN$.
\begin{align}
    \ggam^{(t+1)} &= \ggam^{(t)} + \sum_{k=1}^K\frac{1}{K}\lgam^{(t,\tau)} \nonumber\\
    & = \ggam^{(t)} - \frac{\eta}{nm}\sum_{k=1}^K \sum_{i \in [N]} \sum_{s=0}^{\tau -1} \lderiv^{(t,s)}\cdot\signalderiv{\lmweight^{(t,s)}}\cdot\norm{\bmu} \label{eq:Gamma_jr} 
\end{align}
\begin{align}
    \grho^{(t+1)} &= \grho^{(t)} + \frac{1}{K}\lrho^{(t,\tau)} \nonumber \\
    & = \grho^{(t)} - \frac{\eta}{nm}\sum_{s=0}^{\tau-1}\lderiv^{(t,s)}\cdot\noisederiv{\lmweight^{(t,s)}}\cdot\norm{\bxi_{k,i}}\cdot jy_{k,i}.
\label{eq:noise_update}
\end{align}
Analogously, we can also define the positive and negative global noise coefficients,
\begin{align}
    \gprho^{(t+1)} = \gprho^{(t)} - \frac{\eta}{nm}\sum_{s=0}^{\tau-1}\lderiv^{(t,s)}\cdot\noisederiv{\lmweight^{(t,s)}}\cdot\norm{\bxi_{k,i}}\ind{y_{k,i} = j} \label{eq:overbar_P_jrki}
\end{align}
and,
\begin{align}
    \gnrho^{(t+1)} = \gnrho^{(t)} + \frac{\eta}{nm}\sum_{s=0}^{\tau-1}\lderiv^{(t,s)}\cdot\noisederiv{\lmweight^{(t,s)}}\cdot\norm{\bxi_{k,i}} \ind{y_{k,i} = -j}. \label{eq:underbar_P_jrki}
\end{align}

\begin{lemma}(Measuring local and global signal coefficient)
\label{lemma:measure_signal_coeff}

From \eqref{eq:w_client_decomp_sig_noise}, it follows that
\begin{align}
    \langle \lmweight^{(t,s)} - \gmweight^{(t)}, y_{k,i}\bmu \rangle = jy_{k,i}\lgam^{(t,s)}.
\label{eq:measure_local_signal_coeff}
\end{align}
and from \eqref{eq:gmweight_decomp}, it follows that
\begin{align}
    \langle \gmweight^{(t)} - \gmweight^{(0)}, \bmu \rangle = j\ggam^{(t)}.
\label{eq:measure_global_signal_coeff}
\end{align}
\end{lemma}

Since $\{ \ggam^{(t)} \}_t$ are non-negative and non-decreasing in $t$, the global weights $\{ \bw_{j,r}^{(t)} \}_r$ become increasing aligned with the \textit{actual} signal $y_{k,i} \bmu$ corresponding to the filters $j = y_{k,i}$. 
Similarly, as $\{ \lgam^{(t,s)} \}_t$ are non-negative and non-decreasing in $s$ for fixed $t$, the local weights $\{ \bw_{y_{k,i},r,k}^{(t,s)} \}_r$ become increasing aligned with the signal $y_{k,i} \bmu$ corresponding to the filters $j=y_{k,i}$.

\section{Training Error Convergence of FedAvg with Random Initialization}
\label{sec:assumptions_and_results}

For the sake of completeness, we state the conditions used in our analysis (\Cref{assum:main_assump}) in full detail.

\paragraph{Assumptions.}

Let $\epsilon$ be a desired training error threshold and $\delta \in (0,1)$ be some failure probability. Let $T^* = \frac{1}{\eta} \mathrm{poly}(\epsilon^{-1},m,n,d)$ be the maximum admissible rounds. 

Suppose there exists a sufficiently large constant $C$, such that the following hold.

\begin{assumption}
\label{assump:d}
    Dimension $d$ is sufficiently large, i.e.,
    \begin{align}
         d \geq C \max \left\{\mfrac{n\norm{\bmu}\log(T^*\tau)}{\sigma_p^2}, n^2\log(nm/\delta)(\log(T^*\tau))^2 \right\}. \nn
    \end{align}
\end{assumption}

\begin{assumption}
\label{assump:m_n}
    Training sample size $n$ and neural network width $m$ satisfy
    \begin{align}
        m \geq C \log (n/\delta), n \geq C \log (m/\delta). \nn
    \end{align}
\end{assumption}

\begin{assumption}
\label{assump:mu_strength}
    The norm of the signal satisfies,
    \begin{align}
        \norm{\bmu} \geq C \sigma_p^2 \log(n/\delta). \nn
    \end{align}
\end{assumption}

\begin{assumption}
\label{assump:sigma_0}
    Standard deviation of Gaussian initialization is sufficiently small, i.e.,
    \begin{align}
        \sigma_0 \leq \mfrac{1}{C}\min\left\{\mfrac{\sqrt{n}}{ \sigma_p d \tau}, \mfrac{1}{\sqrt{\log(m/\delta)}\normt{\bmu}}\right\}. \nn
    \end{align}
\end{assumption}

\begin{assumption}
\label{assump:eta}
    Learning rate is sufficiently small, i.e., 
    \begin{align}
        \eta \leq \mfrac{1}{C} \min \left\{ \mfrac{nm\sqrt{\log (m/\delta)}}{\sigma_p^2 d}, \mfrac{1}{\norm{\bmu}}, \mfrac{1}{\sigma_p^2d}  \right\}. \nn
    \end{align}
\end{assumption}

The assumptions are primarily used to ensure that the model is sufficiently overparameterized, i.e., training loss can be made arbitrarily small, and that we do not begin optimization from a point where the gradient is already zero or unbounded. We provide a more intuitive reasoning behind each of the assumptions below:
\begin{itemize}
\item \textit{Bounded number of communication rounds:}
This is needed to ensure that the magnitude of filter weights remains bounded throughout training since they grow logarithmically with the number of updates (see Theorem 3). We note that this is quite a mild condition since the max rounds can have polynomial dependence on $1/\epsilon$ where $\epsilon$ is our desired training error. 
\item \textit{Dimension $d$ is sufficiently large:} 
This is needed to ensure that the model is sufficiently overparameterized and the training loss can be made arbitrarily small. Recall that our input $\mathbf{x}$ consists of a signal component $\bmu\in \mathbb{R}^d$ that is common across all datapoints and noise component $\bxi \in \mathbb{R}^d$ that is independently drawn from $\mathcal{N}(0,\sigma_p^2\cdot \bm{I})$. Having a sufficiently large $d$ ensures that the correlation between any two noise vectors, i.e. $\langle \bxi, \bxi' \rangle/ \|\bxi\|^2$ is not too large. Otherwise if the correlation between two noise vectors is large and negative, then minimizing the loss on one data point could end up increasing the loss on another training point which complicates convergence and prevents loss from becoming arbitrarily small.
\item \textit{Training set size and network width is sufficiently large:}
The condition ensures that a sufficient number of filters get activated at initialization with high probability (see Lemma 6 and Lemma 7) and prevents cases where the initial gradient is zero. The condition on training set size also ensures that there are a sufficient number of datapoints with negative and positive labels (see Lemma 8). 
\item \textit{Standard deviation of Gaussian random initialization is sufficiently small:} This condition is needed to ensure that the magnitude of the initial correlation between the filter weights and the signal and noise components, i.e $|\langle \mathbf{w}_{j,r}^{(0)}, \bmu \rangle|$, $|\langle \mathbf{w}_{j,r}^{(0)}, \bxi \rangle|$ is not too large. This simplifies the analysis and prevents cases where none of the filters get activated at initialization (see Lemma 21). It also ensures that after some number of rounds all filters get aligned with the signal (see Lemma 30). 
\item \textit{Norm of signal is larger than noise variance:} This condition is needed to ensure that all misaligned filters at initialization eventually become aligned with the signal after some rounds (see Lemma 30). This allows us to derive a meaningful bound on test performance that is not dominated by noise memorization.  
\item \textit{Learning rate is sufficiently small:} This is a standard condition to ensure that gradient descent does not diverge. The conditions are derived from ensuring that the signal and noise coefficient remain bounded in the first stage of training and that the loss decreases monotonically in every round in the second stage of training. 
\end{itemize}

For ease of reference, we restate \Cref{thm:train_loss} below.

\begin{thm}[Training Loss Convergence]
Let $T_1 = \bigO{\frac{mn}{\eta\sigma_p^{2}d\tau}}$. With probability $1-\delta$ over the random initialization, for all $T_1 \leq T \leq T^*$ we have,
\begin{align}
    \frac{1}{T-T_1 + 1}\sum_{t=T_1}^T L(\bW^{(t)}) \leq \frac{\norm{\bW^{(T_1)}-\bW^*}}{\eta(T-T_1 + 1)} + \epsilon. \nn
\end{align}
Therefore we can find an iterate with training error smaller than $2\epsilon$ within $T = T_1 + \norm{\bW^{(T_1)}-\bW^*}/(\eta \epsilon) = \bigO{\frac{mn}{\eta\sigma_p^{2}d\tau}} + \bigO{\frac{mn \log(\tau/\epsilon)}{\eta\sigma_p^{2}d\epsilon} }$ rounds.
\end{thm}

\paragraph{Proof Sketch.}

The template follows that of \cite{kou2023benign} and is divided into $3$ parts. In the first part (\Cref{subsec:scale_coeff}), we show that the magnitude of the signal and noise memorization coefficients for the global model is bounded for the entire duration of training (see \Cref{thm:coeff_bound}), where $|\ggam^{(t)}| \leq 4\log(T^*\tau)$ and $|\grho^{(t)}| \leq 4\log(T^*\tau)$ for all $0\leq t \leq T^*-1$. Next, we divide our training into two stages. 
In the first stage (\Cref{subsec:first_stage}), we show (see \Cref{lemma:activated_noise_filters}) that the noise (and also signal) memorization coefficients grow fast and are lower bounded by some constant after $T_1$ rounds i.e., $|\gprho^{(T_1)}| = \bigw{1}$. 
In the second stage (\Cref{subsec:2nd_stage}), the growth of the noise and signal coefficients becomes relatively slower and the model reaches a neighborhood of a global minimizer where the loss landscape is nearly convex (see \Cref{lemma:pl_convex_prop}). 
Using this we can show that our objective is monotonically decreasing in every round (see \Cref{lemma:local_model_convergence}), which establishes convergence (in \Cref{subsec:proof_train_loss}). 
We begin by stating (in \Cref{subsec:prelim_lemma}) some intermediate results that we use in the subsequent analysis.

\subsection{Preliminary Lemmas}
\label{subsec:prelim_lemma}

\begin{lemma}
\label{lemma:noise_corr}
    (Lemma B.4 in \cite{cao2022benign}) Suppose that $\delta > 0$ and $d = \bigw{\log (4n/\delta)}$. Then with probability at least $1-\delta$,
    \begin{align}
        \sigma_p^2d/2 \leq \norm{\bxi_{k,i}} \leq 3\sigma_p^2d/2, \nn
    \end{align}
    \begin{align}
        \abs{\inner{\bxi_{k,i}}{\bxi_{k',i'}}} \leq 2\sigma_p^2 \sqrt{d \log (6n^2/\delta)}, \nn
    \end{align}
for all $k,k' \in [K]$, $ i,i'\in [N]$, and $(k,i) \neq (k',i')$.
\end{lemma}

\begin{lemma}
\label{lemma:init_filter_corr}
    (Lemma B.5 in \cite{kou2023benign}). Suppose that $d = \bigw{\log (mn/\delta)}$, $m = \bigw{\log (1/\delta)}$. Then with probability at least $1-\delta$,
\begin{align}
        \sigma_0^2d/2 \leq \norm{\gmweight^{(0)}} \leq 3\sigma_0^2 d/2, \nn
\end{align}
\begin{align}
    \abs{\inner{\gmweight^{(0)}}{\bmu}} & \leq \sqrt{2 \log (12m/\delta)} \cdot \sigma_0 \normt{\bmu},  \nn
    \abs{\inner{\gmweight^{(0)}}{\bxi_{k,i}}} & \leq 2\sqrt{\log (12mn/\delta)} \cdot \sigma_0 \sigma_p \sqrt{d}, \nn
\end{align}
for all $r \in [m]$, $j \in \{\pm 1\}$, $k \in [K]$ and $i \in [N]$. 

\end{lemma}

\begin{lemma}
\label{lemma:min_size_activated_noise_filters} (Lemma B.6 in \cite{kou2023benign}). 
    Let $S_{k,i}^{(0)} = \left\{ r \in [m]: \inner{\bw_{y_{k,i},r}^{(0)}}{\bxi_{k,i}} \geq 0\right\}$. Suppose $\delta > 0$ and $m \geq 50 \log (2n/\delta)$. Then with probability at least $1-\delta$,
    \begin{align}
        \left|S_{k,i}^{(0)}\right| \geq 0.4m, \forall i \in [n]. \nn
    \end{align}
\end{lemma}

\begin{lemma}
\label{lemma:min_size_activated_noise_data} 
(Lemma B.7 in \cite{kou2023benign}) Let $\tilde{S}_{j,r}^{(0)} = \left\{ k \in [K], i \in [N]: y_{k,i} = j, \inner{\bw_{j,r}^{(0)}}{\bxi_{k,i}} \geq 0\right\}$. Suppose $\delta > 0$ and $n \geq 32 \log (4m/\delta)$. Then with probability at least $1-\delta$,
    \begin{align}
    \left|\tilde{S}_{j,r}^{(0)}\right| \geq n/8, \forall i \in [n]. \nn
    \end{align}
\end{lemma}

\begin{lemma}
\label{lemma:S_j_size}

Let $D_j = \left\{ k \in [K], i \in [N]: y_{k,i} = j \right\}$. Suppose $\delta > 0$ and $n \geq 8 \log (4/\delta)$. Then with probability at least $1-\delta$,
\begin{align}
    \abs{D_j} \geq \frac{n}{4}, \forall j \in \{\pm 1\}. \nn
\end{align}
\end{lemma}

\begin{proof}
We have $\abs{D_j} = \sum_{k,i} \ind{y_{k,i} = j}$ and therefore $\expt \abs{D_j} = \sum_{k,i} \mathbb{P} (y_{k,i} = j) = n/2$. Applying Hoeffding's inequality we have with probability $1-2\delta$,
\begin{align}
    \abs{\frac{\abs{D_j}}{n} - \frac{1}{2}} \leq \sqrt{\frac{\log (4/
    \delta)}{2n}}. \nn
\end{align}
Now if $n \geq 8\log(4/\delta)$, by applying union bound, we have with probability at least $1-\delta$,
\begin{align}
    \abs{D_j} \geq \frac{n}{4}, \forall j \in \{ \pm 1 \}. \nn
\end{align}
\end{proof}

\subsection{Bounding the Scale of Signal and Noise Memorization Coefficients}
\label{subsec:scale_coeff}

Our first goal is to show that the coefficients of the global model, i.e., $\Gamma_{j,r}^{(t)}$, $ \gprho^{(t)}$ and $\abs{ \gnrho^{(t)} }$ are bounded as $\bigo{\log (T^*\tau})$. To do so, we look at a \textit{virtual} iteration index given by $v= 0,1,2,3,\dots, T^*\tau-1$. For any $v$, we can define the filter weights at virtual iteration $v$ in terms of the filter weights we have seen so far. In particular,
\begin{align}
    \lmweightt^{(v)} \triangleq \lmweight^{\left(\floor{\frac{v}{\tau}}, v\bmod \tau\right)}. \nn
\end{align}

We also define the following \textit{virtual sequence of local coefficients} which will be used in our proof. Let $\ggamv^{(0)} = 0, \gprhov^{(0)} = 0, \gnrhov^{(0)} = 0$. We have the following update equation for $\ggamv^{(v)}, \gprhov^{(v)}$ and $\gnrhov^{(v)}$ for $v \geq 1$.

\begin{align}
    \label{eq:ggamv_update}
    \ggamv^{(v)} = 
    \begin{cases}
        \ggamv^{(v-1)} - \frac{\eta}{Nm}\displaystyle\sum_{i \in [N]} \lderiv^{(v-1)} \signalderiv{\lmweightt^{(v-1)}}\norm{\bmu}, \text{ if } v\pmod\tau & \neq 0, \\
        \ggamv^{(v-\tau)} - \frac{\eta}{nm} \displaystyle\sum_{s=0}^{\tau-1} \sum_{k'} \sum_{i \in [N]} \lderivk^{(v-\tau+s)} \signalderiv{\lmweightt^{(v-\tau+s)}}\norm{\bmu} & \text{ else,}
    \end{cases}
\end{align}
where we slightly abuse notation, using $\lderiv^{(v)}$ to denote $\lderiv^{\left(\floor{\frac{v}{\tau}}, v\bmod \tau\right)}$.

\begin{align}
\label{eq:gprhov_update}
   \gprhov^{(v)} =
   \begin{cases}
       \gprhov^{(v-1)} - \frac{\eta}{Nm} \lderiv^{(v-1)} \noisederiv{\lmweightt^{(v-1)}} \norm{\bxi_{k,i}}\ind{j = y_{k,i}}, \text{ if } v\pmod\tau & \neq 0, \\
       \gprhov^{(v-\tau)} - \frac{\eta}{nm} \displaystyle \sum_{s = 0}^{\tau-1} \lderiv^{(v-\tau + s)} \noisederiv{\lmweightt^{(v-\tau + s)}} \norm{\bxi_{k,i}} \ind{j = y_{k,i}} & \text{ else.}
   \end{cases}
\end{align}

\begin{align}
\label{eq:gnrhov_update}
   \gnrhov^{(v)} = 
   \begin{cases}
       \gnrhov^{(v-1)} + \frac{\eta}{Nm} \lderiv^{(v-1)} \noisederiv{\lmweightt^{(v-1)}} \norm{\bxi_{k,i}}\ind{j = - y_{k,i}}, \text{ if } v\pmod\tau & \neq 0, \\
       \gnrhov^{(v-\tau)} + \frac{\eta}{nm} \displaystyle \sum_{s = 0}^{\tau-1} \lderiv^{(v-\tau + s)} \noisederiv{\lmweightt^{(v-\tau + s)}} \norm{\bxi_{k,i}} \ind{j = - y_{k,i}} & \text{ else.}
   \end{cases}
\end{align}

Note that we have the relation $\ggamv^{(t\tau)} = \ggam^{(t)}, \gprhov^{(t\tau)} = \gprho^{(t)}, \gnrhov^{(t\tau)} = \gnrho^{(t)}$ 

for all $t = 0,1,2,\dots, T^*-1$. Intuitively, if we can bound the virtual sequence of coefficients, we can also bound the actual coefficients of the global model at every round. 

\subsubsection{Decomposition of Virtual Local Filter Weights}

The purpose of introducing the virtual sequence of coefficients is to write the local filter weight at each client as the following decomposition.

\begin{align}
\label{eq:lm_decomp}
    \lmweightt^{(v)} &= \gmweight^{(0)} + j\ggamv^{(v)}\normt{\bmu}^{-2}\bmu + \sum_{k',k' \neq k}\sum_{i' \in [N]} (\overline{\mathbb{P}}_{j,r,k',i'}^{(\tau \floor{v/\tau})} +  \underline{\mathbb{P}}_{j,r,k',i'}^{(\tau \floor{v/\tau})}) \normt{\bxi_{k',i'}}^{-2}\bxi_{k',i'} \nonumber\\
    & \hspace{10pt} + \sum_{i \in [N]}  (\gprhov^{(v)} +  \gnrhov^{(v)}) \normt{\bxi_{k,i}}^{-2}\bxi_{k,i}. 
\end{align}

Note that $(\tau \floor{v/\tau})$ denotes the last iteration at which communication happened. If $ v \pmod \tau = 0$, then $\lmweightt^{(v)}$ is the same for all $k \in [K]$.

\subsubsection{Theorem on Scale of Coefficients}

We will now state the theorem that bounds our virtual sequence of coefficients and give the proof below. We first define some quantities that will be used throughout the proof.
\begin{align}
    \alpha := 4 \log (T^*\tau); \hspace{2pt} \beta := 2 \max_{i,j,k,r}\left\{ \abs{\inner{\gmweight^{(0)}}{\bmu}},  \abs{\inner{\gmweight^{(0)}}{\bxi_{k,i}}} \right\}; \hspace{2pt} \effsnr = \frac{n \norm{\bmu}}{\sigma_p^2 d}. \nn
\end{align}

\begin{theorem}
\label{thm:coeff_bound} 
    Under assumptions,  for all $v = 0,1,2,\dots, T^*\tau-1$, we have that,
    \begin{align}
       \ggamv^{(0)} = 0, \gprhov^{(0)} = 0, \gnrhov^{(0)} = 0, \nn
    \end{align}
    \begin{align}
        0 \leq \gprhov^{(v)} \leq \alpha,
    \label{eq:rhop_bound}
    \end{align}
    \begin{align}
        0 \geq \gnrhov^{(v)} \geq -\beta -8\sqrt{\frac{\log (6n^2/\delta)}{d}}n\alpha \geq -\alpha,
     \label{eq:rhov_bound}
    \end{align}
    \begin{align}
        0 \leq \ggamv^{(v)} \leq C'\effsnr\alpha,
     \label{eq:gamma_bound}
    \end{align}
for all $r \in [m], j \in \{\pm 1\}, k \in [K], i \in [N]$, where $C'$ is some positive constant.
\end{theorem}

We will use induction to prove this theorem. The statement is clearly true at $v = 0$. Now assuming the statement holds at $v = v'$ we will show that it holds at $v = v'+1$. We first state and prove some intermediate lemmas that we will use in our proof.

\subsubsection{Intermediate Steps to Prove the Induction in \Cref{thm:coeff_bound}}

\begin{lemma}
\label{lemma:beta_mag}
\begin{align}
     \max \left\{\beta, 4\sqrt{\frac{\log (6n^2/\delta)}{d}}n\alpha\right\} \leq \frac{1}{12}.\nn
\end{align}
\end{lemma}

\begin{proof}
From  \Cref{lemma:init_filter_corr} we have $\beta = 4\sigma_0 \max\left\{ \sqrt{\log(12mn/\delta)} \cdot \sigma_p \sqrt{d}, \sqrt{\log (12m/\delta)} \cdot \normt{\bmu}\right\}$. Now from Assumptions \ref{assump:d} and \ref{assump:sigma_0}, by choosing $C$ large enough, the inequality is satisfied.
\end{proof}

\begin{lemma}
\label{lemma:inner_prod_bound}
Suppose, \eqref{eq:rhop_bound}, \eqref{eq:rhov_bound} and  \eqref{eq:gamma_bound} holds for all iterations $0 \leq v \leq v'$. Then for all $r \in [m]$, $j \in \{\pm 1\}, k \in [K], i \in [N]$ we have,

\begin{align}
    \inner{\lmweightt^{(v')} - \gmweight^{(0)}}{\bmu} &= j\ggamv^{(v')},
    \label{eq:mu_inner_prod_bound} \\
    \abs{\inner{\lmweightt^{(v')} - \gmweight^{(0)}}{\bxi_{k,i}} - \gprhov^{(v')}} & \leq 4\sqrt{\frac{\log (6n^2/\delta)}{d}}n\alpha, j =  y_{k,i}, \label{eq:rhop_inner_prod_bound} \\
    \abs{\inner{\lmweightt^{(v')} - \gmweight^{(0)}}{\bxi_{k,i}} - \gnrhov^{(v')}} & \leq 4\sqrt{\frac{\log (6n^2/\delta)}{d}}n\alpha, j \neq y_{k,i}.
    \label{eq:rhon_inner_prod_bound}
\end{align}
\end{lemma}

\begin{proof}[Proof of \eqref{eq:mu_inner_prod_bound}]
It follows directly from \eqref{eq:lm_decomp} by using our assumption that $\inner{\bmu}{\bxi_{k,i}} = 0$ for all $k \in [K], i \in [N]$.
\end{proof}

\begin{proof}[Proof of \eqref{eq:rhop_inner_prod_bound}]
Note that 

for $y_{k,i} = j$ we have $\gnrhov^{(v')} = 0$. Now using \eqref{eq:lm_decomp} for $j = y_{k,i}$ we have,
\begin{align}
    & \abs{\inner{\lmweightt^{(v')} - \gmweight^{(0)}}{\bxi_{k,i}} - \gprhov^{(v')}} \nonumber \\
    &= \abs{\sum_{k',k' \neq k}\sum_{i' \in [N]} (\overline{\mathbb{P}}_{j,r,k',i'}^{(\tau \floor{v'/\tau})} +  \underline{\mathbb{P}}_{j,r,k',i'}^{(\tau \floor{v'/\tau})})\tfrac{\inner{\bxi_{k,i}}{\bxi_{k',i'}}}
    {\norm{\bxi_{k',i'}}}  +  \sum_{i' \in [N], i' \neq i}  (\overline{\mathbb{P}}_{j,r,k,i'}^{(v')} +  \underline{\mathbb{P}}_{j,r,k,i'}^{(v')}) \tfrac{\inner{\bxi_{k,i}}{\bxi_{k,i'}}}
    {\norm{\bxi_{k,i'}}}} \nonumber \\
    %%%%
    & \overset{(a)}{\leq}\left( \sum_{k',k' \neq k}\sum_{i' \in [N]} \left( \abs{\overline{\mathbb{P}}_{j,r,k',i'}^{(\tau \floor{v'/\tau})}} + \abs{\underline{\mathbb{P}}_{j,r,k',i'}^{(\tau \floor{v'/\tau})}} \right) + \sum_{i' \in [N]} \left( \abs{\overline{\mathbb{P}}_{j,r,k,i'}^{(v')}} + \abs{\underline{\mathbb{P}}_{j,r,k,i'}^{(v')}} \right) \right)4\sqrt{\tfrac{\log (6n^2/\delta)}{d}} \nonumber \\
    & \overset{(b)}{\leq} 4\sqrt{\mfrac{\log (6n^2/\delta)}{d}}n\alpha, \nn
\end{align}
where $(a)$ follows from triangle inequality and \Cref{lemma:noise_corr}; $(b)$ follows from the induction hypothesis.
\end{proof}

\begin{proof}[Proof of \eqref{eq:rhon_inner_prod_bound}]
Note that for

$j \neq y_{k,i} $ we have $\gprhov^{(v')} = 0$. Using \eqref{eq:lm_decomp} for $j \neq y_{k,i}$ we have,
\begin{align}
    & \abs{\inner{\lmweightt^{(v')} - \gmweight^{(0)}}{\bxi_{k,i}} - \gnrhov^{(v')}} \nonumber \\
    &= \abs{\sum_{k',k' \neq k}\sum_{i' \in [N]} (\overline{\mathbb{P}}_{j,r,k',i'}^{(\tau \floor{v'/\tau})} +  \underline{\mathbb{P}}_{j,r,k',i'}^{(\tau \floor{v'/\tau})})\tfrac{\inner{\bxi_{k,i}}{\bxi_{k',i'}}}
    {\norm{\bxi_{k',i'}}}  +  \sum_{i' \in [N], i' \neq i}  (\overline{\mathbb{P}}_{j,r,k,i'}^{(v')} +  \underline{\mathbb{P}}_{j,r,k,i'}^{(v')}) \tfrac{\inner{\bxi_{k,i}}{\bxi_{k,i'}}}
    {\norm{\bxi_{k,i'}}}} \nonumber \\
    & \overset{(a)}{\leq} \left( \sum_{k',k' \neq k}\sum_{i' \in [N]} \left( \abs{\overline{\mathbb{P}}_{j,r,k',i'}^{(\tau \floor{v'/\tau})}} + \abs{\underline{\mathbb{P}}_{j,r,k',i'}^{(\tau \floor{v'/\tau})}} \right) + \sum_{i' \in [N]} \left( \abs{\overline{\mathbb{P}}_{j,r,k,i'}^{(v')}} + \abs{\underline{\mathbb{P}}_{j,r,k,i'}^{(v')}} \right) \right)4\sqrt{\tfrac{\log (6n^2/\delta)}{d}} \nonumber \\
    & \overset{(b)}{\leq} 4\sqrt{\mfrac{\log (6n^2/\delta)}{d}}n\alpha, \nn
\end{align}
where $(a)$ follows from triangle inequality and \Cref{lemma:noise_corr}; $(b)$ follows from the induction hypothesis.

This concludes the proof of \Cref{lemma:beta_mag}.
\end{proof}

\begin{lemma}
\label{lemma:bounded_func_output}
Suppose \eqref{eq:rhop_bound}, \eqref{eq:rhov_bound} and  \eqref{eq:gamma_bound} hold at iteration $v'$. Then for all $k \in [K]$ and $i \in [N]$,
\begin{enumerate}[leftmargin=*]
    \item \label{bounded_func_output_1} For $j \neq y_{k,i}$,  $F_j(\widetilde{\bW}_{j,k}^{(v')}, \bx_{k,i}) \leq 0.5$. 
    
    \item \label{bounded_func_output_2} For $j = y_{k,i}$, $F_j(\widetilde{\bW}_{j,k}^{(v')}, \bx_{k,i}) \geq \frac{1}{m}\sum_{r=1}^m \gprhov^{(v')}  - 0.25$. 
    \item \label{bounded_func_output_3} $ y_{k,i}f(\lmt^{(v')}, \bx_{k,i}) \geq \frac{1}{m} \sum_{r=1}^m \gprhovy^{(v')} -0.75. $
\end{enumerate}
    
\end{lemma}

\begin{proof}[Proof of \ref{bounded_func_output_1}]
First note that for $j \neq y_{k,i}$ from \Cref{lemma:inner_prod_bound} we have,
\begin{align}
\label{eq:inner_prod_signal_neg}
    \inner{\lmweightt^{(v')}}{\bmu} \leq \inner{\gmweight^{(0)}}{\bmu}. 
\end{align}
since $\ggamv^{(v')} \geq 0$ by the induction hypothesis. Also from \Cref{lemma:inner_prod_bound} for $j \neq y_{k,i}$ we have,
\begin{align}
\label{eq:inner_prod_noise_neg}
    \inner{\lmweightt^{(v')}}{\bxi_{k,i}} & \leq   \inner{\gmweight^{(0)}}{\bxi_{k,i}} + \gnrhov^{(v')} + 4\sqrt{\frac{\log (6n^2/\delta)}{d}}n\alpha \nonumber \\
    & \overset{(a)}{\leq}  \inner{\gmweight^{(0)}}{\bxi_{k,i}} + 4\sqrt{\frac{\log (6n^2/\delta)}{d}}n\alpha
\end{align}
where $(a)$ follows from $\gnrhov^{(v')} \leq 0$ (induction hypothesis). Now using the definition of $F_j(\bW,\bx)$ for $j \neq y_{k,i}$ we have,
\begin{align}
\label{eq:max_negative_output}
    F_j(\widetilde{\bW}_{j,k}^{(v')}, \bx_{k,i}) &= \frac{1}{m}\sum_{r=1}^m \left[ \sigma\left(\inner{\lmweightt^{(v')}}{y_{k,i}\bmu} \right) + \sigma\left( \inner{\lmweightt^{(v')}}{\bxi_{k,i}} \right) \right] \nonumber\\
    & \overset{(a)}{\leq} 3 \max_{r \in [m]} \left\{ \abs{\inner{\gmweight^{(0)}}{\bmu}}, \abs{\inner{\gmweight^{(0)}}{\bxi_{k,i}} }, 4\sqrt{\frac{\log (6n^2/\delta)}{d}}n\alpha \right\} \nonumber\\
    & \overset{(b)}{\leq} 3 \max \left\{ \beta, 4\sqrt{\frac{\log (6n^2/\delta)}{d}}n\alpha \right\} \nonumber\\
    & \overset{(c)}{\leq} 0.5.  
\end{align}
Here $(a)$ follows from \eqref{eq:inner_prod_signal_neg} and \eqref{eq:inner_prod_noise_neg}; $(b)$ follows from the definition of $\beta$; $(c)$ follows from \Cref{lemma:beta_mag}.

\end{proof}

\begin{proof}[Proof of \ref{bounded_func_output_2}]
For $j = y_{k,i}$ we have,

\begin{align}
    F_j(\widetilde{\bW}_{j,k}^{(v')}, \bx_{k,i}) &= \frac{1}{m}\sum_{r=1}^m \left[ \sigma\left(\inner{\lmweightt^{(v')}}{y_{k,i}\bmu} \right) + \sigma\left( \inner{\lmweightt^{(v')}}{\bxi_{k,i}} \right) \right] \nonumber\\
    & \overset{(a)}{\geq} \frac{1}{m}\sum_{r=1}^m \left[ \inner{\lmweightt^{(v')}}{y_{k,i}\bmu} +  \inner{\lmweightt^{(v')}}{\bxi_{k,i}} \right] \nonumber\\
    & \overset{(b)}{\geq} \frac{1}{m}\sum_{r=1}^m \left[ \inner{\gmweight^{(0)}}{y_{k,i}\bmu} +  \inner{\gmweight^{(0)}}{\bxi_{k,i}} + \gprhov^{(v')} - 4\sqrt{\frac{\log (6n^2/\delta)}{d}}n\alpha\right] \nonumber\\
    & \overset{(c)}{\geq} \frac{1}{m}\sum_{r=1}^m \gprhov^{(v')} - 2\beta - 4\sqrt{\frac{\log (6n^2/\delta)}{d}}n\alpha 
    \nn \\
    & \overset{(d)}{\geq} \frac{1}{m}\sum_{r=1}^m \gprhov^{(v')}  - 0.25. \label{eq:min_positive_output}
\end{align}
Here $(a)$ follows from $\sigma(z) \geq z$; $(b)$ follows from \Cref{lemma:inner_prod_bound} and that $\ggamv^{(v')} \geq 0$; $(c)$ follows from the definition of $\beta$; $(d)$ follows from \Cref{lemma:beta_mag}.

\end{proof}

\begin{proof}[Proof of \ref{bounded_func_output_3}]
Combining the results in \eqref{eq:max_negative_output} and \eqref{eq:min_positive_output} we have,
\begin{align}
     y_{k,i}f(\lmt^{(v')}, \bx_{k,i}) &= F_{y_{k,i}}(\widetilde{\bW}_{y_{k,i},k}^{(v')},\bx_{k,i}) - F_{-y_{k,i}}(\widetilde{\bW}_{-y_{k,i},k}^{(v')},\bx_{k,i}) \nonumber\\
    & \overset{(a)}{\geq} F_{y_{k,i}}(\widetilde{\bW}_{y_{k,i},k}^{(v')},\bx_{k,i}) - 0.5 \nonumber \\
    & \overset{(b)}{\geq} \frac{1}{m}\sum_{r=1}^m \gprhovy^{(v')} -0.75. \nn
\end{align}
where $(a)$ follows from \eqref{eq:max_negative_output}; $(b)$ follows from \eqref{eq:min_positive_output}.

This concludes the proof of \Cref{lemma:bounded_func_output}.
\end{proof}

\begin{lemma}
\label{lemma:max_deriv}
Suppose \eqref{eq:rhop_bound},  \eqref{eq:rhov_bound} and \eqref{eq:gamma_bound} hold at iteration $v'$. Then for all $j \in \{\pm 1\}$, $k \in [K]$ and $i \in [N]$, $\abs{\lderiv^{(v')}} \leq \exp\left(-F_{y_{k,i}}(\widetilde{\bW}_{y_{k,i},k}^{(v')},\bx_i) + 0.5\right )$. 
\end{lemma}

\begin{proof}
We have,
\begin{align}
    \abs{\lderiv^{(v')}} 
    &= \frac{1}{1 + \exp\left(y_{k,i}\left[F_{+1}(\widetilde{\bW}_{+1,k}^{(v')},\bx_{k,i}) - F_{-1}(\widetilde{\bW}_{+1,k}^{(v')},\bx_{k,i})\right] \right)} \nonumber \\
    & \overset{(a)}{\leq} \exp\left( -y_{k,i}\left[F_{+1}(\widetilde{\bW}_{+1,k}^{(v')},\bx_{k,i}) - F_{-1}(\widetilde{\bW}_{+1,k}^{(v')},\bx_{k,i})\right]\right) \nonumber \\
    & = \exp\left( -F_{y_{k,i}}(\widetilde{\bW}_{y_{k,i},k}^{(v')},\bx_{k,i}) + F_{-y_{k,i}}(\widetilde{\bW}_{-y_{k,i},k}^{(v')},\bx_{k,i})\right) \nn \\
    & \overset{(b)}{\leq} \exp\left( -F_{y_{k,i}}(\widetilde{\bW}_{y_{k,i},k}^{(v')},\bx_{k,i}) +0.5 \right), \nn
\end{align}
where $(a)$ uses $1/(1+\exp(z)) \leq \exp(-z)$; $(b)$ uses part 1 of \Cref{lemma:bounded_func_output}.
\end{proof}

\begin{lemma}
\label{lemma:bounded_lderiv_frac}
Let $g(z) = \ell'(z) = -1/(1+\exp(z))$. Further suppose $z_2 - z_1 \leq c$ where $c \geq 0$. Then, 
\begin{align}
    \frac{g(z_1)}{g(z_2)} \leq \exp(c).
\end{align}

\end{lemma}

\begin{proof}
We have,
\begin{align}
    \frac{g(z_1)}{g(z_2)}  = \frac{1+\exp(z_2)}{1+\exp(z_1)} \leq \max\{1, \exp(z_2 - z_1)\} \overset{(a)}{\leq} \exp(c), \nn
\end{align}
where $(a)$ follows from $c \geq 0$. 
\end{proof}

\begin{lemma}
\label{lemma:min_and_max_noise_inner_prod}

Suppose \eqref{eq:rhop_bound},  \eqref{eq:rhov_bound} and \eqref{eq:gamma_bound} hold at iteration $v'$. Then for all $k \in [K]$ and $i \in [N]$, 
\begin{align}
\label{eq:min_noise_inner_prod}
    \inner{\widetilde{\bw}_{y_{k,i}, r, k}^{(v')}}{\bxi_{k,i}} \geq -0.25,
\end{align}
\begin{align}
\label{eq:max_noise_inner_prod}
    \inner{\widetilde{\bw}_{y_{k,i}, r,k}^{(v')}}{\bxi_{k,i}} \leq \sigma\left(\inner{\widetilde{\bw}_{y_{k,i},r,k}^{(v')}}{\bxi_{k,i}} \right) \leq \inner{\widetilde{\bw}_{y_{k,i},r,k}^{(v')}}{\bxi_{k,i}} + 0.25.
\end{align}
\end{lemma}

\begin{proof}[Proof of \eqref{eq:min_noise_inner_prod}]
From \Cref{lemma:inner_prod_bound} we have,
\begin{align}
  \inner{\widetilde{\bw}_{y_{k,i},r,k}^{(v')}}{\bxi_{k,i}} &\geq \inner{\bw_{y_{k,i},r,k}^{(0)}}{\bxi_{k,i}} + \overline{\mathbb{P}}_{y_{k,i},r,k,i}^{(v')} - 4\sqrt{\frac{\log (6n^2/\delta)}{d}}n\alpha \nn \\
  & \overset{(a)}{\geq} -\beta - 4\sqrt{\frac{\log (6n^2/\delta)}{d}}n\alpha \nonumber\\
  & \overset{(b)}{\geq} -0.25. \nn
\end{align}
Here $(a)$ follows from the definition of $\beta$ and $\overline{\mathbb{P}}_{y_{k,i},r,k,i}^{(v')} \geq 0$ for all $v' \geq 0$; $(b)$ follows from \Cref{lemma:beta_mag}.
\end{proof}

\begin{proof}[Proof of \eqref{eq:max_noise_inner_prod}]
The first inequality of \eqref{eq:max_noise_inner_prod} follows naturally since $\sigma(z) \geq z$ for all $z \in \mathbb{R}$. For the second inequality we have,
\begin{align}
   \sigma \left( \inner{\widetilde{\bw}_{y_{k,i},r,k}^{(v')}}{\bxi_{k,i}} \right) = 
   \begin{cases}
       \inner{ \widetilde{\bw}_{y_{k,i},r,k}^{(v')}}{\bxi_{k,i}}  \leq \inner{\widetilde{\bw}_{y_{k,i},r,k}^{(v')}}{\bxi_{k,i}} + 0.25, & \text{ if } \inner{\widetilde{\bw}_{y_{k,i},r,k}^{(v')}}{\bxi_{k,i}} \geq 0 \\
       0 \overset{(a)}{\leq} \inner{\widetilde{\bw}_{y_{k,i},r,k}^{(v')}}{\bxi_{k,i}} + 0.25, & \text{ if } \inner{\widetilde{\bw}_{y_{k,i},r,k}^{(v')}}{\bxi_{k,i}} <0,
   \end{cases} \nn
\end{align}
where $(a)$ follows from  $\inner{\widetilde{\bw}_{y_{k,i},r,k}^{(v')}}{\bxi_{k,i}} \geq - 0.25$. This completes the proof.

This concludes the proof of \Cref{lemma:min_and_max_noise_inner_prod}.
\end{proof}

\begin{lemma}
\label{lemma:bounded_func_diff}

Suppose \eqref{eq:rhop_bound},  \eqref{eq:rhov_bound} and \eqref{eq:gamma_bound} hold at iteration $v'$. Then for all $k,k' \in [K]$ and $i, i' \in [N]$,
\begin{align}
    \abs{y_{k,i}f(\lmt^{(v')}, \bx_{k,i})-  y_{k',i'}f(\widetilde{\bW}_{k'}^{(v')}, \bx_{k',i'}) - \frac{1}{m}\sum_{r=1}^m \left[ \overline{\mathbb{P}}_{y_{k,i},r,k,i}^{(v')}  - \overline{\mathbb{P}}_{y_{k',i'},r,k',i'}^{(v')} \right]} \leq 1.75. \nn  
\end{align}
\end{lemma}

\begin{proof}
We can write,
\begin{align}
    &y_{k,i}f(\lmt^{(v')}, \bx_{k,i})-  y_{k',i'}f(\widetilde{\bW}_{k'}^{(v')},  \bx_{k',i'}) \nonumber\\
    & = F_{y_{k,i}}(\widetilde{\bW}_{y_{k,i},k}^{(v')},\bx_{k,i}) - F_{-y_{k,i}}(\widetilde{\bW}_{-y_{k,i},k}^{(v')},\bx_{k,i}) \nn \\
    & \quad - F_{y_{k',i'}}(\widetilde{\bW}_{y_{k',i'},k'}^{(v')},\bx_{k',i'}) +  F_{-y_{k',i'}}(\widetilde{\bW}_{-y_{k',i'},k'}^{(v')},\bx_{k',i'}) \nonumber\\
    & = F_{-y_{k',i'}}(\widetilde{\bW}_{-y_{k',i'},k'}^{(v')},\bx_{k',i'})  - F_{-y_{k,i}}(\widetilde{\bW}_{-y_{k,i},k}^{(v')},\bx_{k,i}) \nn \\
    & \quad + F_{y_{k,i}}(\widetilde{\bW}_{y_{k,i},k}^{(v')},\bx_{k,i}) - F_{y_{k',i'}}(\widetilde{\bW}_{y_{k',i'},k'}^{(v')},\bx_{k',i'}) \nonumber \\
    & = \underbrace{F_{-y_{k',i'}}(\widetilde{\bW}_{-y_{k',i'},k'}^{(v')},\bx_{k',i'})  - F_{-y_{k,i}}(\widetilde{\bW}_{-y_{k,i},k}^{(v')},\bx_{k,i})}_{I_1} \nonumber\\
    & \hspace{5pt} + \underbrace{\frac{1}{m}\sum_{r=1}^m \left[\sigma\left( \inner{\widetilde{\bw}_{y_{k,i},r,k}^{(v')}}{y_{k,i}\bmu} \right) - \sigma\left( \inner{\widetilde{\bw}_{y_{k',i'},r,k'}^{(v')}}{y_{k',i'
    }\bmu} \right) \right]}_{I_2} \nonumber \\
    & \hspace{5pt} + \underbrace{\frac{1}{m}\sum_{r=1}^m \left[\sigma\left( \inner{\widetilde{\bw}_{y_{k,i},r,k}^{(v')}}{\bxi_{k,i}} \right) - \sigma\left( \inner{\widetilde{\bw}_{y_{k',i'},r,k'}^{(v')}}{\bxi_{k',i'}} \right) \right]}_{I_3}. \nonumber
\end{align}
Next we bound $I_1, I_2$ and $I_3$ as follows.
\begin{align}
    |I_1| \leq F_{-y_{k',i'}}(\widetilde{\bW}_{-y_{k',i'},k'}^{(v')},\bx_{k',i'}) + F_{-y_{k,i}}(\widetilde{\bW}_{-y_{k,i},k}^{(v')},\bx_{k,i}) \overset{(a)}{\leq} 1, \nn
\end{align}
where $(a)$ follows from part 1 of \Cref{lemma:bounded_func_output}.
For $|I_2|$ we have the following bound,
\begin{align}
    |I_2| &\leq \max\left\{\frac{1}{m}\sum_{r=1}^m \sigma\left( \inner{\widetilde{\bw}_{y_{k,i},r,k}^{(v')}}{y_{k,i}\bmu}\right) ,  \frac{1}{m}\sum_{r=1}^m \sigma\left( \inner{\widetilde{\bw}_{y_{k',i'},r,k'}^{(v')}}{y_{k',i'}\bmu}\right)\right\} \nonumber \\
    & \overset{(a)}{\leq} 2 \max_{r \in [m]} \left\{\abs{\inner{\bw_{y_{k,i},r}^{(0)}}{\bmu}}, \abs{\inner{\bw_{y_{k',i'},r}^{(0)}}{\bmu}}, \mathbb{G}_{y_{k,i},r,k}^{(v')}, \mathbb{G}_{y_{k',i'},r,k'}^{(v')} \right\} \nn \\
    & \overset{(b)}{\leq} 2 \max_{r \in [m]} \left\{\beta, C'\hat{\gamma}\alpha \right\} \nn
    \\
    & \overset{(c)}{\leq} 0.25. \nn
\end{align}
Here $(a)$ follows \Cref{lemma:inner_prod_bound}, $(b)$ follows from the definition of $\beta$ and the induction hypothesis, $(c)$ follows from \Cref{lemma:beta_mag} and \Cref{assump:d}. 

Next we derive an upper bound on $I_3$ as follows.
\begin{align}
    I_3 &= \frac{1}{m}\sum_{r=1}^m \left[\sigma\left( \inner{\widetilde{\bw}_{y_{k,i},r,k}^{(v')}}{\bxi_{k,i}} \right) - \sigma\left( \inner{\widetilde{\bw}_{y_{k',i'},r,k'}^{(v')}}{\bxi_{k',i'}} \right) \right] \nonumber \\
    & \overset{(a)}{\leq} \frac{1}{m}\sum_{r=1}^m \left[ \inner{\widetilde{\bw}_{y_{k,i},r,k}^{(v')}}{\bxi_{k,i}}  - \inner{\widetilde{\bw}_{y_{k',i'},r,k'}^{(v')}}{\bxi_{k',i'}}\right] + 0.25 \nonumber \\
    & \overset{(b)}{\leq} \frac{1}{m}\sum_{r=1}^m \left[ \overline{\mathbb{P}}_{y_{k,i},r,k,i}^{(v')}  - \overline{\mathbb{P}}_{y_{k',i'},r,k',i'}^{(v')} \right]+ 2\beta + 8\sqrt{\frac{\log (6n^2/\delta)}{d}}n\alpha + 0.25 \nonumber \\
    & \overset{(c)}{\leq} \frac{1}{m}\sum_{r=1}^m \left[ \overline{\mathbb{P}}_{y_{k,i},r,k,i}^{(v')}  - \overline{\mathbb{P}}_{y_{k',i'},r,k',i'}^{(v')} \right] + 0.5. \nn
\end{align}
Here $(a)$ follows from \Cref{lemma:min_and_max_noise_inner_prod}; $(b)$ follows from \Cref{lemma:inner_prod_bound}; $(c)$ follows from \Cref{lemma:beta_mag}.

Similarly, we can get a lower bound for $I_3$ as follows,
\begin{align}
    I_3 &= \frac{1}{m}\sum_{r=1}^m \left[\sigma\left( \inner{\widetilde{\bw}_{y_{k,i},r,k}^{(v')}}{\bxi_{k,i}} \right) - \sigma\left( \inner{\widetilde{\bw}_{y_{k',i'},r,k'}^{(v')}}{\bxi_{k',i'}} \right) \right] \nonumber \\
    & \overset{(a)}{\geq} \frac{1}{m}\sum_{r=1}^m \left[ \inner{\widetilde{\bw}_{y_{k,i},r,k}^{(v')}}{\bxi_{k,i}}  - \inner{\widetilde{\bw}_{y_{k',i'},r,k'}^{(v')}}{\bxi_{k',i'}}\right] - 0.25 \nonumber \\
    & \overset{(b)}{\geq} \frac{1}{m}\sum_{r=1}^m \left[ \overline{\mathbb{P}}_{y_{k,i},r,k,i}^{(v')}  - \overline{\mathbb{P}}_{y_{k',i'},r,k',i'}^{(v')} \right] -2\beta - 8\sqrt{\frac{\log (6n^2/\delta)}{d}}n\alpha - 0.25 \nonumber \\
    & \overset{(c)}{\geq} \frac{1}{m}\sum_{r=1}^m \left[ \overline{\mathbb{P}}_{y_{k,i},r,k,i}^{(v')}  - \overline{\mathbb{P}}_{y_{k',i'},r,k',i'}^{(v')} \right] - 0.5. \nn
\end{align}
Here $(a)$ follows from \Cref{lemma:min_and_max_noise_inner_prod}; $(b)$ follows from \Cref{lemma:inner_prod_bound}; $(c)$ follows from \Cref{lemma:beta_mag}. 

Combining the above results, we have
\begin{align}
     y_{k,i}f(\lmt^{(v')}, \bx_{k,i})-  y_{k',i'}f(\widetilde{\bW}_{k'}^{(v')}, \bx_{k',i'}) &\leq |I_1| + |I_2| + I_3 \nonumber \\
    & \leq \frac{1}{m}\sum_{r=1}^m \left[ \overline{\mathbb{P}}_{y_{k,i},r,k,i}^{(v')}  - \overline{\mathbb{P}}_{y_{k',i'},r,k',i'}^{(v')} \right] + 1.75, \nn
\end{align}
and,
\begin{align}
    y_{k,i}f(\lmt^{(v')}, \bx_{k,i})-  y_{k',i'}f(\widetilde{\bW}_{k'}^{(v')}, \bx_{k',i'}) &\geq -|I_1| - |I_2| + I_3 \nonumber \\
    & \geq \frac{1}{m}\sum_{r=1}^m \left[ \overline{\mathbb{P}}_{y_{k,i},r,k,i}^{(v')}  - \overline{\mathbb{P}}_{y_{k',i'},r,k',i'}^{(v')} \right] - 1.75. \nn
\end{align}
This implies,
\begin{align}
    \abs{y_{k,i}f(\lmt^{(v')}, \bx_{k,i})-  y_{k',i'}f(\widetilde{\bW}_{k'}^{(v')}, \bx_{k',i'}) - \frac{1}{m}\sum_{r=1}^m \left[ \overline{\mathbb{P}}_{y_{k,i},r,k,i}^{(v')}  - \overline{\mathbb{P}}_{y_{k',i'},r,k',i'}^{(v')} \right]} \leq 1.75. \nn
\end{align}
\end{proof}

We will now state and prove a version of Lemma C.7 that appears in \cite{cao2022benign}. Note that \cite{cao2022benign} only considers the heterogeneity arising due to different datapoints for the same model. Interestingly, we show that the lemma can be extended to the case with different local models and different datapoints as long as the local models start from the same initialization. 

\begin{lemma}
\label{lemma:bounded_deriv_frac}
Suppose \eqref{eq:rhop_bound}, \eqref{eq:rhov_bound} and \eqref{eq:gamma_bound} hold for all $0 \leq v \leq v'$. Then the following holds for all $0 \leq v \leq v'$.  

\begin{enumerate}[leftmargin=*]
    \item \label{bounded_deriv_frac_1} $\frac{1}{m}\sum_{r=1}^m \left[ \overline{\mathbb{P}}_{y_{k,i},r,k,i}^{(v)}  - \overline{\mathbb{P}}_{y_{k',i'},r,k',i'}^{(v)} \right] \leq \kappa$ for all $k,k' \in [K], i,i' \in [N]$.
    \item \label{bounded_deriv_frac_2} $y_{k,i}f(\lmt^{(v)}, \bx_{k,i})-  y_{k',i'}f(\widetilde{\bW}_{k'}^{(v)},  \bx_{k',i'}) \leq C_1$ for all $k, k' \in [K]$ and $i,i' \in [N]$.
    \item \label{bounded_deriv_frac_3} $\frac{\lderivd^{(v)}}{\lderiv^{(v)}} \leq C_2 = \exp(C_1)$ for all $k,k' \in [K]$ and $i, i' \in [N]$.
    \item \label{bounded_deriv_frac_4} $S_{k,i}^{(0)} \subseteq S_{k,i}^{(v)}$ where $S_{k,i}^{(v)} := \left\{r \in [m]: \inner{\widetilde{\bw}_{y_{k,i},r,k}^{(v)}}{\bxi_{k,i}} \geq 0 \right\}$, and hence $\abs{S_{k,i}^{(v)}} \geq 0.4m$ for all $k \in [K], i \in [N]$.
    \item \label{bounded_deriv_frac_5} $\tilde{S}_{j,r}^{(0)} \subseteq \tilde{S}_{j,r}^{(v)}$ where $\tilde{S}_{j,r}^{(0)} := \left\{k \in [K], i \in [N]: y_{k,i} = j, \inner{\widetilde{\bw}_{j,r,k}^{(v)}}{\bxi_{k,i}} \geq 0 \right\}$, and hence $\abs{\tilde{S}_{j,r}^{(v)}} \geq \frac{n}{8}$.
\end{enumerate}
Here we take $\kappa = 5$ and $C_1 = 6.75$.
\end{lemma}

\begin{proof}[Proof of \ref{bounded_deriv_frac_1}]
We will use a proof by induction. For $v=0$, it is simple to verify that \ref{bounded_deriv_frac_1} holds since $\overline{\mathbb{P}}_{j,r,k,i}^{(0)} = 0$ for all $j \in \{\pm 1\}, r \in [m], k \in [K], i \in [N]$ by definition. Now suppose \ref{bounded_deriv_frac_1} holds for all $0 \leq v \leq \tilde{v} < v'$. Then we will show that \ref{bounded_deriv_frac_1} also holds at $v = \tilde{v}+1$. We have the following cases.

\textbf{Case 1: $(\tilde{v}+1) \pmod \tau \neq 0$}

In this case, from \eqref{eq:gprhov_update}
\begin{align}
     \overline{\mathbb{P}}_{y_{k,i}, r, k, i}^{(\tilde{v}+1)} =  \overline{\mathbb{P}}_{y_{k,i},r,k,i}^{(\tilde{v})} - \frac{\eta}{Nm} \lderiv^{(\tilde{v})} \noisederiv{\widetilde{\bw}_{y_{k,i},r,k,i}^{(\tilde v)}} \norm{\bxi_{k,i}}. \nn
\end{align}

Thus, 

\begin{align}
\label{eq:rho_pos_update}
    \frac{1}{m}\sum_{r=1}^m \left[ \overline{\mathbb{P}}_{y_{k,i},r,k,i}^{(\tilde{v}+1)}  - \overline{\mathbb{P}}_{y_{k',i'},r,k',i'}^{(\tilde{v}+1)} \right] &= \frac{1}{m}\sum_{r=1}^m \left[ \overline{\mathbb{P}}_{y_{k,i},r,k,i}^{(\tilde{v})}  - \overline{\mathbb{P}}_{y_{k',i'},r,k',i'}^{(\tilde{v})} \right] \nonumber \\
    & \hspace{2pt} + \frac{\eta }{Nm^2} \lbr \abs{S_{k,i}^{(\tilde{v})}}(-\lderiv^{(\tilde{v})}) \norm{\bxi_{k,i}} - \abs{S_{k',i'}^{(\tilde{v})}} (-\lderivd^{(\tilde{v})}) \norm{\bxi_{k',i'}} \rbr,
\end{align}
where $S_{k,i}^{(\tilde{v})}, S_{k',i'}^{(\tilde{v})}$ are defined in \ref{bounded_deriv_frac_4}.

We bound \eqref{eq:rho_pos_update} in two cases, depending on the value of $\frac{1}{m}\sum_{r=1}^m \left[ \overline{\mathbb{P}}_{y_{k,i},r,k,i}^{(\tilde{v})}  - \overline{\mathbb{P}}_{y_{k',i'},r,k',i'}^{(\tilde{v})} \right]$.

\begin{itemize}
    \item[\textbf{i)}] If $\frac{1}{m}\sum_{r=1}^m \left[ \overline{\mathbb{P}}_{y_{k,i},r,k,i}^{(\tilde{v})}  - \overline{\mathbb{P}}_{y_{k',i'},r,k',i'}^{(\tilde{v})} \right] \leq 0.9\kappa$. From \eqref{eq:rho_pos_update} we have,
    \begin{align}
        \frac{1}{m}\sum_{r=1}^m \left[ \overline{\mathbb{P}}_{y_{k,i},r,k,i}^{(\tilde{v}+1)}  - \overline{\mathbb{P}}_{y_{k',i'},r,k',i'}^{(\tilde{v}+1)} \right] &\leq 0.9\kappa +  \frac{\eta }{Nm^2} \abs{S_{k,i}^{(\tilde{v})}}(-\lderiv^{(\tilde{v})}) \norm{\bxi_{k,i}} \nonumber \\
        & \overset{(a)}{\leq}  0.9\kappa + \frac{\eta}{Nm}\norm{\bxi_{k,i}} \nonumber \\
        & \overset{(b)}{\leq} \kappa. \nn
    \end{align}
    $(a)$ follows from $\abs{S_{k,i}^{(\tilde{v})}} \leq m, -\ell'(\cdot) \leq 1$;$(b)$ follows from \Cref{lemma:noise_corr} and \Cref{assump:eta}.
    %%%%%%%%%%%%%%%%%%%%%%
    \item[\textbf{ii)}] If $\frac{1}{m}\sum_{r=1}^m \left[ \overline{\mathbb{P}}_{y_{k,i},r,k,i}^{(\tilde{v})}  - \overline{\mathbb{P}}_{y_{k',i'},r,k',i'}^{(\tilde{v})} \right] > 0.9\kappa$. From \Cref{lemma:bounded_func_diff} we know that,
    \begin{align}
    \label{eq:bounded_func_diff_1.75}
        y_{k,i}f(\lmt^{(\tilde{v})}, \bx_{k,i})-  y_{k',i'}f(\widetilde{\bW}_{k'}^{(\tilde{v})}, \bx_{k',i'}) &\geq \frac{1}{m}\sum_{r=1}^m \left[ \overline{\mathbb{P}}_{y_{k,i},r,k,i}^{(\tilde{v})}  - \overline{\mathbb{P}}_{y_{k',i'},r,k',i'}^{(\tilde{v})} \right] -1.75 \nonumber \\
        & \overset{(a)}{\geq} 0.9\kappa - 0.35\kappa \nonumber \\
        & = 0.55\kappa.
    \end{align}
    where $(a)$ follows from $\kappa = 5$. Also note that since $\frac{1}{m} \sum_{r=1}^m  \overline{\mathbb{P}}_{y_{k,i}, r, k, i}^{(\tilde{v})} \geq \frac{1}{m}\sum_{r=1}^m \overline{\mathbb{P}}_{y_{k',i'},r,k',i'}^{(\tilde{v})} + 0.9\kappa  \geq 0.9\kappa = 4.5$, we have from \Cref{lemma:bounded_func_output} that 
    \begin{align}
    \label{eq:min_func_output_3.75}
        y_{k,i}f(\lmt^{(\tilde{v})}, \bx_{k,i}) \geq 3.75.
    \end{align}
    Now from the definition of $\ell(\cdot)$ we have,
    \begin{align}
    \label{eq:bounded_lderiv_1}
        \frac{(-\lderiv^{(\tilde{v})})}{(-\ell_{k',i'}^{(\tilde{v})})} & = \frac{1+\exp(y_{k',i'}f(\widetilde{\bW}_{k'}^{(\tilde{v})}, \bx_{k',i'}))}{1+\exp( y_{k,i} f(\lmt^{(\tilde{v})}, \bx_{k,i}))} \nonumber \\
        & \overset{(a)}{\leq} \frac{1 + \exp(y_{k,i} f(\lmt^{(\tilde{v})}, \bx_{k,i}) - 0.55\kappa)}{1+\exp( y_{k,i}f(\lmt^{(\tilde{v})}, \bx_{k,i}))} \nonumber \\
        & \overset{(b)}{<} 1/7.5.
    \end{align}
    Here $(a)$ follows from \eqref{eq:bounded_func_diff_1.75}; $(b)$ follows from \eqref{eq:min_func_output_3.75}.

    Thus,
    \begin{align}
        \frac{\abs{S_{k, i}^{(\tilde{v})}} \norm{\bxi_{k,i}}(-\lderiv^{(\tilde{v})})} {\abs{S_{k',i'}^{(\tilde{v})}} \norm{\bxi_{k',i'}}(-\ell_{k',i'}^{(\tilde{v})})} \overset{(a)}{\leq} 2.5 \frac{\norm{\bxi_{k,i}}(-\lderiv^{(\tilde{v})})}{\norm{\bxi_{k',i'}}(-\ell_{k',i'}^{(\tilde{v})})}  \overset{(b)}{\leq} 2.5 \cdot 3  \frac{(-\lderiv^{(\tilde{v})})}{(-\ell_{k',i'}^{(\tilde{v})})} \overset{(c)}{<}1. \nn
    \end{align}
    Here $(a)$ follows from $\abs{S_{k,i}^{(\tilde{v})}} \leq m, \abs{S_{k',i'}^{(\tilde{v})}} \geq 0.4m$ using our induction hypothesis; $(b)$ follows from \Cref{lemma:noise_corr}; $(c)$ follows from \eqref{eq:bounded_lderiv_1}. 
    This implies $\abs{S_{k,i}^{(\tilde{v})}}\norm{\bxi_{k,i}}(-\lderiv^{(\tilde{v})}) < \abs{S_{k',i'}^{(\tilde{v})}} \norm{\bxi_{k',i'}}(-\ell_{k',i'}^{(\tilde{v})})$. Now from \eqref{eq:rho_pos_update} we have,
    \begin{align}
         \frac{1}{m}\sum_{r=1}^m \left[ \overline{\mathbb{P}}_{y_{k,i},r,k,i}^{(\tilde{v}+1)}  - \overline{\mathbb{P}}_{y_{k',i'},r,k',i'}^{(\tilde{v}+1)} \right] &\leq \frac{1}{m}\sum_{r=1}^m \left[ \overline{\mathbb{P}}_{y_{k,i},r,k,i}^{(\tilde{v})}  - \overline{\mathbb{P}}_{y_{k',i'},r,k',i'}^{(\tilde{v})} \right] \leq  \nn \kappa,
    \end{align}
    where the last inequality follows from our induction hypothesis. 
\end{itemize}

\textbf{Case 2: $(\tilde{v}+1) \pmod \tau = 0$}

In this case, using \eqref{eq:gprhov_update} we can write our update equation as follows:
\begin{align}
\label{eq:rho_diff_update_scaling}
   &  \frac{1}{m}\sum_{r=1}^m \left[ \overline{\mathbb{P}}_{y_{k,i},r,k,i}^{(\tilde{v}+1)}  - \overline{\mathbb{P}}_{y_{k',i'},r,k',i'}^{(\tilde{v}+1)} \right] \nonumber \\
    & = \frac{1}{m}\sum_{r=1}^m \left[ \overline{\mathbb{P}}_{y_{k,i},r,k,i}^{(\tilde{v}+1-\tau)}  - \overline{\mathbb{P}}_{y_{k',i'},r,k',i'}^{(\tilde{v}+1-\tau)} \right] \nonumber \\
    & \hspace{2pt} + \frac{1}{n} \underbrace{\frac{\eta }{m^2} \sum_{s=0}^{\tau-1}\Bigg( \abs{S_{k,i}^{(\tilde{v}+1-\tau+s)}}(-\lderiv^{(\tilde{v}+1-\tau+s)}) \norm{\bxi_{k,i}} - \abs{S_{k',i'}^{(\tilde{v}+1-\tau+s)}} (-\ell_{k',i'}^{(\tilde{v}+1-\tau+s)}) \norm{\bxi_{k',i'}}\Bigg)}_{:= I_1} \nonumber \\
    & = \frac{1}{m}\sum_{r=1}^m \left[ \overline{\mathbb{P}}_{y_{k,i},r,k,i}^{(\tilde{v}+1-\tau)}  - \overline{\mathbb{P}}_{y_{k',i'},r,k',i'}^{(\tilde{v}+1-\tau)} \right] + \frac{I_1}{n}.
\end{align}

From our induction hypothesis

we know that
\begin{align}
\label{eq:inter_step_pos}
    \frac{1}{m}\sum_{r=1}^m \left[ \overline{\mathbb{P}}_{y_{k,i},r,k,i}^{(\tilde{v})}  - \overline{\mathbb{P}}_{y_{k',i'},r,k',i'}^{(\tilde{v})} \right] \leq \kappa.
\end{align}

Now unrolling the LHS expression in \eqref{eq:inter_step_pos} using \eqref{eq:gprhov_update}, we see that this implies

\begin{align}
\label{eq:rho_diff_update_no_scaling}
    & \frac{1}{m}\sum_{r=1}^m \left[ \overline{\mathbb{P}}_{y_{k,i},r,k,i}^{(\tilde{v}+1-\tau)}  - \overline{\mathbb{P}}_{y_{k',i'},r,k',i'}^{(\tilde{v}+1-\tau)} \right] + \frac{I_1}{N} \leq \kappa
\end{align}

\textbf{Case 2a):} $I_1 \geq  0$.

In this case it directly follows \eqref{eq:rho_diff_update_scaling} and \eqref{eq:rho_diff_update_no_scaling} that $\frac{1}{m}\sum_{r=1}^m \left[ \overline{\mathbb{P}}_{y_{k,i},r,k,i}^{(\tilde{v}+1)}  - \overline{\mathbb{P}}_{y_{k',i'},r,k',i'}^{(\tilde{v}+1)} \right] \leq \kappa$ since $N \leq n$.

\textbf{Case 2b):} If $I_1 < 0$.

In this case from \eqref{eq:rho_diff_update_scaling} we have,
\begin{align}
   \frac{1}{m}\sum_{r=1}^m \left[ \overline{\mathbb{P}}_{y_{k,i},r,k,i}^{(\tilde{v}+1)}  - \overline{\mathbb{P}}_{y_{k',i'},r,k',i'}^{(\tilde{v}+1)} \right] &\leq  \frac{1}{m}\sum_{r=1}^m \left[ \overline{\mathbb{P}}_{y_{k,i},r,k,i}^{(\tilde{v}+1-\tau)}  - \overline{\mathbb{P}}_{y_{k',i'},r,k',i'}^{(\tilde{v}+1-\tau)} \right] \leq \kappa. \nn
\end{align}
where the last inequality follows from our induction hypothesis.
\end{proof}

\begin{proof}[Proof of \ref{bounded_deriv_frac_2}]
For any $0 \leq v \leq v'$ we have,
\begin{align}
    y_{k,i}f(\lmt^{(v)}, \bx_{k,i})-  y_{k',i'}f(\widetilde{\bW}_{k'}^{(v)}, \bx_{k',i'}) & \overset{(a)}{\leq} \frac{1}{m}\sum_{r=1}^m \left[ \overline{\mathbb{P}}_{y_{k,i},r,k,i}^{(v)}  - \overline{\mathbb{P}}_{y_{k',i'},r,k',i'}^{(v)} \right] + 1.75 \nonumber \\
    & \overset{(b)}{\leq} \kappa + 1.75 = C_1. \nn
\end{align}
Here $(a)$ follows from \Cref{lemma:bounded_func_diff}; $(b)$ follows from \ref{bounded_deriv_frac_1}.
\end{proof}

\begin{proof}[Proof of \ref{bounded_deriv_frac_3}]
For any $0 \leq v \leq v'$ we have,
\begin{align}
    \frac{\lderivd^{(v)}}{\lderiv^{(v)}} 
    & \overset{(a)}{\leq} \max \left\{1,  \exp\left(y_{k,i}f(\lmt^{(v)}, \bx_{k,i})-  y_{k',i'}f(\widetilde{\bW}_{k'}^{(v)}, \bx_{k',i'})  \right) \right\} \overset{(b)}{\leq} \exp(C_1).\nn
\end{align}
Here $(a)$ follows from \Cref{lemma:bounded_lderiv_frac};$(b)$ follows from \ref{bounded_deriv_frac_2}.
\end{proof}

\begin{proof}[Proof of \ref{bounded_deriv_frac_4}]
To prove \ref{bounded_deriv_frac_4}, we will use the result in \ref{bounded_deriv_frac_3} and show that $\inner{\widetilde{\bw}_{y_{k,i}, r, k}^{(0)}}{\bxi_{k,i}} > 0$ implies $\inner{\widetilde{\bw}_{y_{k,i}, r, k}^{(v)}}{\bxi_{k,i}} > 0$ for all $1 \leq v \leq v'$. We use a proof by induction.  Assuming $\inner{\widetilde{\bw}_{y_{k,i}, r, k}^{(v)}}{\bxi_{k,i}} > 0$ for all $0 \leq v \leq \tilde{v} < v'$, we will show that $\inner{\widetilde{\bw}_{y_{k,i}, r, k}^{(\tilde{v}+1)}}{\bxi_{k,i}} > 0$. We have the following cases.

\textbf{Case 1: $(\tilde{v}+1) \pmod \tau \neq 0$}.

Using the fact that $\inner{\widetilde{\bw}_{y_{k,i}, r, k}^{(\tilde{v})}}{\bxi_{k,i}} > 0$ we have,

\begin{align}
    \inner{\widetilde{\bw}_{y_{k,i}, r, k}^{(\tilde{v}+1)}}{\bxi_{k,i}} & = \inner{\widetilde{\bw}_{y_{k,i}, r, k}^{(\tilde{v})}}{\bxi_{k,i}} + \frac{\eta}{Nm}(-\lderiv^{(\tilde{v})})\norm{\bxi_{k,i}} \nonumber \\
    & \hspace{10pt} + \frac{\eta}{Nm}\sum_{i' \in [N], i' \neq i} (-{\ell'}_{k,i'}^{(\tilde{v})})\sigma'\left(\inner{\widetilde{\bw}_{y_{k,i},r,k}^{(\tilde{v})}}{\bxi_{k,i'}} \right) \inner{\bxi_{k,i}}{\bxi_{k,i'}} \nonumber\\
    & \overset{(a)}{\geq} \inner{\widetilde{\bw}_{y_{k,i}, r, k}^{(\tilde{v})}}{\bxi_{k,i}} + \frac{\eta \sigma_p^2 d}{2Nm}(-\lderiv^{(\tilde{v})}) -\frac{\eta}{Nm}2\sigma_p^2 \sqrt{d \log (4n^2/\delta)}\sum_{i' \in [N], i' \neq i} (-{\ell'}_{k,i'}^{(\tilde{v})}) \nonumber \\
    & \overset{(b)}{\geq} \inner{\widetilde{\bw}_{y_{k,i}, r, k}^{(\tilde{v})}}{\bxi_{k,i}} + \frac{\eta \sigma_p^2 d}{2Nm}(-\lderiv^{(\tilde{v})}) -\frac{\eta}{m}2\sigma_p^2 \sqrt{d \log (4n^2/\delta)} C_2(-\lderiv^{(\tilde{v})})\nonumber \\
    & \overset{(c)}{\geq} \inner{\widetilde{\bw}_{y_{k,i}, r, k}^{(\tilde{v})}}{\bxi_{k,i}} \nonumber \\
    & > 0. \nn
\end{align}

Here $(a)$ follows from \Cref{lemma:noise_corr}; $(b)$ follows from \ref{bounded_deriv_frac_3}; $(c)$ follows from \Cref{assump:d} by choosing a sufficiently large $d$.

\textbf{Case 2: $(\tilde{v}+1) \pmod \tau = 0$}.

From our induction hypothesis we know that $\inner{\widetilde{\bw}_{y_{k,i}, r, k}^{(\tilde{v}+1-\tau+s)}}{\bxi_{k,i}} > 0$ for all $0\leq s \leq \tau-1$. Then,

\begin{align}
    \inner{\widetilde{\bw}_{y_{k,i}, r, k}^{(\tilde{v})}}{\bxi_{k,i}} &= \inner{\widetilde{\bw}_{y_{k,i}, r,k}^{(\tilde{v}+1-\tau)}}{\bxi_{k,i}} + \underbrace{\frac{\eta}{nm}\sum_{s=0}^{\tau-1}(-\lderiv^{(\tilde{v}+1-\tau+s)})\norm{\bxi_{k,i}}}_{I_1} \nonumber \\
    & \hspace{10pt} + \underbrace{\frac{\eta}{nm} \sum_{s=0}^{\tau-1} \sum_{i' \in [N], i' \neq i}  (-{\ell'}_{k,i'}^{(\tilde{v}+1-\tau+s)})\sigma'\left(\inner{\widetilde{\bw}_{y_{k,i},r,k}^{(\tilde{v}+1-\tau+s)}}{\bxi_{k,i'}} \right) \inner{\bxi_{k,i}}{\bxi_{k,i'}}}_{I_2}\nonumber\\
    & \hspace{10pt} + 
    \underbrace{\frac{\eta}{nm}\sum_{s=0}^{\tau-1}  \sum_{k',k' \neq k}  \sum_{i' \in [N]} 
    (-{\ell'}_{k',i'}^{(\tilde{v}+1-\tau+s)})\sigma'\left(\inner{\widetilde{\bw}_{y_{k,i},r,k'}^{(\tilde{v}+1-\tau+s)}}{\bxi_{k',i'}} \right)\inner{\bxi_{k,i}}{\bxi_{k',i'}}}_{I_3}
    \label{eq:inner_prod_update}
\end{align}

Using \Cref{lemma:noise_corr} we can lower bound $I_1$ as follows:
\begin{align}
    I_1 \geq \frac{\eta\sigma_p^2d}{2nm}\sum_{s=0}^{\tau-1}(-\lderiv^{(\tilde{v}+1-\tau+s)}), \nn
\end{align}
where the inequality follows from \Cref{lemma:noise_corr}.

For $|I_2|$ we have,

\Cref{lemma:noise_corr} as follows:
\begin{align}
    |I_2| &\overset{(a)}{\leq} \frac{\eta 2\sigma_p^2 \sqrt{d \log (4n^2/\delta)}  }{nm} 
    \sum_{s=0}^{\tau-1} \sum_{i' \in [N], i' \neq i}  (-{\ell'}_{k,i'}^{(\tilde{v}+1-\tau+s)}) \nonumber \\
    & \overset{(b)}{\leq} \frac{\eta(N-1)C_2 2\sigma_p^2 \sqrt{d \log (4n^2/\delta)}}{nm}  \sum_{s=0}^{\tau-1} (-\lderiv^{(\tilde{v}+1-\tau+s)}). \nn
\end{align}
Here $(a)$ follows from \Cref{lemma:noise_corr}; $(b)$ follows from \ref{bounded_deriv_frac_3}.
Similarly we can bound $|I_3|$ as follows,
\begin{align}
    |I_3| & \overset{(a)}{\leq} \frac{\eta 2\sigma_p^2 \sqrt{d \log (4n^2/\delta)}  }{nm} 
    \sum_{s=0}^{\tau-1} \sum_{k',k' \neq k}  \sum_{i' \in [N]} 
    (-{\ell'}_{k',i'}^{(\tilde{v}+1-\tau+s)}) \nonumber \\
    & \overset{(b)}{\leq} \frac{\eta(n-N)C_2 2\sigma_p^2 \sqrt{d \log (4n^2/\delta)} }{nm} \sum_{s=0}^{\tau-1} (-\lderiv^{(\tilde{v}+1-\tau+s)}). \nn
\end{align}
Here $(a)$ follows from \Cref{lemma:noise_corr}; $(b)$ follows from \ref{bounded_deriv_frac_3}.
Substituting the bounds for $I_1, |I_2|, |I_3|$ in \eqref{eq:inner_prod_update} we have,
\begin{align}
   \inner{\widetilde{\bw}_{y_{k,i}, r, k}^{(\tilde{v})}}{\bxi_{k,i}} & \geq \inner{\widetilde{\bw}_{y_{k,i}, r,k}^{(\tilde{v}+1-\tau)}}{\bxi_{k,i}} + I_1 - |I_2| - |I_3| \nonumber \\
   & \geq \inner{\widetilde{\bw}_{y_{k,i}, r,k}^{(\tilde{v}+1-\tau)}}{\bxi_{k,i}} + \frac{\eta\sigma_p^2d}{2nm}\sum_{s=0}^{\tau-1}(-\lderiv^{(\tilde{v}+1-\tau)+s}) \nn \\
   & \hspace{2pt} - \frac{\eta C_2}{m} 2\sigma_p^2 \sqrt{d \log (4n^2/\delta)} \sum_{s=0}^{\tau-1} (-\lderiv^{(\tilde{v}+1-\tau+s)}) \nonumber \\
   & \overset{(a)}{\geq} \inner{\widetilde{\bw}_{y_{k,i}, r,k}^{(\tilde{v}+1-\tau)}}{\bxi_{k,i}} \nonumber \\
   & \geq 0. \nn
\end{align}
Here $(a)$ follows from \Cref{assump:d} by choosing a sufficiently large $d$. 
Thus we have shown that $\inner{\widetilde{\bw}_{y_{k,i}, r, k}^{(v)}}{\bxi_{k,i}} \geq 0$ for all $0 \leq v \leq v'$ and $r$ such that $\inner{\bw_{y_{k,i}, r, k}^{(0)}}{\bxi_{k,i}} \geq 0$. This implies $S_{k,i}^{(0)} \subseteq S_{k,i}^{(v)}$ for all $0 \leq v \leq v'$. Furthermore we know that $\abs{S_{k,i}^{(0)}} \geq 0.4m$ for all $k \in [K], i \in [N]$ from \Cref{lemma:min_size_activated_noise_filters} and thus $\abs{S_{k,i}^{(v)}} \geq 0.4m$ for all $k \in [K], i \in [N], 0 \leq v \leq v'$.
\end{proof}

\begin{proof}[Proof of \ref{bounded_deriv_frac_5}]
Note that as part of the proof of \ref{bounded_deriv_frac_4} we have already shown that $\inner{\widetilde{\bw}_{j, r, k}^{(v)}}{\bxi_{k,i}} \geq 0$  for all $0 \leq v \leq v'$ and $k,i$ such that $y_{k,i} = j$ and $\inner{\widetilde{\bw}_{j, r, k}^{(0)}}{\bxi_{k,i}} \geq 0$. This implies $\tilde{S}_{j,r}^{(0)} \subseteq \tilde{S}_{j,r}^{(v)}$ for all $0 \leq v \leq v'$. Furthermore we know that $\abs{\tilde{S}_{j,r}^{(0)}} \geq n/8$ for all $j \in \{\pm 1\}, r\in [m]$ from \Cref{lemma:min_size_activated_noise_data} and thus $\abs{\tilde{S}_{j,r}^{(v)}} \geq n/8$ for all $j \in \{\pm 1\}, r\in [m]$. 

This concludes the proof of \Cref{lemma:bounded_deriv_frac}.
\end{proof}

We are now ready to prove \Cref{thm:coeff_bound}.

\subsubsection{Proof of \Cref{thm:coeff_bound}}
\label{subsec:thm_coeff_bound_proof}

We will again use a proof by induction to prove this theorem.

\begin{proof}[Proof of \eqref{eq:rhov_bound}]
For $j = y_{k,i}$ we know from \eqref{eq:gnrhov_update} that $\gnrhov^{(v'+1)} = 0$ and hence we look at the case where $ j \neq y_{k,i}$.

\textbf{Case 1: $(v'+1) \pmod \tau \neq 0$}.

\textbf{a)} If $ \gnrhov^{(v')} < -0.5\beta -4\sqrt{\frac{\log (6n^2/\delta)}{d}}n\alpha$, then from \eqref{eq:rhon_inner_prod_bound} in \Cref{lemma:inner_prod_bound} we know that,
\begin{align}
    \inner{\lmweightt^{(v')}}{\bxi_{k,i}} &\leq  
    \inner{\gmweight^{(0)}}{\bxi_{k,i}} + \gnrhov^{(v')} + 4\sqrt{\frac{\log (6n^2/\delta)}{d}}n\alpha \nonumber \\
    & \overset{(a)}{\leq} 0.5\beta + \gnrhov^{(v')} + 4\sqrt{\frac{\log (6n^2/\delta)}{d}}n\alpha \nonumber \\
    & \overset{(b)}{<} 0. \nn
\end{align}

Here $(a)$ follows from definition of $\beta$ in \Cref{thm:coeff_bound}; $(b)$ follows from $\gnrhov^{(v')} < -0.5\beta -4\sqrt{\frac{\log (6n^2/\delta)}{d}}n\alpha$. Now using the fact that $\inner{\lmweightt^{(v')}}{\bxi_{k,i}}  < 0$ we have $\sigma'\left( \inner{\lmweightt^{(v')}}{\bxi_{k,i}}\right) = 0$,  which implies $\gnrhov^{(v'+1)} = \gnrhov^{(v')} \geq  -\beta -8\sqrt{\frac{\log (6n^2/\delta)}{d}}n\alpha$ using the induction hypothesis.

\textbf{b).} If $ \gnrhov^{(v')} \geq -0.5\beta -4\sqrt{\frac{\log (6n^2/\delta)}{d}}n\alpha$, then from \eqref{eq:gnrhov_update} we have,

\begin{align}
    \gnrhov^{(v'+1)} &=  \gnrhov^{(v')} + \frac{\eta}{Nm} \lderiv^{(v')} \noisederiv{\lmweightt^{(v')}} \norm{\bxi_{k,i}}\ind{j = - y_{k,i}} \nonumber \\
    & \overset{(a)}{\geq} -0.5\beta -4\sqrt{\frac{\log (6n^2/\delta)}{d}}n\alpha - \frac{3\eta \sigma_p^2d}{2Nm} \nonumber \\
    & \overset{(b)}{\geq} -\beta -8\sqrt{\frac{\log (6n^2/\delta)}{d}}n\alpha.
\end{align}
Here $(a)$ follows from $\abs{\ell'(\cdot)} \leq 1$ and \Cref{lemma:noise_corr}; $(b)$ follows from $\frac{3\eta \sigma_p^2d}{2Nm}  \leq 4\sqrt{\frac{\log (6n^2/\delta)}{d}}n\alpha $
using \Cref{assump:eta}. 

% \ps{Didn't follow this.}

\textbf{Case 2: $(v'+1) \pmod \tau = 0$}.

In this case, from \eqref{eq:gnrhov_update} we have,

\begin{align}
\label{eq:true_update}
     \gnrhov^{(v'+1)} &= \gnrhov^{(v'+1-\tau)} + \frac{\eta}{nm} \underbrace{\sum_{s = 0}^{\tau-1} \lderiv^{(v'+1-\tau + s)} \noisederiv{\lmweightt^{(v'+1-\tau + s)}} \norm{\bxi_{k,i}} \ind{j = - y_{k,i}}}_{:=I_2} \nonumber \\
     & = \gnrhov^{(v'+1-\tau)}  + \frac{\eta}{nm}I_2.
\end{align}

Now suppose instead of doing the update in \eqref{eq:true_update}, we performed the following hypothetical update:

\begin{align}
    \dot{\underline{{\mathbb{P}}}}_{j,r,k,i}^{(v'+1)} &=  \gnrhov^{(v')} + \frac{\eta}{Nm} \lderiv^{(v')} \noisederiv{\lmweightt^{(v')}} \norm{\bxi_{k,i}}\ind{j = - y_{k,i}} \nonumber \\
    & \overset{(a)}{=} \gnrhov^{(v'+1-\tau)} + \frac{\eta}{Nm} \sum_{s = 0}^{\tau-1} \lderiv^{(v'+1-\tau + s)} \noisederiv{\lmweightt^{(v'+1-\tau + s)}} \norm{\bxi_{k,i}} \ind{j = - y_{k,i}} \nonumber \\
     & = \gnrhov^{(v'+1-\tau)}  + \frac{\eta}{Nm}I_2. \nn
\end{align}

Here $(a)$ uses \eqref{eq:gnrhov_update}  for $v = [v'+1-\tau: v']$. From the argument in Case 1 we know that $\dot{\underline{{\mathbb{P}}}}_{j,r,k,i}^{(v'+1)} \geq -\beta -8\sqrt{\frac{\log (6n^2/\delta)}{d}}n\alpha$. Observe that $ \gnrhov^{(v'+1)} \geq \dot{\underline{{\mathbb{P}}}}_{j,r,k,i}^{(v'+1)}$ since $I_2 \leq 0$ and $N \leq n$ and thus $\gnrhov^{(v'+1)} \geq -\beta -8\sqrt{\frac{\log (6n^2/\delta)}{d}}n\alpha$.
\end{proof}

\begin{proof}[Proof of \eqref{eq:rhop_bound}]
We know from \eqref{eq:gprhov_update} that for $j \neq y_{k,i}$, $\gprhov^{(v')} = 0$ for all $0 \leq v' \leq T^*\tau-1$ and hence we focus on the case where $j = y_{k,i}$.

\textbf{Case 1: $(v'+1) \pmod \tau \neq 0$}.

Let $v'_{j,r,k,i}$ be the last iteration such that $v'_{j,r,k,i} \pmod \tau = 0$ and $\gprhov^{(v'_{j,r,k,i} )} \leq 0.5 \alpha$ and let $s$ be the  maximum value in $\{0,1,\dots,\tau-1\}$ such that $\gprhov^{(v'_{j,r,k,i} + s)} \leq 0.5\alpha$. Define $v_{j,r,k,i} = v'_{j,r,k,i} + s$. We see that for all $v > v_{j,r,k,i}$ we have $\gprhov^{(v)} > 0.5\alpha$. Furthermore,
\begin{align}
    \gprhov^{(v'+1)} & \overset{(a)}{\leq} \gprhov^{(v_{j,r,k,i})} - \underbrace{\frac{\eta}{Nm} \lderiv^{(v_{j,r,k,i})} \noisederiv{\lmweightt^{(v_{j,r,k,i})}} \norm{\bxi_{k,i}}\ind{j =  y_{k,i}}}_{L_1} \nonumber\\
    & \hspace{10pt}
    - \underbrace{\sum_{v_{j,r,k,i} < v \leq v'} \frac{\eta}{Nm} \lderiv^{(v)} \noisederiv{\lmweightt^{(v)}} \norm{\bxi_{k,i}}\ind{j =  y_{k,i}}}_{L_2}.
\label{eq:gprhov_ub}
\end{align}

Here $(a)$ uses the fact that we are avoiding the scaling down by a factor of $\frac{1}{K}$ which occurs at every $v \pmod \tau = 0$ (see \eqref{eq:gprhov_update}) for $v'_{j,r,k,i} < v \leq v'$.

We know $\gprhov^{(v_{j,r,k,i})} \leq 0.5\alpha$. We can bound $L_1$ and $L_2$ as follows:

\begin{align}
    L_1 \overset{(a)}{\leq} \frac{\eta}{Nm }\norm{\bxi_{k,i}} \overset{(b)}{\leq} \frac{3\eta\sigma_p^2d}{2Nm} \overset{(c)}{\leq} 1 \overset{(d)}{\leq} 0.25\alpha. \nn
\end{align}

Here $(a)$ uses $|\ell'(\cdot)| \leq 1, \sigma'(\cdot) \leq 1$; $(b)$ uses \Cref{lemma:noise_corr}; $(c)$ uses \Cref{assump:eta}; $(d)$ uses $T^*\tau \geq e$.

Now note that for $v_{j,r,k,i} < v \leq v'$  since $\gprhov^{(v)} \geq 0.5\alpha$ we have,
\begin{align}
    \inner{\lmweightt^{(v)}}{\bxi_{k,i}} & \overset{(a)}{\geq}  \inner{\lmweight^{(0)}}{\bxi_{k,i}} + \gprhov^{(v)} - 4\sqrt{\frac{\log (6n^2/\delta)}{d}}n\alpha \nonumber \\
    & \overset{(b)}{\geq} -0.5\beta + 0.5\alpha -  4\sqrt{\frac{\log (6n^2/\delta)}{d}}n\alpha \nonumber \\
    & \overset{(c)}{\geq} 0.25\alpha.
    \label{eq:noise_ip_pos}
\end{align}
Here $(a)$ follows from \Cref{lemma:inner_prod_bound}, $(b)$ follows from the definition of $\beta$ (see \Cref{thm:coeff_bound}) and $\gprhov^{(v)} \geq 0.5\alpha$, $(c)$ follows from $\beta \leq \frac{1}{12} \leq 0.1\alpha$ and $4\sqrt{\frac{\log (6n^2/\delta)}{d}}n\alpha \leq 0.2\alpha$ using \Cref{assump:d}.

Substituting the bound above in $L_2$ we have,
\begin{align}
    |L_2| & \overset{(a)}{\leq} \sum_{v_{j,r,k,i} < v \leq v'} \frac{\eta}{Nm} \exp\left(-\inner{\lmweightt^{(v)}}{\bxi_{k,i}} + 0.5\right)\noisederiv{\lmweightt^{(v)}} \norm{\bxi_{k,i}}\ind{j =  y_{k,i}} \nonumber \\
    & \overset{(b)}{\leq} \sum_{v_{j,r,k,i} < v \leq v'}\frac{2\eta}{Nm}\exp\left(-\inner{\lmweightt^{(v)}}{\bxi_{k,i}}\right)\norm{\bxi_{k,i}} \\
    & \overset{(c)}{\leq} \sum_{v_{j,r,k,i} < v \leq v'}\frac{2\eta}{Nm}\exp(-0.25\alpha) \frac{3\sigma_p^2d}{2} \nonumber \\
    & =  \frac{2\eta (v' - v_{j,r,k,i}-1)}{Nm} \exp(-\log T^*\tau)\frac{3\sigma_p^2d}{2} \nonumber \\
    & \leq \frac{2\eta(T^*\tau)}{Nm}\exp(-\log T^*\tau)\frac{3\sigma_p^2d}{2} \nonumber \\
    & = \frac{3\eta\sigma_p^2d}{Nm} \nonumber \\ 
    & \overset{(d)}{\leq} 0.25\alpha. \nn
\end{align}

For $(a)$ we use \Cref{lemma:max_deriv};  for $(b)$ we use $\exp(0.5) \leq 2$ and $\inner{\lmweightt^{(v)}}{\bxi_{k,i}} \geq 0$ from \eqref{eq:noise_ip_pos}, $(c)$ follows from \Cref{lemma:noise_corr} and \eqref{eq:noise_ip_pos}; $(d)$ follows from \Cref{assump:eta}. 

Thus substituting the bounds for $L_1$ and $L_2$ we have,
\begin{align}
     \gprhov^{(v'+1)} \leq \alpha, \nn
\end{align}
which completes our proof.

\textbf{Case 2: $(v'+1) \pmod \tau = 0$}.

Suppose instead of doing the update in \eqref{eq:gprhov_update}, we performed the following hypothetical update

\begin{align}
\dot{\overline{{\mathbb{P}}}}_{j,r,k,i'}^{(v'+1)} &=  \gprhov^{(v')} - \frac{\eta}{Nm} \lderiv^{(v')} \noisederiv{\lmweightt^{(v')}} \norm{\bxi_{k,i}}\ind{j = y_{k,i}}. 
\end{align}

From the argument in Case 1 we know that $\dot{\overline{{\mathbb{P}}}}_{j,r,k,i'}^{(v'+1)} \leq \alpha$. Observe that $ \gprhov^{(v'+1)}  \leq \dot{\overline{{\mathbb{P}}}}_{j,r,k,i'}^{(v'+1)}$ and thus $\gprhov^{(v'+1)} \leq \alpha$. 
\end{proof}

\begin{proof}[Proof of \eqref{eq:gamma_bound}]
This part bounds $\ggamv^{(v'+1)}$. To do so we show that the growth of $\ggamv^{(v'+1)}$ is upper bounded by the growth of $\overline{\mathbb{P}}_{y_{k,1},r^*,k,1}^{(v'+1)}$ for any $r^* \in S_{k,1}^{(0)}$, that is,
\begin{align}
\label{eq:ratio_ub}
    \frac{\ggamv^{(v'+1)}}{\overline{\mathbb{P}}_{y_{k,1},r^*,k,1}^{(v'+1)}} \leq C'\effsnr. \nn
\end{align}
We will again use a proof by induction. We first argue the base case of our induction. Since $r^* \in S_{k,1}^{(0)} \subseteq S_{k,1}^{(v)}$, so, 
\begin{align*}
    \overline{\mathbb{P}}_{y_{k,1}, r^*, k, 1}^{(1)} &= \underbrace{\overline{\mathbb{P}}_{y_{k,1}, r^*, k, 1}^{(0)}}_{=0} - \frac{\eta}{Nm} {\ell'}_{k,1}^{(0)} \underbrace{\sigma' \left( \left\langle \bw_{y_{k,1},r^*,k}^{(0)}, \bxi_{k,1} \right\rangle \right)}_{=1 (\because r^* \in S_{k,1}^{(0)})} \norm{\bxi_{k,1}} \\
    &= \frac{\eta \norm{\bxi_{k,1}}}{Nm} \left( -{\ell'}_{k,1}^{(0)} \right) \overset{(a)}{\geq} \frac{\eta \sigma_p^2 d}{2Nm},
\end{align*}
where $(a)$ follows from \Cref{lemma:noise_corr}.
On the other hand,
\begin{align*}
    \ggamv^{(1)} &= \underbrace{\ggamv^{(0)}}_{=0} - \frac{\eta}{Nm} \sum_{i \in [N]} \lderiv^{(0)} \signalderiv{\lmweight^{(0)}} \norm{\bmu} \leq \frac{\norm{\bmu} \eta}{m}.
\end{align*}
Therefore,
\begin{align*}
    \frac{ \ggamv^{(1)}}{\overline{\mathbb{P}}_{y_{k,1},r^*,k,1}^{(1)}} \leq \frac{2 N \norm{\bmu}}{\sigma_p^2 d} \leq C'\effsnr,
\end{align*}
if $C' \geq 2$.
Now assuming \eqref{eq:ratio_ub} holds at $v'$ we have the following cases for $(v'+1)$.
\begin{align}
    \frac{\ggamv^{(v)}}{\overline{\mathbb{P}}_{y_{k,1},r^*,k,1}^{(v)}} \leq C'\effsnr. \nn
\end{align}

\textbf{Case 1: $(v'+1) \pmod \tau \neq 0$}.
From \eqref{eq:ggamv_update} 
we have,
\begin{align}
    \ggamv^{(v'+1)} &= \ggamv^{(v')} + \frac{\eta}{Nm}\sum_{i \in [N]} (-\lderiv^{(v')}) \signalderiv{\lmweightt^{(v')}}\norm{\bmu} \nonumber \\
    & \overset{(a)}{\leq} \ggamv^{(v')} + \frac{\eta C_2}{m}(-\lderivone^{(v')})\norm{\bmu} \nonumber \\
\end{align}
where $(a)$ follows from part $(3)$ in \Cref{lemma:bounded_deriv_frac}.
At the same time since $\inner{\bw_{y_{k,1},r^*,k}^{(v)}}{\bxi_{k,1}} \geq 0 $ for any $r^* \in S_{k,1}^{(0)}$ and for all $0 \leq v \leq T^*\tau-1$, we have from \eqref{eq:gprhov_update}:
\begin{align}
    \overline{\mathbb{P}}_{y_{k,1},r^*,k,1}^{(v'+1)} &= \overline{\mathbb{P}}_{y_{k,1},r^*,k,1}^{(v')} + \frac{\eta}{Nm} (-\lderivone^{(v')})\norm{\bxi_{k,1}} \nn \\
    & \overset{(a)}{\geq} \overline{\mathbb{P}}_{y_{k,1},r^*,k,1}^{(v')} + \frac{\eta}{Nm} (-\lderivone^{(v')})\frac{\sigma_p^2d}{2}, \nn
\end{align}
where $(a)$ follows from \Cref{lemma:noise_corr}.

Thus,
\begin{align}
    \frac{ \ggamv^{(v'+1)}}{\overline{\mathbb{P}}_{y_{k,1},r^*,k,1}^{(v'+1)}} &\leq \max \left\{  \frac{ \ggamv^{(v')}}{\overline{\mathbb{P}}_{y_{k,1},r^*,k,1}^{(v')}}, \frac{2C_2N\norm{\bmu}}{\sigma_p^2 d}\right\} \overset{(a)}{\leq} \max\{C'\effsnr, 2C_2\effsnr \} \overset{(b)}{\leq} C'\effsnr. \nn
\end{align}
Here $(a)$ follows from the definition of $\effsnr$; $(b)$ follows from setting $C' = 2C_2$.

\textbf{Case 2: $(v'+1) \pmod \tau = 0$}.

We have from \eqref{eq:ggamv_update},
\begin{align}
    \ggamv^{(v'+1)} &= \ggamv^{(v'+1-\tau)} +\frac{\eta}{nm}\sum_{s=0}^{\tau-1}\sum_{k'}\sum_{i \in [N]} (-\lderivk^{(v'+1-\tau+s)}) \signalderiv{\lmweightt^{(v-\tau+s)}}\norm{\bmu} \nonumber \\
    & \overset{(a)}{\leq} \ggamv^{(v'+1-\tau)} + \frac{\eta C_2}{m}\sum_{s=0}^{\tau-1}(-\lderivone^{(v'+1-\tau+s)})\norm{\bmu}, \nonumber
\end{align}
where $(a)$ follows from part $(3)$ in \Cref{lemma:bounded_deriv_frac}.
At the same time since $\inner{\bw_{y_{k,1},r^*,k}^{(v)}}{\bxi_{k,1}} \geq 0 $ for any $r^* \in S_{k,1}^{(0)}$ and for all $0 \leq v \leq T^*\tau-1$, we have from \eqref{eq:gprhov_update},
\begin{align}
    \overline{\mathbb{P}}_{y_{k,1},r^*,k,1}^{(v'+1)} &= \overline{\mathbb{P}}_{y_{k,1},r^*,k,1}^{(v'+1-\tau)} + \frac{\eta}{nm}\sum_{s=0}^{\tau-1} (-\lderivone^{(v'+1-\tau+s)})\norm{\bxi_{k,1}} \nonumber \\
    & \overset{(a)}{\geq} \overline{\mathbb{P}}_{y_{k,1},r^*,k,1}^{(v'+1-\tau)} + \frac{\eta}{nm} \sum_{s=0}^{\tau-1}(-\lderivone^{(v'+1-\tau+s)})\frac{\sigma_p^2d}{2}, \nn
\end{align}
where $(a)$ follows from \Cref{lemma:noise_corr}.
Thus,
\begin{align}
    \frac{ \ggamv^{(v'+1)}}{\overline{\mathbb{P}}_{y_{k,1},r^*,k,1}^{(v'+1)}} &\leq \max \left\{  \frac{ \ggamv^{(v'+1-\tau)}}{\overline{\mathbb{P}}_{y_{k,1},r^*,k,1}^{(v'+1-\tau)}}, \frac{2C_2n\norm{\bmu}}{\sigma_p^2 d}\right\} \overset{(a)}{\leq} \max\{C'\effsnr, 2C_2\effsnr \} \overset{(b)}{\leq} C'\effsnr. \nn
\end{align}
Here $(a)$ follows from the definition of $\effsnr$; $(b)$ follows from setting $C' = 2C_2$.
Thus we have shown $\ggamv^{(v'+1)} \leq C'\effsnr\overline{\mathbb{P}}_{y_{k,1},r^*,k,1}^{(v'+1)} \leq C'\effsnr \alpha$ where the last inequality follows from $\overline{\mathbb{P}}_{y_{k,1},r^*,k,1}^{(v'+1)} \leq \alpha$. 
\end{proof}

Now that we have proved \Cref{thm:coeff_bound}, that is, \eqref{eq:rhop_bound}, \eqref{eq:rhov_bound} and \eqref{eq:gamma_bound} hold for all $0 \leq v \leq T^*\tau-1$, we state a simple proposition that extends the result in \Cref{lemma:bounded_deriv_frac} for all $0 \leq v \leq T^*\tau-1$.

\begin{proposition}
\label{prop:bounded_deriv_frac}
Under assumptions, for all $0 \leq v \leq T^*\tau-1$ we have
\begin{enumerate}[leftmargin=*]
    \item \label{prop_bounded_deriv_frac_1} $\frac{1}{m}\sum_{r=1}^m \left[ \overline{\mathbb{P}}_{y_{k,i},r,k,i}^{(v)}  - \overline{\mathbb{P}}_{y_{k',i'},r,k',i'}^{(v)} \right] \leq \kappa$ for all $k,k' \in [K], i,i' \in [N]$.
    \item \label{prop_bounded_deriv_frac_2} $y_{k,i}f(\lmt^{(v)}, \bx_{k,i})-  y_{k',i'}f(\widetilde{\bW}_{k'}^{(v)},  \bx_{k',i'}) \leq C_1$ for all $k, k' \in [K]$ and $i,i' \in [N]$.
    \item \label{prop_bounded_deriv_frac_3} $\frac{\lderivd^{(v)}}{\lderiv^{(v)}} \leq C_2 = \exp(C_1)$ for all $k,k' \in [K]$ and $i, i' \in [N]$.
    \item \label{prop_bounded_deriv_frac_4} $S_{k,i}^{(0)} \subseteq S_{k,i}^{(v)}$ where $S_{k,i}^{(v)} := \left\{r \in [m]: \inner{\widetilde{\bw}_{y_{k,i},r,k}^{(v)}}{\bxi_{k,i}} \geq 0 \right\}$, and hence $\abs{S_{k,i}^{(v)}} \geq 0.4m$ for all $k \in [K], i \in [N]$.
    \item \label{prop_bounded_deriv_frac_5} $\tilde{S}_{j,r}^{(0)} \subseteq \tilde{S}_{j,r}^{(v)}$ where $\tilde{S}_{j,r}^{(v)} := \left\{k \in [K], i \in [N]: y_{k,i} = j, \inner{\widetilde{\bw}_{j,r,k}^{(v)}}{\bxi_{k,i}} \geq 0 \right\}$, and hence $\abs{\tilde{S}_{j,r}^{(v)}} \geq \frac{n}{8}$.
\end{enumerate}
Here we take $\kappa = 5$ and $C_1 = 6.75$.
\end{proposition}

\subsection{First Stage of Training.}
\label{subsec:first_stage}
Define, 
\begin{align}
    T_1 = \frac{C_3nm}{\eta \sigma_p^2 d\tau}
\label{eq:T_1_defn}
\end{align}
where $C_3 = \Theta(1)$ is some large constant. In this stage, our goal is to show that $\overline{P}_{y_{k,i},r^*,k,i}^{(T_1)} \geq 2$ for all $r^*$ such that $r^* \in S_{k,i}^{(0)} := \left\{ r \in [m]: \inner{\bw_{y_{k,i}, r^*}^{(0)}}{\bxi_{k,i}} \geq  0\right\}$. To do so, we first introduce the following lemmas.

\begin{lemma}
\label{lemma:max_signal_coeff}
For all $0 \leq t \leq T_1-1$ and $0 \leq s \leq \tau-1$ we have,
\begin{align}
    \max_{j,r,k} \left\{\ggam^{(t)} + \lgam^{(t,s)}\right\} \leq  \frac{C_3n\norm{\bmu}}{\sigma_p^2d} = \bigO{1}.  \nn
\end{align}
\end{lemma}

\begin{proof}
We have,
\begin{align}
    \ggam^{(t)} + \lgam^{(t,s)}  
    & =  -\frac{\eta}{nm}\sum_{t' = 0}^{t-1} \sum_{k} \sum_{i \in [N]} \sum_{s=0}^{\tau -1} \lderiv^{(t',s)} \signalderiv{\lmweight^{(t',s)}} \norm{\bmu} \nonumber \\
    & \hspace{10pt} -\frac{\eta}{Nm}\sum_{s'=0}^{s} \sum_{i \in [N]} \lderiv^{(t,s')} \signalderiv{\lmweight^{(t,s')}}\norm{\bmu} \nonumber \\
    & \overset{(a)}{\leq} -\frac{\eta}{nm}\sum_{t' = 0}^{t-1} \sum_{k} \sum_{i \in [N]} \sum_{s=0}^{\tau -1} \lderiv^{(t',s)} \norm{\bmu} -\frac{\eta}{Nm}\sum_{s'=0}^{s} \sum_{i \in [N]} \lderiv^{(t,s')} \norm{\bmu} \nonumber \\
    & \overset{(b)}{\leq} \frac{\eta (t+1) \tau \norm{\bmu}}{m} \nonumber \\
    & \leq \frac{\eta T_1 \tau \norm{\bmu}}{m} \nonumber \\
    & = \frac{C_3n\norm{\bmu}}{\sigma_p^2d} \nonumber\\
    & \overset{(c)}{=} \bigO{1}. \nn
\end{align}
Here $(a)$ follows from $\sigma'(\cdot) \in \{0,1\}$, $(b)$ follows from $\abs{\ell'(\cdot)} \leq 1$, $(c)$ follows from \Cref{assump:d}.
\end{proof}

\begin{lemma}
\label{lemma:max_noise_coeff}
For all $0 \leq t \leq T_1-1$ and $0 \leq s \leq \tau-1$ we have,
\begin{align}
    \max_{j,r,k,i} \left\{\gprho^{(t)} + \lprho^{(t,s)}\right\} = \bigO{1}. \nn
\end{align}
\end{lemma}

\begin{proof}
We have from \eqref{eq:overbar_rho_jrki} and \eqref{eq:overbar_P_jrki},
\begin{align}
    \gprho^{(t)} + \lprho^{(t,s)} &=  -\frac{\eta}{nm}\sum_{t' = 0}^{t-1}\sum_{s=0}^{\tau -1} \lderiv^{(t',s)}\noisederiv{\lmweightt^{(v')}}\norm{\bxi_{k,i}}\ind{y_{k,i} = j} \nonumber \\
    & \hspace{10pt} - \frac{\eta}{Nm}\sum_{s'=0}^{s}\lderiv^{(t,s')}\noisederiv{\lmweight^{(t,s')}}\norm{\bxi_{k,i}}\ind{y_{k,i} = j} \nn \\
    & \overset{(a)}{\leq} -\frac{\eta}{nm}\sum_{t' = 0}^{t-1}\sum_{s=0}^{\tau -1} \lderiv^{(t',s)}\norm{\bxi_{k,i}} - \frac{\eta}{Nm}\sum_{s'=0}^{s}\lderiv^{(t,s')}\norm{\bxi_{k,i}} \nonumber \\
    & \leq \frac{\eta (t+1) \tau \norm{\bxi_{k,i}}}{Nm} \nonumber \\
    & \overset{(b)}{\leq} \frac{3\eta T_1 \tau \sigma_p^2d}{2Nm} \nonumber \\
    & \leq \frac{3C_3n}{2N} \nonumber \\
    & = \bigO{1}. \nn
\end{align}
Here $(a)$ follows from $\sigma'(\cdot) \leq 1$, $(b)$ follows from $t \leq T_1-1$ and \Cref{lemma:noise_corr}.
\end{proof}

\begin{lemma}
\label{lemma:max_func_output}
For any $k \in [K]$ and $i \in [N]$,  we have $F_j(\bW_{j,k}^{(t,s)},\bx_{k,i}) = \bigO{1}$ for all $j \in \{\pm 1\}$, $0 \leq t \leq T_1-1$ and $0 \leq s \leq \tau-1$.
\end{lemma}

\begin{proof}
We have,
\begin{align}
    & F_j(\bW_{j,k}^{(t,s)}, \bx_{k,i}) \nn \\
    &= \frac{1}{m}\sum_{r=1}^m \left[ \sigma\left(\inner{\lmweight^{(t,s)}}{y_{k,i}\bmu} \right) + \sigma\left( \inner{\lmweight^{(t,s)}}{\bxi_{k,i}} \right) \right] \nonumber\\
    & \overset{(a)}{\leq} \frac{1}{m}\sum_{r=1}^m \left[ \abs{\inner{\lmweight^{(t,s)}}{y_{k,i}\bmu}}  +  \abs{\inner{\lmweight^{(t,s)}}{\bxi_{k,i}}} \right] \nonumber\\
    & \overset{(b)}{\leq} \frac{1}{m}\sum_{r=1}^m \left[ \abs{\inner{\gmweight^{(0)}}{\bmu}} + \ggam^{(t)} + \lgam^{(t,s)}  + \abs{\inner{\gmweight^{(0)}}{\bxi_{k,i}}} + \gprho^{(t)} + \lprho^{(t,s)} + 4\sqrt{\frac{\log (6n^2/\delta)}{d}}n\alpha \right] \nonumber\\
    & \leq 5 \max_{r \in [m]} \left\{ \abs{\inner{\gmweight^{(0)}}{\bmu}}, \ggam^{(t)} + \lgam^{(t,s)},  \abs{\inner{\gmweight^{(0)}}{\bxi_{k,i}} }, \gprho^{(t)} + \lprho^{(t,s)}, 4\sqrt{\frac{\log (6n^2/\delta)}{d}}n\alpha \right\} \nonumber\\
    & \overset{(c)}{\leq} 5 \max_{r \in [m]} \left\{ \beta, \ggam^{(t)} + \lgam^{(t,s)}, \gprho^{(t)} + \lprho^{(t,s)}, 4\sqrt{\frac{\log (6n^2/\delta)}{d}}n\alpha \right\} \nonumber \\
    & \overset{(d)}{=} \bigO{1}. \nn
\end{align}
Here $(a)$ follows from $\sigma(z) \leq \abs{z}$, $(b)$ follows from \Cref{lemma:inner_prod_bound}, $(c)$ follows from the definition of $\beta$, $(d)$ follows from \Cref{lemma:beta_mag}, \Cref{lemma:max_signal_coeff} and \Cref{lemma:max_noise_coeff}. 
\end{proof}

\begin{lemma}
\label{lemma:activated_noise_filters}
For all $t \geq T_1$ and $0 \leq s \leq \tau-1$ we have, 
\begin{align}
  \overline{P}_{y_{k,i},r^*,k,i}^{(t)} + \overline{\rho}_{y_{k,i}, r^*, k, i}^{(t,s)} \geq \overline{P}_{y_{k,i},r^*,k,i}^{(T_1)} \geq 2.
\end{align}
where $r^* \in S_{k,i}^{(0)} := \left\{ r \in [m]: \inner{\bw_{y_{k,i}, r, k}^{(0)}}{\bxi_{k,i}} > 0\right\}.$ 
\end{lemma}

\begin{proof}
First note that from \Cref{lemma:max_func_output}, we have for any $k \in [K]$, $i \in [N]$, $F_{+1}(\bW_{+1,k}^{(t,s)},\bx_{k,i}), F_{-1}(\bW_{-1,k}^{(t,s)},\bx_{k,i}) = \bigO{1}$ for all $t \in \{0,1,\dots, T_1-1\}$, $s \in \{0,1,\dots, \tau-1\}$. Thus there exists a positive constant $C$ such that for all $0 \leq t \leq T_1-1$ and $0 \leq s \leq \tau-1$ we have,
\begin{align}
\label{eq:first_stage_min_deriv}
  -\lderiv^{(t',s)} \geq C.
\end{align}

Next we know from \Cref{prop:bounded_deriv_frac} part \ref{bounded_deriv_frac_4} that,
\begin{align}
     \inner{\bw_{y_{k,i}, r^*, k}^{(t, s)}}{\bxi_{k,i}} > 0 \hspace{10pt} \text{ for all } 0 \leq t \leq T_1-1, 0 \leq s \leq \tau-1, \nn
\end{align}
where $r^* \in S_{k,i}^{(0)} := \left\{ r \in [m]: \inner{\bw_{y_{k,i}, r, k}^{(0)}}{\bxi_{k,i}} > 0\right\}.$ 
This implies that for $t \geq T_1$,
\begin{align}
\label{eq:min_gprho_pos}
    \overline{P}_{y_{k,i},r^*,k,i}^{(t)} + \overline{\rho}_{y_{k,i}, r^*, k, i}^{(t,s)}
    &\geq \overline{P}_{y_{k,i},r^*,k,i}^{(T_1)} \nonumber \\
    & \overset{(a)}{=} -\sum_{t'=0}^{T_1}\frac{\eta}{nm}\sum_{s=0}^{\tau-1}\lderiv^{(t',s)}\cdot\norm{\bxi_{k,i}} \nn \\
    & \overset{(b)}{\geq} \frac{\eta C T_1 \tau \sigma_p^2 d}{2nm} \nonumber \\
    & \overset{(b)}{\geq} 2.
\end{align}

Here $(a)$ follows from \eqref{eq:overbar_P_jrki}; $(b)$ follows from \eqref{eq:first_stage_min_deriv} and \Cref{lemma:noise_corr}; $(b)$ follows from the definition of $T_1$ in \eqref{eq:T_1_defn} and setting $C_3 = 4/C$.

\end{proof}

\subsection{Second Stage of Training}
\label{subsec:2nd_stage}

In the first stage we have shown that for any $k \in [K]$ and $i \in [N]$,  $\overline{P}_{y_{k,i},r^*,k,i}^{(t)} + \overline{\rho}_{y_{k,i}, r^*, k, i}^{(t,s)} \geq 2$ for all $t \geq T_1$ and $s \in [0:\tau-1]$. Our goal in the second stage is to show that for every round in $T_1 \leq t \leq T^*-1$, the loss of the global model is decreasing. To do so, we will show that our objective  satisfies the following property 
\begin{align}
    \inner{\nabla L_k(\lm^{(t,s)})}{\lm^{(t,s)} - \bW^*} \geq L_k(\lm^{(t,s)}) - \frac{\epsilon}{2\tau}, \nn
\end{align}
where $\bW^*$ is defined as follows.
\begin{align}
\label{eq:W_star_def}
    \bw_{j,r}^* := \bw_{j,r}^{(0)} + 5\log(2\tau/\epsilon)\left[\sum_{k}\sum_{i \in [N]} \ind{j = y_{k,i}} \frac{\bxi_{k,i}}{\norm{\bxi_{k,i}}}\right].
\end{align}

Using this we can easily show that the loss of the global model is decreasing in every round leading to convergence.
We now state and prove some intermediate lemmas.

\begin{lemma}
\label{lemma:frob_distance}
Under \Cref{assum:main_assump}, we have 
\begin{align}
    \normt{ \bW^{(T_1)} - \bW^*} = \bigO{ \sqrt{\frac{mn}{\sigma_p^2 d}} \log(\tau/\epsilon)}. \nn
\end{align}
\end{lemma}

\begin{proof}
\begin{align}
        & \normt{\bW^{(T_1)} - \bW^*} \leq \normt{\bW^{(T_1)}-\bW^{(0)}} + \normt{\bW^* - \bW^{(0)}} \nonumber \\
        & \overset{(a)}{=} \bigO{m^{1/2}\normt{\bmu}^{-1}\max_{j,r} \ggam^{(T_1)}} + \bigO{m^{1/2}n^{1/2} \sigma_p^{-1}d^{-1/2}\max_{j, r, k, i}\left\{ \gprho^{(T_1)}, \gnrho^{(T_1)}\right\}} \nonumber\\
        & + \bigO{m^{1/2}n\sigma_p^{-1}d^{-3/4}} + \normt{\bW^* - \bW^{(0)}} \nonumber \\
        & \overset{(b)}{=} \bigO{m^{1/2} n \normt{\bmu} \sigma_p^{-2}d^{-1}} + \bigO{m^{1/2}n^{1/2} \sigma_p^{-1}d^{-1/2}} + \bigO{m^{1/2}n^{1/2} \log(\tau/\epsilon)\sigma_p^{-1}d^{-1/2}} \nonumber \\
        & \overset{(c)}{=} \bigO{m^{1/2}n^{1/2} \sigma_p^{-1}d^{-1/2}} + \bigO{m^{1/2}n^{1/2} \log(\tau/\epsilon)\sigma_p^{-1}d^{-1/2}} \nonumber \\
        & = \bigO{m^{1/2}n^{1/2} \log(\tau/\epsilon)\sigma_p^{-1}d^{-1/2}}. \nn
\end{align}
Here $(a)$ follows from the following argument:
\begin{align*}
    & \norm{\bW^{(T_1)}-\bW^{(0)}} \nn \\
    &= \sum_{j,r} \norm{\ggam^{(T_1)}\cdot\normt{\bmu}^{-2} \cdot \bmu} + \sum_{j,r} \norm{\sum_{k=1}^K\sum_{i \in [N]}\grho^{(T_1)}\cdot\normt{\bxi_{k,i}}^{-2}\cdot\bxi_{k,i}} \\
    & \quad + 2m \underbrace{\left\langle \ggam^{(t)}\cdot\normt{\bmu}^{-2}\bmu, \sum_{k=1}^2\sum_{i \in [N]}\grho^{(t)}\cdot\normt{\bxi_{k,i}}^{-2}\cdot\bxi_{k,i} \right\rangle}_{=0} \\
    &= \mc O \left( \frac{m}{\normt{\bmu}^{2}} \max_{j,r} (\ggam^{(t)})^2 \right) + \mc O \left( \frac{mn}{\normt{\bxi_{k,i}}^{2}} \max_{j,r,k,i} (\grho^{(t)})^2 \right) + \mc O \left( mn^2 \max_{k,k,k',i'} \frac{\left\langle \bxi_{k,i}, \bxi_{k',i'} \right\rangle}{\normt{\bxi_{k,i}}^4} \right) \\
    &= \mc O \left( \frac{m}{\normt{\bmu}^{2}} \max_{j,r} (\ggam^{(t)})^2 \right) + \mc O \left( \frac{mn}{\normt{\bxi_{k,i}}^{2}} \max_{j,r,k,i} (\grho^{(t)})^2  \right) + \mc O \left( \frac{mn^2}{\sigma_p^2 d^{3/2}} \right) \\
\end{align*}
where the last equality follows from \Cref{lemma:noise_corr}. Getting back to our proof, we see that $(b)$ follows from \Cref{lemma:max_signal_coeff}, \Cref{lemma:max_noise_coeff} and definition of $\bW^*$ in \eqref{eq:W_star_def}; $(c)$ follows from \Cref{assump:d}.
\end{proof}

\begin{lemma}
\label{eq:grad_greater_than_eps}
For any $k \in [K]$, $i \in [N]$ we have for all $t \in  \{T_1, T_1+1,\dots, T^*-1\}$, $s \in \{0,1,\dots, \tau-1\}$,
\begin{align}
    y_{k,i} \langle \nabla f(\lm^{(t,s)}, \bx_{k,i}), \bW^* \rangle \geq \log (2\tau/\epsilon). \nn
\end{align}
\end{lemma}

\begin{proof}
\begin{align}
    & y_{k,i} \langle \nabla f(\lm^{(t,s)},  \bx_{k,i}), \bW^* \rangle  \nonumber \\
    %%%%
    %%%%
    %%%%
    & = \frac{1}{m} \sum_{j,r} \sigma'\left( \inner{\lmweight^{(t,s)}}{y_{k,i}\bmu}\right)\inner{\bmu}{j\woptw_{j,r}} + \frac{1}{m} \sum_{j,r} \sigma'\left( \inner{\lmweight^{(t,s)}}{\bxi_{k,i}}\right)\inner{y_{k,i}\bxi_{k,i}}{j\woptw_{j,r}} \nonumber \\
    %%%%
    %%%%
    %%%%
    & = \frac{1}{m} \sum_{j,r} \sum_{k',i'} \sigma'\left( \inner{\lmweight^{(t,s)}}{\bxi_{k,i}}\right)5\log(2/\epsilon)\ind{j = y_{k',i'}} \frac{\inner{y_{k,i}\bxi_{k,i}}{j\bxi_{k',i'}}}{\norm{\bxi_{k',i'}}} \nonumber\\
     & \hspace{10pt} + 
     \frac{1}{m} \sum_{j,r} \sum_{k',i'} \sigma'\left( \inner{\lmweight^{(t,s)}}{y_{k,i}\bmu}\right)5\log(2/\epsilon)\ind{j = y_{k',i'}} \frac{\inner{\bmu}{j\bxi_{k',i'}}}{\norm{\bxi_{k',i'}}} \nonumber\\
      & \hspace{10pt} +
      \frac{1}{m}\sum_{j,r} \sigma'\left( \inner{\lmweight^{(t,s)}}{y_{k,i}\bmu}\right) \inner{\bmu}{j \gmweight^{(0)}} +   \frac{1}{m}\sum_{j,r} \sigma'\left( \inner{\lmweight^{(t,s)}}{\bxi_{k,i}}\right) \inner{y_{k,i}\bxi_{k,i}}{j\gmweight^{(0)}} \nn \\
      %%%%
      %%%%
      %%%%
      & \geq \underbrace{\frac{1}{m}\sum_{j = y_{k,i}, r} \sigma'\left( \inner{\lmweight^{(t,s)}}{\bxi_{k,i}}\right)5\log(2\tau/\epsilon)}_{I_1} \nonumber \\
      & \hspace{10pt} - \underbrace{\frac{1}{m}\sum_{j,r}\sum_{(k',i') \neq (k, i) } \sigma'\left( \inner{\lmweight^{(t,s)}}{\bxi_{k,i}}\right)5\log(2\tau/\epsilon) \frac{\left|\inner{\bxi_{k,i}}{\bxi_{k',i'}}\right|}{\norm{\bxi_{k',i'}}}}_{I_2}  \nn \\
      & \hspace{10pt} -  \underbrace{\frac{1}{m} \sum_{j,r} \sum_{k',i'} \sigma'\left( \inner{\lmweight^{(t,s)}}{y_{k,i}\bmu}\right)5\log(2\tau/\epsilon)\frac{\left|\inner{\bmu}{\bxi_{k',i'}}\right|}{\norm{\bxi_{k',i'}}}}_{I_3} \nonumber \\
      & \hspace{10pt}  - \underbrace{\frac{1}{m}\sum_{j,r} \sigma'\left( \inner{\lmweight^{(t,s)}}{y_{k,i}\bmu}\right) \left|\inner{\bmu}{j \gmweight^{(0)}}\right|}_{I_4}  - \underbrace{\frac{1}{m}\sum_{j,r} \sigma'\left( \inner{\lmweight^{(t,s)}}{\bxi_{k,i}}\right) \left| \inner{y_{k,i}\bxi_{k,i}}{j\gmweight^{(0)}} \right|}_{I_5}. \nn
\end{align}
Now noting that $\sigma'(z) \leq 1$ and $\inner{\bmu}{\bxi_{k,i}} 
 = 0 \hspace{5pt} \forall k \in [K], i \in [N]$ we have the following bounds for $I_2, I_3, I_4, I_5$ using \Cref{lemma:noise_corr}, \Cref{lemma:init_filter_corr} and \Cref{lemma:beta_mag}. 
\begin{align}
    & I_2 = \log(2\tau/\epsilon)\bigo{n\sqrt{\log (n^2/\delta)}/\sqrt{d}}, I_3 = 0, \nn \\
    & I_4 = \bigo{\sqrt{\log (m/\delta)}\cdot \sigma_0 \normt{\bmu}}, I_5 = \bigo{\sqrt{\log (mn/\delta)}\cdot \sigma_0 \sigma_p\sqrt{d}}. \nn
\end{align}
For $I_1$ we know that, $\inner{\bw_{y_{k,i},r^*,k}^{(t,s)}}{\bxi_{k,i}} \geq 0$ $\forall t \in [0: T^*-1], \forall s \in [0:\tau-1]$ (\Cref{lemma:activated_noise_filters} ) and $r^*$ such that $ r^* \in S_{k,i}^{(0)} := \left\{ r \in [m]: \inner{\bw_{y_{k,i}, r,k}^{(0)}}{\bxi_{k,i}} \geq 0\right\}$.
Thus,
\begin{align}
    I_1 & \geq \frac{1}{m} |S_{k,i}^{(0)}|5\log (2\tau/\epsilon) \geq 2 \log (2\tau/\epsilon). \nn
\end{align}
where the last inequality follows from \Cref{lemma:min_size_activated_noise_filters}. 
Applying triangle inequality we have,
\begin{align}
   y_{k,i} \langle \nabla f(\lm^{(t,s)}, \bx_{k,i} ), \bW^* \rangle & \geq I_1 - |I_2| - |I_3| - |I_4| - |I_5| \geq \log (2\tau/\epsilon), \nn
\end{align}
where the last inequality follows from \Cref{assump:d} and \Cref{assump:sigma_0}.
\end{proof}

\begin{lemma} (Lemma D.4 in \cite{kou2023benign})
\label{lemma:grad_mag}
Under assumptions, for $0 \leq t \leq T^*$ and $0 \leq s \leq \tau-1$, the following result holds,
\begin{align}
    \norm{\nabla L_{k}(\bW_k^{(t,s)})} \leq \bigO{\max\left\{ \norm{\bmu},  \sigma_p^2 d\right\}} L_k(\bW_k^{(t,s)}). \nn
\end{align}

\end{lemma}

\begin{lemma}
\label{lemma:pl_convex_prop} 
For all $k \in [K]$, $T_1 \leq t \leq T^*-1$, $0 \leq s \leq \tau-1$ we have,
\begin{align}
    \inner{\nabla L_k(\lm^{(t,s)})}{\lm^{(t,s)} - \bW^*} \geq L_k(\lm^{(t,s)}) - \frac{\epsilon}{2\tau}. \nn
\end{align}
\end{lemma}

\begin{proof}
\begin{align}
    &\inner{\nabla L_k(\lm^{(t,s)})}{\lm^{(t,s)} - \bW^*} \nonumber \\
    & \hspace{10pt} = \frac{1}{N}\sum_{i \in [N]} \lderiv^{(t,s)} \inner{y_{k,i}\nabla f(\lm^{(t,s)},\bx_{k,i})}{\lm^{(t,s)} - \bW^*} \nonumber \\
    & \hspace{10pt} \overset{(a)}{=} \frac{1}{N}\sum_{i \in [N]} \lderiv^{(t,s)} \left[ y_{k,i}f(\lm^{(t,s)},\bx) - y_{k,i}\inner{\nabla f(\lm^{(t,s)},\bx_{k,i})}{\bW^*}\right] \nn \\
    & \hspace{10pt} \overset{(b)}{\geq} \frac{1}{N}\sum_{i \in [N]} \lderiv^{(t,s)} \left[ y_{k,i}f(\lm^{(t,s)},\bx_{k,i}) - \log (2\tau/\epsilon) \right] \nn  \\
    & \hspace{10pt} \overset{(c)}{\geq} \frac{1}{N}\sum_{i \in [N]} \left[ \ell(y_{k,i}f(\lm^{(t,s)},\bx_{k,i})) - \epsilon/2\tau\right] \nonumber \\
    & \hspace{10pt} = L_k(\lm^{(t,s)}) - \frac{\epsilon}{2\tau}. \nn 
\end{align}

Here $(a)$ follows from the property that $\inner{\nabla f(\bW,\bx)}{\bW} = f(\bW,\bx)$ for our two-layer CNN model; $(b)$ follows from \eqref{eq:grad_greater_than_eps} (note that $\lderiv^{(t,s)} \leq 0$), $(c)$ follows from $\ell'(z)(z-z') \geq \ell(z) - \ell(z')$ since $\ell(\cdot)$ is convex and $\log(1+z) \leq z$.
\end{proof}

\begin{lemma}(Local Model Convergence)
\label{lemma:local_model_convergence}
Under assumptions, for all $t \geq T_1$ we have,
\begin{align}
    \norm{\lm^{(t,\tau)} - \wopt} \leq \norm{\gm^{(t)}-\wopt} - \eta \sum_{s=0}^{\tau - 1} L_k(\lm^{(t,s)}) + \eta \epsilon. \nn
\end{align}
\end{lemma}

\begin{proof}
\begin{align}
    & \norm{\lm^{(t,s+1)} - \wopt}  \nonumber \\
    & \hspace{10pt} = \norm{\lm^{(t,s)}-\wopt} -2\eta \inner{\nabla L_k(\lm^{(t,s)})}{\lm^{(t,s)} - \bW^*} + \eta^2  \norm{\nabla L_{k}(\bW_k^{(t,s)})} \nonumber \\
    & \hspace{10pt} \overset{(a)}{\leq} \norm{\lm^{(t,s)}-\wopt} -2\eta L_k(\lm^{(t,s)}) + \frac{\eta \epsilon}{\tau} + \eta^2  \norm{\nabla L_{k}(\bW_k^{(t,s)})} \nonumber \\
    & \hspace{10pt} \overset{(b)}{\leq} \norm{\lm^{(t,s)}-\wopt} -\eta L_k(\lm^{(t,s)}) + \frac{\eta \epsilon}{\tau}, \nn
\end{align}
where $(a)$ follows from \Cref{lemma:pl_convex_prop}; $(b)$ follows from \Cref{lemma:grad_mag} and \Cref{assump:eta}. 
Now starting from $s = \tau-1$ and unrolling the recursion we have,
\begin{align}
     \norm{\lm^{(t,\tau)} - \wopt} \leq \norm{\lm^{(t,0)}-\wopt} - \eta \sum_{s=0}^{\tau - 1} L_k(\lm^{(t,s)}) + \eta \epsilon. \nn
\end{align}
\end{proof}

\subsection{\texorpdfstring{Proof of \Cref{thm:train_loss}}{Proof of Theorem}}
\label{subsec:proof_train_loss}

For any $t \geq T_1$ we have,
\begin{align}
    \norm{\gm^{(t+1)} - \wopt} &= \norm{\sum_{k=1}^K \frac{1}{K} \lm^{(t,\tau)} - \wopt} \nonumber \\
    & \overset{(a)}{\leq} \sum_{k=1}^K \frac{1}{K} \norm{\lm^{(t,\tau)} - \wopt} \nonumber \\
    & \overset{(b)}{\leq} \norm{\gm^{(t)} - \wopt} - \eta \frac{1}{K} \sum_{k=1}^K \sum_{s=0}^{\tau-1} L_k(\lm^{(t,s)}) + \eta\epsilon \nonumber \\
    & \overset{(c)}{\leq} \norm{\gm^{(t)} - \wopt} - \eta \frac{1}{K} \sum_{k=1}^K  L_k(\gm^{(t)}) + \eta \epsilon \nonumber \\
    & = \norm{\gm^{(t)} - \wopt} - \eta L(\gm^{(t)}) + \eta \epsilon, \label{eq:global_model_decrease}
\end{align}
where $(a)$ follows from Jensen's inequality, $(b)$ follows from \Cref{lemma:local_model_convergence}; $(c)$ follows from $\sum_{s=0}^{\tau-1} L_k(\lm^{(t,s)}) \leq L_k(\lm^{(t,0)}) = L_k(\gm^{(t)})$.
From \eqref{eq:global_model_decrease} we get,
\begin{align}
    \eta L(\gm^{(t)}) \leq \norm{\gm^{(t)} - \wopt} - \norm{\gm^{(t+1)} - \wopt} + \eta \epsilon. \nn
\end{align}
Summing over $t= T_1, T_1+1,\dots, T$ and dividing by $\eta(T - T_1 + 1)$ we have,
\begin{align}
\label{eq:main_descent}
    \frac{1}{T - T_1 + 1}\sum_{t=T_1}^{T} L(\gm^{(t)}) \leq \frac{\norm{\gm^{(T_1)} - \wopt}}{\eta(T-T_1 + 1)} + \epsilon,
\end{align}
for all $T_1 \leq T \leq T^*-1$. 
Now \eqref{eq:main_descent} implies that we can find an iterate with training error less than $2\epsilon$ within,
\begin{align}
    T &= T_1 + \frac{\norm{\gm^{(t)} - \wopt} }{\eta \epsilon} = \bigO{\frac{mn}{\eta\sigma_p^{2}d\tau}} + \bigO{\frac{mn \log(\tau/\epsilon)}{\eta\sigma_p^{2}d\epsilon} } \nn
\end{align}
rounds where the last equality follows from the definition of $T_1$ in \eqref{eq:T_1_defn}
and \Cref{lemma:frob_distance}. This completes our proof of \Cref{thm:train_loss}.

\qed

\newpage

\section{Proof of \texorpdfstring{\Cref{thm:test_error}}{Theorem 2}}

We first state some intermediate lemmas that will be used in the proof.

\begin{lemma}
\label{lemma:gamma_ip_always_pos}
Suppose $\inner{\bw_{j,r}^{(t')}}{j\bmu} \geq 0$ for some $t' \geq 0$. Then for all $t \geq t', s \in [0:\tau-1]$, $k \in [K]$, we have $\inner{\bw_{j,r,k
 }^{(t,s)}}{j\bmu} \geq 0$.
\end{lemma}

\begin{proof}
We will use a proof by induction. We will show that our claim holds for $t = t', s \in [0:\tau-1]$ and also $t = (t'+1), s = 0$. Using this fact we can argue that the claim holds for all $t \geq t'$ and $s \in [0:\tau-1]$.

\textbf{Case 1:} First let us look at the local iterations $s \in [0:\tau-1]$ for $t = t'$. From \Cref{lemma:measure_signal_coeff} we have,
\begin{align}
  \inner{\bw_{j,r,k}^{(t', s)}}{j\bmu} &= \inner{\bw_{j,r}^{(t')}}{j\bmu} + \lgam^{(t',s)} \nonumber\\
  & \overset{(a)}{\geq} \inner{\bw_{y,r}^{(t')}}{j\bmu} \nonumber \\
  & \overset{(b)}{\geq} 0, \nn 
\end{align}
where $(a)$ uses $\lgam^{(\cdot, \cdot)} \geq 0$ by definition; $(b)$ uses $\inner{\bw_{j,r}^{(t')}}{j\bmu} \geq 0$.

\textbf{Case 2:} Now let us look at the round update $t = t'+1, s= 0$. We have,

\begin{align}
  \inner{\bw_{j,r,k}^{(t'+ 1, 0)}}{j\bmu} &=   \inner{\bw_{j,r}^{(t'+1)}}{j\bmu} \nonumber\\
  & = \inner{\bw_{j,r}^{(t')}}{j\bmu} + \frac{1}{K}\sum_{i=1}^K \lgam^{(t',\tau)} \nonumber\\
  & \overset{(a)}{\geq} \inner{\bw_{j,r}^{(t')}}{j\bmu} \nonumber \\
  & \overset{(b)}{\geq} 0, \nn 
\end{align}
where $(a)$ uses $\lgam^{(\cdot,\cdot)} \geq 0$ by definition; $(b)$ uses $\inner{\bw_{j,r}^{(t')}}{j\bmu} \geq 0$.
\end{proof}

\begin{lemma}
\label{lemma:gamma_upper_and_lower}
Under \Cref{assum:main_assump}, for any $0 \leq t \leq T^*-1$ we have,

\begin{align}
\label{eq:test_acc_gamma_min_init_pos}
    \ggam^{(t)}  \geq  \ggam^{(t-1)} + \frac{\eta \norm{\bmu}}{4m} \sum_{s=0}^{\tau - 1} \min_{k, i} \abs{\lderiv^{(t-1,s)}}  \hspace{10pt} \text{ if } \inner{\bw_{j,r}^{(t-1)}}{j \bmu} \geq 0,
\end{align}
and,
\begin{align}
\label{eq:test_acc_gamma_min_init_neg}
    \ggam^{(t)} & \geq   \ggam^{(t-1)} + \frac{\eta \norm{\bmu}}{4m} \left( \min_{k, i} \abs{\lderiv^{(t-1,0)}} + h \sum_{s=1}^{\tau-1} \min_{k, i} \abs{\lderiv^{(t-1,s)}} \right)  \hspace{10pt} \text{ if } \inner{\bw_{j,r}^{(0)}}{j \bmu} < 0.
\end{align}
\end{lemma}

\textbf{Proof.}

From \eqref{eq:Gamma_jr} we have the following update equation for $\ggam^{(t)}$,
\begin{align}
     \ggam^{(t)} &= \ggam^{(t-1)} - \frac{\eta}{nm} \sum_{s=0}^{\tau -1} 
 \sum_{k,i} \lderiv^{(t-1,s)} \cdot \signalderiv{\lmweight^{(t-1,s)}}\cdot\norm{\bmu}  \label{eq:gamma_simplified_1}.
\end{align}

\begin{proof}[Proof of \eqref{eq:test_acc_gamma_min_init_pos}]
In this case we know from \Cref{lemma:gamma_ip_always_pos} that
if $\inner{\bw_{j,r}^{(t)}}{j\bmu} \geq 0$, then
\begin{align}
    \inner{\bw_{j,r,k}^{(t,s)}}{j \bmu} \geq 0 \text{ for all } k \in [K], s \in [0:\tau-1].
\end{align}
Using this observation we have from \eqref{eq:gamma_simplified_1},
\begin{align}
    \ggam^{(t)} & \overset{(a)}{\geq}  \ggam^{(t-1)} + \frac{\eta |D_j| \norm{\bmu}}{nm}\sum_{s=0}^{\tau - 1} \min_{(k,i) \in D_j} \abs{\lderiv^{(t-1,s)}} \nonumber \\
    & \overset{(b)}{\geq} \ggam^{(t-1)} + \frac{\eta \norm{\bmu}}{4m}\sum_{s=0}^{\tau - 1} \min_{k, i} \abs{\lderiv^{(t-1,s)}} 
\end{align}
where $(a)$ follows from the definition of $D_j := \{ k \in [K], i \in [N]: y_{k,i} = j\}$; $(b)$ follows from \Cref{lemma:S_j_size} and $\min_{(k,i) \in D_j} \abs{\lderiv^{(t',s)}} \geq \min_{k, i} \abs{\lderiv^{(t',s)}} $.
\end{proof}

\begin{proof}[Proof of \eqref{eq:test_acc_gamma_min_init_neg}]
First let us look at the iteration $s = 0$. In this case we know that $ \inner{\bw_{j,r,k}^{(t-1,0)}}{j \bmu}  = \inner{\bw_{j,r}^{(t-1)}}{j \bmu} < 0$ and thus $\inner{\bw_{j,r}^{(t-1)}}{y_{k,i} \bmu} > 0$ for $y_{k,i} = -j$. Using this observation we have,
\begin{align}
    -\frac{\eta}{nm}\sum_{k,i} \lderiv^{(t-1,0)} \cdot \signalderiv{\lmweight^{(t-1,0)}}\cdot\norm{\bmu} &\geq \frac{\eta\abs{D_{-j}}\norm{\bmu}}{nm} \min_{(k,i) \in D_{-j}} \abs{\lderiv^{(t-1,0)}} \nonumber\\
    & \overset{(a)}{\geq} \frac{\eta\norm{\bmu}}{4m} \min_{k, i} \abs{\lderiv^{(t-1,0)}}  \nn 
\end{align}
where $(a)$ follows from \Cref{lemma:S_j_size} and $\min_{(k,i) \in D_j} \abs{\lderiv^{(t',s)}} \geq \min_{k, i} \abs{\lderiv^{(t',s)}}$.

Now let us look at the case $1 \leq s \leq \tau - 1$. In this case if $\inner{\bw_{j,r,k}^{(t-1,s)}}{j \bmu} < 0$ then,
\begin{align}
\label{eq:min_gamma_s_geq_0}
    -\frac{\eta}{nm}\sum_{i} \lderiv^{(t-1,s)} \cdot \signalderiv{\lmweight^{(t-1,s)}}\cdot\norm{\bmu} &\geq \frac{\eta \abs{D_{-j,k}}\norm{\bmu}}{nm} \min_{(k,i) \in D_{-j,k}} \abs{\lderiv^{(t-1,s)}},
\end{align}
and if $\inner{\bw_{j,r,k}^{(t-1,s)}}{j \bmu} \geq 0$ then,
\begin{align}
    -\frac{\eta}{nm}\sum_{i} \lderiv^{(t-1,s)} \cdot \signalderiv{\lmweight^{(t-1,s)}}\cdot\norm{\bmu} \geq \frac{\eta \abs{D_{j,k}}\norm{\bmu}}{nm} \min_{(k,i) \in D_{j,k}} \abs{\lderiv^{(t-1,s)}}. \nn 
\end{align}
Thus,
\begin{align}
\label{eq:min_gamma_s_geq_1}
  -\frac{\eta}{nm}\sum_{i} \lderiv^{(t-1,s)} \cdot \signalderiv{\lmweight^{(t-1,s)}}\cdot\norm{\bmu} \geq   \frac{\eta \min\{\abs{D_{+,k}}, \abs{D_{-, k}}\}\norm{\bmu}}{nm} \min_{(k,i) \in D_k} \abs{\lderiv^{(t-1,s)}}.
\end{align}

Using the results in \eqref{eq:min_gamma_s_geq_0} and \eqref{eq:min_gamma_s_geq_1} we have,
\begin{align}
    \ggam^{(t)} &\geq \ggam^{(t-1)} + \frac{\eta\norm{\bmu}}{4m} \min_{k, i} \abs{\lderiv^{(t-1,0)}}  + \frac{\eta \norm{\bmu}}{m} \sum_{k}\frac{\min\{\abs{D_{+, k}}, \abs{D_{-, k}}\}}{n} \sum_{s=1}^{\tau-1} \min_{(k,i)} \abs{\lderiv^{(t-1,s)}} \nonumber\\
    & \overset{(a)}{\geq}  \ggam^{(t-1)} + \frac{\eta\norm{\bmu}}{4m} \left( \min_{k, i} \abs{\lderiv^{(t-1,0)}} + h\sum_{s=1}^{\tau-1} \min_{k, i} \abs{\lderiv^{(t-1,s)}} \right), \nn 
\end{align}
where $(a)$ follows from our definition of $h$ in \eqref{eq:data_hetero}.
\end{proof}

\begin{lemma}
\label{prop:test_acc_min_gamma}

Let $A_j := \{ r \in [m]: \inner{\bw_{j,r}^{(0)}}{j \bmu} \geq 0\}$. For any $0 \leq t \leq T^*-1$ we have,

 \begin{enumerate}
     \item \label{eq:test_acc_gamma_max} For any $j \in \{ \pm 1\}, r \in [m]:$ $\ggam^{(t)} \leq \frac{\eta \norm{\bmu} }{m} \sum_{t'=0}^{t-1}\sum_{s=0}^{\tau - 1} \max_{k, i} \abs{\lderiv^{(t',s)}}$. 

     \item \label{eq:test_acc_min_gamma_aligned} For any $r \in A_j:$ $\ggam^{(t)} \geq \frac{\eta \norm{\bmu}}{4m}\sum_{t'=0}^{t-1}\sum_{s=0}^{\tau - 1} \min_{(k,i)} \abs{\lderiv^{(t',s)}}$.

     \item \label{eq:test_acc_min_gamma_misaligned} For any $r \notin A_j:$ $ \ggam^{(t)} \geq \frac{\eta \norm{\bmu}}{4m}\sum_{t'=0}^{t-1} \left( \min_{k, i} \abs{\lderiv^{(t',0)}} + h\sum_{s=1}^{\tau-1} \min_{k, i} \abs{\lderiv^{(t',s)}} \right)$.

\end{enumerate}

\end{lemma}

\qed

\textbf{Proof.} 

Unrolling the iterative update in \eqref{eq:Gamma_jr} we have,
\begin{align}
    \ggam^{(t)} & = \frac{\eta}{nm} \sum_{t'=0}^{t-1} \sum_{s=0}^{\tau -1} \sum_{k,i} (-\lderiv^{(t',s)}) \cdot \signalderiv{\lmweight^{(t',s)}}\cdot\norm{\bmu}. \label{eq:gamma_simplified_2}
\end{align}

\textit {Proof of \eqref{eq:test_acc_gamma_max}}. Using \eqref{eq:gamma_simplified_2}, we can get an upper bound on $\ggam^{(t)}$ as follows.
\begin{align}
  \ggam^{(t)} \leq \frac{\eta \norm{\bmu} }{m} \sum_{t'=0}^{t-1}\sum_{s=0}^{\tau - 1} \max_{k, i} \abs{\lderiv^{(t',s)}}, \nn   
\end{align}
where the inequality follows from $\sigma'(\cdot) \leq 1$.

\textit{Proof of \eqref{eq:test_acc_min_gamma_aligned}.} From \Cref{lemma:gamma_ip_always_pos} we know that if $\inner{\bw_{j,r}^{(0)}}{j \bmu} \geq 0$ then $\inner{\bw_{j,r}^{(t')}}{j \bmu} \geq 0$ for all $t' \geq 0$. Thus using \eqref{eq:test_acc_gamma_min_init_pos} repeatedly for all $0 \leq t' \leq t-1$ we get,
\begin{align}
   \ggam^{(t)} & \geq \frac{\eta \norm{\bmu}}{4m}\sum_{t'=0}^{t-1}\sum_{s=0}^{\tau - 1} \min_{k, i} \abs{\lderiv^{(t',s)}}. \nn 
\end{align}

\textit{Proof of \eqref{eq:test_acc_min_gamma_misaligned}.}  Note that the bound in \eqref{eq:test_acc_gamma_min_init_neg} holds even if $\inner{\bw_{j,r}^{(t-1)}}{j \bmu} \geq 0$. Thus applying \eqref{eq:test_acc_gamma_min_init_neg} repeatedly for all $0 \leq t' \leq t-1$ we get,

\begin{align}
   \ggam^{(t)} & \geq \frac{\eta \norm{\bmu}}{4m}\sum_{t'=0}^{t-1} \left( \min_{k, i} \abs{\lderiv^{(t',0)}} + h\sum_{s=1}^{\tau-1} \min_{k, i} \abs{\lderiv^{(t',s)}} \right). \nn 
\end{align}

\begin{lemma}
\label{lemma:rhop_upper_and_lower}
Under assumptions, for any $0 \leq t \leq T^*-1$ we have,

\begin{enumerate}
    \item \label{eq:sum_rhop_upper_bound}
   $\sum_{k,i} \gprho^{(t)} \leq \frac{3\eta \sigma_p^2 d}{2m}\sum_{t'=0}^{t-1} \sum_{s = 0}^{\tau-1} \max_{k,i} \abs{\lderiv^{(t',s)}}$. 

   \item \label{eq:sum_rhop_lower_bound}
    $\sum_{k,i} \gprho^{(t)} \geq \frac{\eta \sigma_p^2 d}{16m}\sum_{t'=0}^{t-1} \sum_{s = 0}^{\tau-1} \min_{(k,i) \in \tilde{S}_{j,r}^{(t',s)}} \abs{\lderiv^{(t',s)}}$
\end{enumerate}

where $ \tilde{S}_{j,r}^{(t',s)} := \left\{ k \in [K], i \in [N]: \inner{\lmweight^{(t',s)}}{\bxi_{k,i}} \geq  0\right\}$.
\end{lemma}

\textbf{Proof.}

From \eqref{eq:overbar_P_jrki} we have the following update equation for $\gprho^{(t)}$.

\begin{align}
\label{eq:test_acc_grhop_update}
    \sum_{k,i} \gprho^{(t)} & = \sum_{k,i} \gprho^{(t-1)} - \frac{\eta}{nm}\sum_{s = 0}^{\tau-1} \sum_{k,i: y_{k,i} = j} \lderiv^{(t-1,s)}\cdot\noisederiv{\lmweight^{(t-1,s)}}\cdot\norm{\bxi_{k,i}} \nonumber \\
    & = \sum_{k,i} \gprho^{(t-1)} - \frac{\eta}{nm}\sum_{s = 0}^{\tau-1} \sum_{(k,i) \in \tilde{S}_{j,r}^{(t-1,s)}} \lderiv^{(t-1,s)}\cdot\norm{\bxi_{k,i}}.
\end{align}
where the last equality follows from the definition of $\tilde{S}_{j,r}^{(t,s)}$.

\textit{Proof of \eqref{eq:sum_rhop_upper_bound}}. Now using \eqref{eq:test_acc_grhop_update} we have,
\begin{align}
    \sum_{k,i} \gprho^{(t)} \overset{(a)}{\leq} \sum_{k,i} \gprho^{(t-1)} + \frac{3\eta \sigma_p^2 d}{2m}\sum_{s = 0}^{\tau-1} \max_{k,i} \abs{\lderiv^{(t-1,s)}} \nonumber
\end{align}
where $(a)$ follows from \Cref{lemma:noise_corr}. Unrolling the recursion above we have the following upper bound,
\begin{align}
% \label{eq:test_acc_grhop_max}
  \sum_{k,i} \gprho^{(t)} \leq \frac{3\eta \sigma_p^2 d}{2m}\sum_{t'=0}^{t-1} \sum_{s = 0}^{\tau-1} \max_{k,i} \abs{\lderiv^{(t',s)}}.  \nn  
\end{align}

\textit{Proof of \eqref{eq:sum_rhop_lower_bound}}. From \eqref{eq:test_acc_grhop_update} we have,
\begin{align}
    \sum_{k,i} \gprho^{(t)} \overset{(a)}{\geq} \sum_{k,i} \gprho^{(t-1)} + \frac{\eta \sigma_p^2 d}{16m}\sum_{s = 0}^{\tau-1} \min_{(k,i) \in \tilde{S}_{j,r}^{(t-1,s)}} \abs{\lderiv^{(t-1,s)}} \nonumber
\end{align}
where $(a)$ follows from \Cref{lemma:noise_corr} and \Cref{prop:bounded_deriv_frac} part \ref{prop_bounded_deriv_frac_5} which implies $\abs{\tilde{S}_{j,r}^{(t-1,s)}} \geq n/8$. Unrolling the recursion above we have,
\begin{align}
% \label{eq:test_acc_grhop_min}
  \sum_{k,i} \gprho^{(t)} \geq \frac{\eta \sigma_p^2 d}{16m}\sum_{t'=0}^{t-1} \sum_{s = 0}^{\tau-1} \min_{(k,i) \in \tilde{S}_{j,r}^{(t',s)}} \abs{\lderiv^{(t',s)}}. \nn 
\end{align}
\qed

\begin{lemma}
\label{lemma:filter_aligned_after_T1}
    For all $t \geq T_1$, we have $\inner{\bw_{y,r}^{(t)}} {y\bmu} > 0$.
\end{lemma}

\begin{proof}
    We have,
    \begin{align}
        \inner{\bw_{y,r}^{(t)}} {y\bmu} &= \inner{\bw_{y,r}^{(0)}} {y\bmu} + \ggam^{(t)} \nonumber \\
        & \overset{(a)}{\geq} -\bigtheta{\sqrt{\log(m/\delta)} \cdot \sigma_0 \normt{\bmu}} + \ggam^{(t)} \nonumber \\
        & \overset{(b)}{\geq} -\bigtheta{\sqrt{\log(m/\delta)} \cdot \sigma_0 \normt{\bmu}} +  \frac{\eta \norm{\bmu}}{4m}\sum_{t'=0}^{T_1-1} \min_{k, i} \abs{\lderiv^{(t',0)}} \nonumber \\
        & \overset{(c)}{=} -\bigtheta{\sqrt{\log(m/\delta)} \cdot \sigma_0 \normt{\bmu}} +  \bigw{\frac{n \norm{\bmu}}{\sigma_p^2 d \tau}} \nonumber \\
        & \overset{(d)}{\geq} \bigtheta{\sqrt{\log(m/\delta)} \cdot \frac{\sqrt{n} \normt{\bmu}}{\sigma_p d \tau}} +  \bigw{\frac{n \norm{\bmu}}{\sigma_p^2 d \tau}} \nonumber \\
        & \overset{(e)}{\geq} 0.
    \end{align}
Here $(a)$ follows from \Cref{lemma:init_filter_corr}; $(b)$ follows from \Cref{prop:test_acc_min_gamma}; $(c)$ follows from the definition of $T_1$ in \Cref{eq:T_1_defn}; $(d)$ follows from \Cref{assump:sigma_0}; $(e)$ follows from \Cref{assump:mu_strength} and \Cref{assump:m_n}.
\end{proof}

\begin{lemma}
\label{lemma:w_0_signal_norm_bound}
Under \Cref{assum:main_assump}, for any $T_1 \leq t \leq T^*-1$ we have,

\begin{enumerate}[leftmargin=*]
    \item \label{eq:w_0_noise_upper_bound}
    $\frac{\normt{\bw_{j,r}^{(0)}}}{\bigtheta{\sigma_p^{-1}d^{-1/2}n^{-1/2}}\sum_{k,i} \gprho^{(t)}} = \bigO{1}$ 

    \item \label{eq:snr_upper_bound}
    $\frac{\ggam^{(t)} \normt{\bmu}^{-1}}{\bigtheta{\sigma_p^{-1}d^{-1/2}n^{-1/2}}\sum_{k,i} \gprho^{(t)}} = \bigO{1}$
\end{enumerate}
\end{lemma}

\begin{proof}[Proof of \eqref{eq:w_0_noise_upper_bound}]
Note from our proof of \Cref{lemma:activated_noise_filters}, we know that for all $T_1 \leq t \leq T^*-1$ we have $ \overline{P}_{j,r,k^*,i^*}^{(t)} \geq 2$ for all $(k^*,i^*) \in \tilde{S}_{j,r}^{(0)} = \left\{k \in [K], i \in [N]: y_{k,i} = j, \inner{\bw_{j,r,k}^{(0)}}{\bxi_{k,i}} \geq 0 \right\}$.
Thus,
\begin{align}
    \sum_{k,i}  \gprho^{(t)}  & \geq 2 \abs{\tilde{S}_{j,r}^{(0)}} \overset{(a)}{=} \bigw{n}, \label{eq:sum_rho_omega_n}
\end{align}
where $(a)$ follows from \Cref{lemma:min_size_activated_noise_data}.
This implies, 
\begin{align}
    \frac{\normt{\bw_{j,r}^{(0)}}}{\bigtheta{\sigma_p^{-1}d^{-1/2}n^{-1/2}}\sum_{k,i} \gprho^{(t)}}  & \overset{(a)}{=}  \frac{\bigtheta{\sigma_0 \sqrt{d}}}{\bigtheta{\sigma_p^{-1}d^{-1/2}n^{-1/2}}\sum_{k,i} \gprho^{(t)}} \nonumber\\
    & \overset{(b)} = \bigO{\sigma_0 \sigma_p d n^{-1/2}} \nonumber \\
    & \overset{(c)}{=} \bigO{1}. \nn 
\end{align}
Here $(a)$ follows from \Cref{lemma:init_filter_corr}; $(b)$ follows from \eqref{eq:sum_rho_omega_n}; $(c)$ follows from \Cref{assump:sigma_0}. 
\end{proof}

\begin{proof}[Proof of \eqref{eq:snr_upper_bound}]
From \Cref{lemma:gamma_upper_and_lower} and \Cref{lemma:rhop_upper_and_lower} we have,
\begin{align}
    \frac{\ggam^{(t)} }{\sum_{k,i} \gprho^{(t)}} & \leq \frac{16\norm{\bmu}}{\sigma_p^2 d}\frac{\sum_{t'=0}^{t-1}\sum_{s=0}^{\tau - 1} \max_{k, i} \abs{\lderiv^{(t',s)}}}{\sum_{t'=0}^{t-1} \sum_{s = 0}^{\tau-1} \min_{(k,i) \in \tilde{S}_{j,r}^{(t',s)}} \abs{\lderiv^{(t',s)}}} \overset{(a)}{\leq} \frac{16 C_2\norm{\bmu}}{\sigma_p^2 d }, \nn 
\end{align}
where $(a)$ follows from \Cref{prop:bounded_deriv_frac} part \ref{prop_bounded_deriv_frac_3} which implies $\max_{k, i} \abs{\lderiv^{(t'-1,s)}} \leq C_2 \min_{(k,i) \in \tilde{S}_{j,r}^{(t'-1,s)}}  \abs{\lderiv^{(t'-1,s)}} $ for all $0 \leq t' \leq T^*-1, 0\leq s\leq \tau - 1$.
Thus,
\begin{align}
    \frac{\ggam^{(t)} \normt{\bmu}^{-1}}{\bigtheta{\sigma_p^{-1}d^{-1/2}n^{-1/2}} \sum_{k,i} \gprho^{(t)}} &= \bigO{\frac{n^{1/2}\normt{\bmu}}{\sigma_p d^{1/2}}} \overset{(a)}= \bigO{1}. \nn 
\end{align}
where $(a)$ follows from \Cref{assump:d}. 
\end{proof}

\begin{lemma}
\label{lemma:sigma_rhop_bound}
For any $T_1 \leq t \leq T^* -1$ we have,
\begin{align}
% \label{eq:sigma_rhop_bound}
    \frac{ \sum_{r} \sigma \left( \inner{\bw_{y,r}^{(t)}} {y\bmu} \right)}{\sum_{r,k,i} \overline{P}_{-y,r,k,i}^{(t)}} \geq \frac{C_4 
    \norm{\bmu}}{\sigma_p^2 m d} \left(|A_y| + (m-|A_y|)\left(h + \frac{1}{\tau}(1-h)\right)\right), \nn 
\end{align}
where $C_4 > 0$ is some constant.
\end{lemma}

\textbf{Proof}.

We can write,
\begin{align}
\label{eq:sum_relu_filter_signal}
    \sum_{r} \sigma \left( \inner{\bw_{y,r}^{(t)}} {y\bmu} \right) =  \underbrace{\sum_{r: \inner{\bw_{y,r}^{(0)}}{y \bmu} \geq 0} \sigma \left( \inner{\bw_{y,r}^{(t)}} {y\bmu} \right)}_{I_1} + \underbrace{\sum_{r: \inner{\bw_{y,r}^{(0)}}{y \bmu} < 0} \sigma \left( \inner{\bw_{y,r}^{(t)}} {y\bmu} \right)}_{I_2}.
\end{align}

First note that if $\inner{\bw_{y,r}^{(0)}}{y \bmu} \geq 0$ then from \Cref{lemma:gamma_ip_always_pos} we know that ,
\begin{align}
\label{eq:pos_filter_always_pos}
    \inner{\bw_{y,r,k}^{(t,s)}}{y \bmu} \geq 0 \text{ for all } k \in [K], 0 \leq t \leq T^*-1, 0 \leq s \leq \tau-1.
\end{align}
We can bound $I_1$ as follows:
\begin{align}
\label{eq:filter_signal_I_1_bound}
    I_1 &= \sum_{r: \inner{\bw_{y,r}^{(0)}}{y \bmu} \geq 0} \sigma \left( \inner{\bw_{y,r}^{(t)}} {y\bmu} \right) \nonumber \\
    & \overset{(a)}{=} \sum_{r: \inner{\bw_{y,r}^{(0)}}{y \bmu} \geq 0} \inner{\bw_{y,r}^{(t)}} {y\bmu} \nonumber \\
    & \overset{(b)}{\geq} \sum_{r: \inner{\bw_{y,r}^{(0)}}{y \bmu} \geq 0} \Gamma_{y,r}^{(t)} \nonumber \\
    & \overset{(c)} = \bigw{|A_y|  \eta \norm{\bmu} \sum_{t'=0}^{t-1}\sum_{s=0}^{\tau - 1} \min_{ k, i} \abs{\lderiv^{(t',s)}}}.
\end{align}
Here $(a)$ follows from \eqref{eq:pos_filter_always_pos}; $(b)$ follows from \Cref{lemma:measure_signal_coeff}; $(c)$ follows from \Cref{prop:test_acc_min_gamma} part \ref{eq:test_acc_min_gamma_aligned}.
For $I_2$, we have the following bound:
\begin{align}
\label{eq:filter_signal_I_2_bound}
    I_2 &= \sum_{r: \inner{\bw_{y,r}^{(0)}}{y \bmu} < 0} \sigma \left( \inner{\bw_{y,r}^{(t)}} {y\bmu} \right) \nonumber \\
     & \overset{(a)}{\geq} \sum_{r: \inner{\bw_{y,r}^{(0)}}{y \bmu} < 0}  \inner{\bw_{y,r}^{(0)}} {y\bmu} + \ggam^{(t)} \nonumber \\
     & \overset{(b)}{\geq}  -(m-|A_y|)\bigtheta{\sqrt{\log(m/\delta)} \cdot \sigma_0 \normt{\bmu}}  + \sum_{r: \inner{\bw_{y,r}^{(0)}}{y \bmu} < 0}  \ggam^{(t)} \nonumber \\
     & \overset{(c)}{=} \bigw{\sum_{r: \inner{\bw_{y,r}^{(0)}}{y \bmu} < 0}  \ggam^{(t)} } \nonumber \\
    & \overset{(d)}{\geq}  \Omega \Bigg( (m-|A_y|)\eta \norm{\bmu} \left(\sum_{t'=0}^{T_1-1} \min_{k,i} \abs{\lderiv^{(t',0)}} + h\sum_{t'=0}^{T_1-1}\sum_{s=1}^{\tau-1}  \min_{k,i} \abs{\lderiv^{(t',s)}} \right) \nonumber \\
    & \hspace{5pt} + (m-|A_y|) \eta \norm{\bmu} \sum_{t'=T_1}^{t-1}\sum_{s=0}^{\tau - 1} \min_{ k, i} \abs{\lderiv^{(t',s)}} \Bigg).
\end{align}

Here $(a)$ follows from $\sigma(z) \geq z$; $(b)$ follows from \Cref{lemma:init_filter_corr} and \Cref{assump:sigma_0}; $(c)$ follows from \Cref{lemma:filter_aligned_after_T1}; $(d)$ follows from \Cref{prop:test_acc_min_gamma}.
Substituting \eqref{eq:filter_signal_I_1_bound} and \eqref{eq:filter_signal_I_2_bound} in \eqref{eq:sum_relu_filter_signal} we have,
\begin{align}
\label{eq:sum_relu_filter_signal_bound}
    \sum_{r} \sigma \left( \inner{\bw_{y,r}^{(t)}} {y\bmu} \right) & \geq \Omega \Bigg( |A_y|\eta \norm{\bmu} \sum_{t'=0}^{t-1}\sum_{s=0}^{\tau -1} \min_{k,i} \abs{\lderiv^{(t',s)}} \nn \\
    & + (m-|A_y|)\eta \norm{\bmu} \left(\sum_{t'=0}^{T_1-1} \min_{k,i} \abs{\lderiv^{(t',0)}} + h\sum_{t'=0}^{T_1-1}\sum_{s=1}^{\tau-1}  \min_{k,i} \abs{\lderiv^{(t',s)}} \right) \nonumber \\
    & + (m-|A_y|) \eta \norm{\bmu} \sum_{t'=T_1}^{t-1}\sum_{s=0}^{\tau - 1} \min_{ k, i} \abs{\lderiv^{(t',s)}} \Bigg)
\end{align}

Now using \eqref{eq:sum_relu_filter_signal_bound} and \Cref{lemma:rhop_upper_and_lower} we have,

\begin{align}
    & \frac{ \sum_{r} \sigma \left( \inner{\bw_{y,r}^{(t)}} {y\bmu} \right)}{\sum_{r,k,i} \overline{P}_{-y,r,k,i}^{(t)}} \nn \\
    & \overset{(a)}{\geq} \Omega \Bigg( \frac{ \norm{\bmu}}{\sigma_p^2 md}\Bigg( |A_y| \frac{\sum_{t'=0}^{t-1}\sum_{s=0}^{\tau -1} \min_{k, i} \abs{\lderiv^{(t',s)}}}{\sum_{t'=0}^{t-1} \sum_{s = 0}^{\tau-1} \max_{k,i} \abs{\lderiv^{(t',s)}}} \nonumber \\
    & \hspace{5pt} + (m-|A_y|) \frac{\sum_{t'=0}^{T_1-1} \left(\min_{k,i} \abs{\lderiv^{(t',0)}} + h\sum_{s=1}^{\tau-1}  \min_{k,i}\abs{\lderiv^{(t',s)}} \right) + \sum_{t'=0}^{t-1}\sum_{s=0}^{\tau -1} \min_{k, i} \abs{\lderiv^{(t',s)}} } {\sum_{t'=0}^{T_1-1} \sum_{s = 0}^{\tau-1} \max_{k,i} \abs{\lderiv^{(t',s)}} + \sum_{t'=T_1}^{t-1} \sum_{s = 0}^{\tau-1} \max_{k,i} \abs{\lderiv^{(t',s)}}}  \Bigg) \Bigg) \nonumber \\
    & \overset{(b)}{\geq} \Omega \Bigg( \frac{ 
    \norm{\bmu}}{\sigma_p^2 m d} \left(|A_y| + (m-|A_y|)\left(h + \frac{1}{\tau}(1-h)\right)\right) \Bigg) \nn 
\end{align}
where $(a)$ follows from \Cref{lemma:rhop_upper_and_lower}; $(b)$ follows from \Cref{prop:bounded_deriv_frac} part \ref{prop_bounded_deriv_frac_3} and \Cref{eq:first_stage_min_deriv}. 

\qed

\begin{lemma}
\label{lemma:sigma_r_w_0_bound}
Under assumptions, for all $T_1 \leq t \leq T^*-1$ we have
\begin{align}
  \frac{\sum_{r} \sigma \left( \inner{\bw_{y,r}^{(t)}} {y\bmu} \right)}{\sigma_p \sum_{r=1}^m \normt{\bw_{-y,r}^{(t)}}} \geq \bigtheta{\frac{n^{1/2}\norm{\bmu}  }{\sigma_p^2 m d^{1/2}}  \left(|A_y| + (m-|A_y|)\left(h + \frac{1}{\tau}(1-h)\right)\right) }. \nn 
\end{align}
    
\end{lemma}

\begin{proof}
To prove this, we first show that $\normt{\gmweight^{(t)}} = \bigO{\sigma_p^{-1}d^{-1/2}n^{-1/2}} \cdot \sum_{k,i} \gprho^{(t)}$ for all $j \in \{ \pm 1\}$. 

We first bound the norm of the noise components as follows.
\begin{align}
    &\norm{\sum_{k,i} \grho^{(t)}\cdot\normt{\bxi_{k,i}}^{-2}\cdot \bxi_{k,i}} \nonumber\\
    & = \sum_{k,i} \left(\grho^{(t)}\right)^2\cdot\normt{\bxi_{k,i}}^{-2} + 2\sum_{k,k'> k, i, i'> i} P_{j,r,k,i}^{(t)} P_{j,r,k',i'}^{(t)} \cdot\normt{\bxi_{k,i}}^{-2} \cdot\normt{\bxi_{k',i'}}^{-2} \cdot \inner{\bxi_{k,i}}{\bxi_{k',i'}} 
    \nonumber \\
    & \overset{(a)}{\leq} 4\sigma_p^{-2}d^{-1}\sum_{k,i}\left(\grho^{(t)}\right)^2 + 2 \sum_{k,k'> k, i, i'> i} \abs{P_{j,r,k,i}^{(t)} P_{j,r,k',i'}^{(t)}}(16\sigma_p^{-4}d^{-2})(2\sigma_p^2 \sqrt{d \log (6n^2/\delta)}) \nonumber \\
    &= 4\sigma_p^{-2}d^{-1}\sum_{k,i}\left(\grho^{(t)}\right)^2 + 32\sigma_p^{-2}d^{-3/2}\left( \left( \sum_{k,i} \abs{\grho^{(t)}}\right)^2 - \sum_{k,i} \left( \grho^{(t)} \right)^2\right) \nonumber\\
    & = \bigtheta{\sigma_p^{-2}d^{-1}}\sum_{k,i} \left(\grho^{(t)}\right)^2 + \bigthetat{\sigma_p^{-2}d^{-3/2}}\left( \sum_{k,i} \abs{\grho^{(t)}}\right)^2 \nonumber \\
    & \overset{(b)}{\leq} \left[ \bigtheta{\sigma_p^{-2}d^{-1}} + \bigthetat{\sigma_p^{-2}d^{-3/2}}\right] \left(\sum_{k,i} \abs{\gprho^{(t)}} + \sum_{k,i} \abs{\gnrho^{(t)}}\right)^2 \nonumber \\
    & = \bigtheta{\sigma_p^{-2}d^{-1}n^{-1}}\left(\sum_{k,i} \gprho^{(t)} \right)^2. \label{eq:sum_rhop_norm_bound}
\end{align}
Here for $(a)$ uses \Cref{lemma:noise_corr}; $(b)$ uses $\max_{j,r,k,i} \abs{\gnrho^{(t)}} \leq  \beta + 8\sqrt{\frac{\log (6n^2/\delta)}{d}}n\alpha = \bigO{1}$ from \Cref{thm:coeff_bound} and so $\sum_{k,i} \abs{\gnrho^{(t)}}  = \bigO{\sum_{k,i} \gprho^{(t)}}$. 
Now from \eqref{eq:gmweight_decomp} we know that,
\begin{align}
    \gmweight^{(t)} = \gmweight^{(0)} + j\ggam^{(t)}\cdot\normt{\bmu}^{-2}\bmu + \sum_{k=1}^2\sum_{i \in [N]}\grho^{(t)}\cdot\normt{\bxi_{k,i}}^{-2}\cdot\bxi_{k,i}. \nn 
\end{align}
Using triangle inequality and \eqref{eq:sum_rhop_norm_bound} we have,
\begin{align}
    \normt{ \gmweight^{(t)}} & \leq \normt{\gmweight^{(0)}} + \ggam^{(t)}\normt{\bmu}^{-1} + \bigtheta{\sigma_p^{-1}d^{-1/2}n^{-1/2}}\sum_{k,i} \gprho^{(t)} \nonumber \\
    & \overset{(a)}{=} \bigtheta{\sigma_p^{-1}d^{-1/2}n^{-1/2}}\sum_{k,i} \gprho^{(t)} \nn 
\end{align}
where $(a)$ follows from \Cref{lemma:w_0_signal_norm_bound}. 

Thus,
\begin{align}
    \frac{\sum_{r} \sigma \left( \inner{\bw_{y,r}^{(t)}} {y\bmu} \right)}{\sigma_p \sum_{r=1}^m \normt{\bw_{-y,r}^{(t)}}} &\geq \frac{\sum_{r} \sigma \left( \inner{\bw_{y,r}^{(t)}} {y\bmu} \right)}{\bigtheta{d^{-1/2}n^{-1/2}}\sum_{k,i} \gprho^{(t)}} \nonumber \\
    & \overset{(a)}{=} \bigtheta{\frac{n^{1/2}\norm{\bmu}  }{\sigma_p^2 m d^{1/2}} \left(|A_y| + (m-|A_y|)\left(h + \frac{1}{\tau}(1-h)\right)\right) } \nn 
\end{align}
where $(a)$ follows from \Cref{lemma:sigma_rhop_bound}. 
\end{proof}

\begin{lemma} (sub-result in Theorem E.1 in \cite{cao2022benign}.)
\label{lemma:g_xi}
    Denote $g(\bxi) = \sum_{r} \sigma \left(\inner{\bw_{-y,r}^{(t)}}{\bxi} \right)$. Then for any $x \geq 0$ it holds that
    \begin{align}
        \Pr(g(\bxi) - \expt g(\bxi) > x) \leq \exp \left( - \frac{cx^2}{\sigma_p^2 \left(\sum_{r=1}^m \normt{\bw_{-y,r}^{(t)}}\right)^2}\right) \nn 
    \end{align}
\end{lemma}
where $c$ is a constant and $\expt g(\bxi) = \frac{\sigma_p}{\sqrt{2\pi}}\sum_{r=1}^m \normt{\bw_{-y,r}^{(t)}}$.

\subsection{Test Error Upper Bound}

\label{subsec:test_error_upper_bound_proof}
We now prove the upper bound on our test error in the benign overfitting regime as stated in \Cref{thm:test_error}.

First note that for some given $(\bx,y)$ we have, 
\begin{align}
    \mathbb{P}(y \neq \mathrm{sign}(f(\bW^{(t)},\bx)) = \mathbb{P}(y f(\bW^{(t)},\bx) \leq 0). \nn 
\end{align}

We can write,
\begin{align}
\label{eq:yf_equation}
   & y f(\bW^{(t)},\bx) =  F_{y}(\bW_y^{(t)},\bx) - F_{-y}(\bW_{-y}^{(t)},\bx) \nonumber \\
   & = \frac{1}{m}\sum_{r=1}^m \left[ \sigma\left(\inner{\bw_{y,r}^{(t)}}{y\bmu}\right) + \sigma\left( \inner{\bw_{y,r}^{(t)}}{\bxi} \right)\right] - \frac{1}{m}\sum_{r=1}^m \left[ \sigma\left(\inner{\bw_{-y,r}^{(t)}}{y\bmu}\right) + \sigma\left( \inner{\bw_{-y,r}^{(t)}}{\bxi} \right) \right].
\end{align}

Now note that since $t \geq T_1$ we know that $\sigma\left(\inner{\bw_{-y,r}^{(t)}}{y\bmu}\right) = 0$ for all $r \in [m]$ from \Cref{lemma:filter_aligned_after_T1}. Thus, 

\begin{align}
   \mathbb{P}(y f(\bW^{(t)},\bx) \leq 0) & \leq \mathbb{P}\left(\sum_{r=1}^m \sigma\left( \inner{\bw_{-y,r}^{(t)}}{\bxi} \right) \geq  \sum_{r=1}^m \sigma\left(\inner{\bw_{y,r}^{(t)}}{y\bmu}\right)\right) \nonumber \\
   & \overset{(a)}{=} \mathbb{P}\left(g(\bxi) - \expt g(\bxi) \geq  \sum_{r=1}^m \sigma\left(\inner{\bw_{y,r}^{(t)}}{y\bmu}\right) - \frac{\sigma_p}{\sqrt{2\pi}}\sum_{r=1}^m \normt{\bw_{-y,r}^{(t)}}\right) \nonumber \\
   & \overset{(b)}{\leq} \exp \left( -\frac{c\left( \sum_{r=1}^m \sigma\left(\inner{\bw_{y,r}^{(t)}}{y\bmu}\right) - \frac{\sigma_p}{\sqrt{2\pi}}\sum_{r=1}^m \normt{\bw_{-y,r}^{(t)}}\right)^2}{\sigma_p^2 \left(\sum_{r=1}^m \normt{\bw_{-y,r}^{(t)}}\right)^2}
   \right) \nonumber\\
   & = \exp\left( -c \left(\frac{\sum_{r=1}^m \sigma\left(\inner{\bw_{y,r}^{(t)}}{y\bmu}\right)}{\sigma_p \sum_{r=1}^m \normt{\bw_{-y,r}^{(t)}}} - \frac{1}{\sqrt{2\pi}}\right)^2 \right) \nonumber\\
   & \overset{(c)}{\leq} \exp \left(\frac{c}{2\pi} - \frac{c}{2} \left( \frac{\sum_{r=1}^m \sigma\left(\inner{\bw_{y,r}^{(t)}}{y\bmu}\right)}{\sigma_p \sum_{r=1}^m \normt{\bw_{-y,r}^{(t)}}}\right)^2 \right) \nonumber \\
   & \overset{(d)}{\leq} \exp \left(\frac{c}{2\pi} - \frac{n\normt{\bmu}^4\left(|A_y| + (m-|A_y|)\left(h + \frac{1}{\tau}(1-h)\right)\right) ^2}{C_5 \sigma_p^4 m^2 d} \right) \nonumber \\
   & \overset{(e)}{\leq} \exp \left(- \frac{n\normt{\bmu}^4\left(|A_y| + (m-|A_y|)\left(h + \frac{1}{\tau}(1-h)\right)\right) ^2}{2C_5 \sigma_p^4 m^2 d} \right). \nn 
\end{align}
Here $(a)$ follows from the definition of $g(\bxi)$ in \Cref{lemma:g_xi}; $(b)$ follows from the result in \Cref{lemma:g_xi}; $(c)$ uses $(a-b)^2 \geq a^2/2 - b^2, \forall a,b \geq 0$; $(d)$ uses \Cref{lemma:sigma_r_w_0_bound}; $(e)$ follows from the benign overfitting condition $n \normt{\bmu}^4 = \bigw{\sigma_p^4 d}$ and choosing sufficiently large $C_6$. Now note that,
\begin{align}
    L_{\mathcal{D}}^{0-1}(\bW^{(T)}) &= \sum_{j \in \{ \pm 1\}} \mathbb{P}(y = j) \mathbb{P}(y \neq \mathrm{sign}(f(\bW^{(t)},\bx)) \nonumber \\
    & = \frac{1}{2}\sum_{j \in \{\pm 1\}} \exp \left(- \frac{n\normt{\bmu}^4\left(|A_j| + (m-|A_j|)\left(h + \frac{1}{\tau}(1-h)\right)\right) ^2}{2C_5 \sigma_p^4 m^2 d} \right). \nn 
\end{align}

This completes our proof for the upper bound on the test error in the benign overfitting regime.

\subsection{Test Error Lower Bound}

We first state some intermediate lemmas that we use in our proof.

\begin{lemma} (Lemma 5.8 in \cite{kou2023benign})
\label{lemma:lemma_5.8}
Let $g(\bxi) = \sum_{j,r} j \sigma \left( \inner{\bw_{j,r}^{(T)}}{\bxi}\right)$. If $n\normt{\bmu}^4 = \bigO{\sigma_p^4d}$ (harmful overfitting condition) then there exists a fixed vector $v$ with $\norm{\bv} \leq 0.06 \sigma_p$ such that
\begin{align}
    \sum_{j' \in \{\pm 1\}} \left[g(j'\bxi+\bv) - g(j'\bxi) \right] \geq 4C_6 \max_{j \in \{\pm 1\} }\left\{ \sum_{r} \Gamma_{j,r}^{(T)} \right\} \nn 
\end{align}
for all $\bxi \in \mathbb{R}^d$.
\end{lemma}

\begin{lemma}(Proposition 2.1 in \cite{devroye2018total})
\label{lemma:tv_distance}
    The $\mathrm{TV}$ distance between $\mathcal{N}(\mathbf{0}, \sigma_p^2 \mathbf{I}_d)$ and $\mathcal{N}(\mathbf{v}, \sigma_p^2 \mathbf{I}_d)$ is less than $\norm{\bv}/2\sigma_p$.
\end{lemma}

\textbf{Proof.}

We have,

\begin{align}
\label{eq:test_error_0}
     & L_{\mathcal{D}}^{0-1}(\bW^{(T)}) \nonumber\\
     &= \mathbb{P}_{(\bx,y) \sim \mathcal{D}} \left(y \neq \mathrm{sign}(f(\bW, \bx))\right) \nonumber\\
     & = \mathbb{P}_{(\bx,y) \sim \mathcal{D}} \left(yf(\bW, \bx) \leq 0\right) \nonumber \\
     & \overset{(a)}{=}  \mathbb{P}_{(\bx,y) \sim \mathcal{D}} \left(\sum_{r} \sigma\left(\inner{\bw_{-y,r}^{(T)}}{\bxi}\right) - \sum_{r} \sigma\left( \inner{\bw_{y,r}^{(T)}}{\bxi} \right) \geq  \sum_{r} \sigma\left(\inner{\bw_{y,r}^{(T)}}{y\bmu}\right) - \sum_{r} \sigma\left( \inner{\bw_{-y,r}^{(T)}}{y\bmu} \right)\right) \nonumber \\
      & \overset{(b)}{\geq}  \mathbb{P}_{(\bx,y) \sim \mathcal{D}} \left(\sum_{r} \sigma\left(\inner{\bw_{-y,r}^{(T)}}{\bxi}\right) - \sum_{r} \sigma\left( \inner{\bw_{y,r}^{(T)}}{\bxi} \right) \geq  C_6 \max \left\{ \sum_{r} \Gamma_{1,r}^{(T)}, \sum_{r} \Gamma_{-1,r}^{(T)} \right\}\right) \nonumber \\
     & \geq 0.5 \mathbb{P}_{(\bx,y) \sim \mathcal{D}} \left( \abs{\sum_{r} \sigma\left(\inner{\bw_{1,r}^{(T)}}{\bxi}\right) - \sum_{r} \sigma\left( \inner{\bw_{-1,r}^{(T)}}{\bxi} \right)} \geq C_6 \max \left\{ \sum_{r} \Gamma_{1,r}^{(T)}, \sum_{r} \Gamma_{-1,r}^{(T)} \right\}\right) \nonumber \\
     & \overset{(c)}{=}  0.5 \mathbb{P}_{(\bx,y) \sim \mathcal{D}} \left( \abs{g(\bxi)} \geq C_6 \max \left\{ \sum_{r} \Gamma_{1,r}^{(T)}, \sum_{r} \Gamma_{-1,r}^{(T)} \right\}\right) \nonumber \\
     & \overset{(d)}{=} 0.5 \mathbb{P} (\Omega).
\end{align}
Here $(a)$ follows from \eqref{eq:yf_equation};
$\mathbb{P}(y \neq \mathrm{sign}(f(\bW^{(t)},\bx)) = \mathbb{P}(y f(\bW^{(t)},\bx) \leq 0)$; $(b)$ follows from $\sigma\left(\inner{\bw_{-y,r}^{(t)}}{y\bmu}\right) = 0$ (\Cref{lemma:filter_aligned_after_T1}) and $\sigma\left(\inner{\bw_{y,r}^{(t)}}{y\bmu}\right) = \bigtheta{\Gamma_{y,r}^{(t)}}$; $(c)$ follows from defining $g(\bxi) = \sum_{r} \sigma\left(\inner{\bw_{1,r}^{(T)}}{\bxi}\right) - \sum_{r} \sigma\left( \inner{\bw_{-1,r}^{(T)}}{\bxi} \right)$; $(d)$ follows from defining $\Omega := \left\{ \bxi: \abs{g(\bxi)} \geq C_6  \max \left\{ \sum_{r} \Gamma_{1,r}^{(T)}, \sum_{r} \Gamma_{-1,r}^{(T)} \right\}\right\}$.

Now we know from Lemma \Cref{lemma:lemma_5.8}, that $\sum_{j} \left[(g(j\bxi + \bv) - g(j\bxi) \right] \geq 4C_6 \max_j\{\sum_{r} \Gamma_{j,r}^{(T)}\}$. This implies that one one of the $\bxi, \bxi+v, -\bxi, -\bxi+v$ must belong to $\Omega$. Therefore,
\begin{align}
\label{eq:test_error_1}
    \min \left\{ \mathbb{P}(\Omega), \mathbb{P}(-\Omega), \mathbb{P}(\Omega - {\bv}), \mathbb{P}(-\Omega-{\bv})\right\} \geq 0.25
\end{align}

Also note that by symmetry $\mathbb{P}(\Omega) = \mathbb{P}(-\Omega)$. Furthermore,
\begin{align}
\label{eq:test_error_2}
    \abs{\mathbb{P} \left(\Omega \right) - \mathbb{P} \left(\Omega -\bv\right)} & = \abs{\mathbb{P}_{\bxi \sim \mathcal{N}(\mathbf{0},\sigma_p^2\mathbb{I}_d}) (\bxi \ \in \Omega) - \mathbb{P}_{\bxi \sim \mathcal{N}(\bv,\sigma_p^2\mathbf{I}_d)} (\bxi \in \Omega)]} \nonumber\\
    & \overset{(a)}{\leq} \mathrm{TV} \left( \mathcal{N}(\mathbf{0},\sigma_p^2\mathbf{I}_d), \mathcal{N}(\bv,\sigma_p^2\mathbf{I}_d)\right) \nonumber \\
    & \overset{(b)}{\leq} \frac{\norm{\bv}}{2\sigma_p} \nonumber \\
    & \leq 0.03.
\end{align}

Here $(a)$ follows from the definition of $\mathrm{TV}$ distance; $(b)$ follows from Lemma \Cref{lemma:tv_distance}.
Thus we see that \eqref{eq:test_error_2} along with \eqref{eq:test_error_1} implies that $\mathbb{P}(\Omega) = 0.22$. Substituting this in \eqref{eq:test_error_0} we get $L_{\mathcal{D}}^{0-1}(\bW^{(T)})  = 0.1$ as claimed.

\section{Main Paper Lemma Proofs}

\subsection{\texorpdfstring{Proof of \Cref{lem:main_paper_lemma_signal_growth}}{Proof of Lemma}}
This lemma follows from directly from \Cref{prop:test_acc_min_gamma} and the constant lower bound on cross-entropy loss derivatives, i.e., \Cref{eq:first_stage_min_deriv}. 

\subsection{\texorpdfstring{Proof of \Cref{lem:main_paper_noise_growth}}{Proof of Lemma}}
This lemma follows from directly from \Cref{lemma:rhop_upper_and_lower} and the constant lower bound on cross-entropy loss derivatives, i.e., \Cref{eq:first_stage_min_deriv}. 

\subsection{\texorpdfstring{Proof of \Cref{lemma:pretrain}}{Proof of Lemma}}
Using our result in \Cref{lemma:gamma_upper_and_lower} with $\tau = 1$ and $h = 0$, we have after $T_1 = \bigO{\frac{mn}{\eta \sigma_p^2 d}}$ iterations for all $j \in \{\pm 1\}$ and $r \in [m]$,
\begin{align}
    \Gamma_{j,r}^{(\pre, T_1)} & \geq \frac{\eta \norm{\bmu^{(\pre)}}}{4m}\sum_{t=0}^{T_1-1} \min_{i} \abs{{\ell'}_i^{(\pre, t)}} \overset{(a)}{\geq}
    \frac{\eta \norm{\bmu^{(\pre)}}CT_1}{4m} = \bigw{\frac{ n \norm{\bmu^{(\pre)}}}{\sigma_p^2 d}}. \nn 
\end{align}
Here $(a)$ follows from \eqref{eq:first_stage_min_deriv}.
Now for any $t \geq T_1$ we have from \Cref{lemma:measure_signal_coeff},
\begin{align}
    \inner{\bw^{(\text{pre},t)}_{j,r}}{j\bmu^{(\text{pre})}} & = \inner{\bw^{(\text{pre},0)}_{j,r}}{j\bmu^{(\text{pre})}} +  \Gamma_{j,r}^{(\pre, t)} \nonumber \\
    & \overset{(a)}{\geq} \inner{\bw^{(\text{pre},0)}_{j,r}}{j\bmu^{(\text{pre})}} +  \Gamma_{j,r}^{(\pre, T_1)} \nonumber \\
    & \overset{(b)}{\geq} -\bigtheta{\sqrt{\log(m/\delta)} (\sigma_p d)^{-1}\sqrt{n} \normt{\bmu^{(\pre)}}} + \bigw{\sigma_p^{-2}d^{-1}n \norm{\bmu^{(\pre)}}} \nn  \\
    & \overset{(c)}{\geq} 0, \nn 
\end{align}
where $(a)$ follows from the fact that $\Gamma_{j,r}^{(t)}$ is non-decreasing with respect to $t$, $(b)$ follows from \Cref{assump:sigma_0} and \Cref{lemma:init_filter_corr}; $(c)$ follows from \Cref{assump:mu_strength}.

\qed

\section{Additional Experimental Details}
\label{sec:expt_details}

% \subsection{Details for Figures and Tables in Main Paper}

\paragraph{Implementation.} We use PyTorch \cite{paszke2019pytorch} to run all our algorithms and also simulate our synthetic data setting. For experiments on neural network training we use one H100 GPU with $2$ cores and $20$GB memory. For synthetic data experiments we use one T4 GPU. The approximate total run-time for all our experiments on neural networks is about $36$ hours. The approximate total run-time for all experiments on the synthetic data setting is about $1$ hour.

\paragraph{Details for \Cref{fig:test_acc_gap}.} We simulate a FL setup with $K = 10$ clients on the CIFAR10 data partitioned using Dirichlet$(\alpha)$ with $\alpha = 0.1$ for the non-IID setting and $\alpha = 10$ for the IID setting. For pre-training, we consider a Squeezenet model pre-trained on ImageNet \cite{russakovsky2015imagenet} which is available in PyTorch. Following \cite{nguyen2022begin} we replace the BatchNorm layers in the model with GroupNorm \cite{wu2018group}. For FL optimization we use the vanilla \ts{FedAvg} optimizer with server step size $\eta_g = 1$ and train the model for $500$ rounds and $1$ local epoch at each client. For centralized optimization we use \ts{SGD} optimizer and run the optimization for $200$ epochs. Learning rates were tuned using grid search with the grid $\{0.1, 0.01, 0.001\}$. Final accuracies were reported after averaging across $3$ random seeds.

\paragraph{Details for and \Cref{fig:emp_verification} and \Cref{fig:two_layer_cnn_expts}.} For these experiments we simulate a synthetic data setup following our data model in \Cref{sec:problem_setup}. We set the dimension $d = 200$, $n = 20$ datapoints (we keep $n$ small to ensure we are in the over-parameterized regime), $m = 10$ filters, $K=2$ clients, $N = 10$ local datapoints. The signal strength is $\norm{\bmu} = 3$, noise variance is $\sigma_p^2 = 0.1$ and variance of Gaussian initialization is $\sigma_0 = 0.01$. The global dataset has $10$ datapoints with positive labels and $10$ datapoints with negative labels. We also create a test dataset of $1000$ datapoints following the same setup to evaluate our test error.

\paragraph{Details for \Cref{fig:real_world_misalignment} and \Cref{fig:varying_heterogeneity}.} We simulate a FL setup with $K = 20$ clients using Dirichlet$(\alpha)$ \cite{hsu2019noniid}. For pre-training, we consider a ResNet18 model pre-trained on ImageNet \cite{russakovsky2015imagenet} which is available in PyTorch. Following \cite{nguyen2022begin} we replace the BatchNorm layers in the model with GroupNorm \cite{wu2018group}. For FL optimization we use the \ts{FedAvg} optimizer with server step size $\eta_g = 1$ and $1$ local epoch at each client. In the case of random initiation, for local optimization we use SGD optimizer with a learning rate of $0.01$ and $0.9$ momentum. In the case of pre-trained initiation, for local optimization we use SGD optimizer with a learning rate of $0.001$ and $0.9$ momentum. The learning rate is decayed by a factor of $0.998$ in every round in the case for both initializations. Each experiment is repeated with $3$ different random seeds.\\

\end{document}